%% file: 00_main.tex
\documentclass{article} 
\usepackage{tinypaper,times}

\input{math_commands.tex}

\usepackage{hyperref}
\usepackage{url}

\hypersetup{
    colorlinks=true,
    linkcolor=blue,
    filecolor=magenta,      
    urlcolor=blue,
    pdftitle={Metrics as Transform},
    pdfpagemode=FullScreen,
    }

\usepackage{microtype}
\usepackage{graphicx}

\usepackage{caption}
\usepackage{subcaption}

\usepackage{booktabs} 
\usepackage{multicol}
\usepackage{multirow}

\usepackage[frozencache,cachedir=minted]{minted}

\usepackage{algorithmic}
\usepackage{algorithm}

\usepackage{amsmath}
\usepackage{amssymb}
\usepackage{mathtools}
\usepackage{amsthm}

\usepackage[textsize=tiny]{todonotes}

\usepackage{amssymb}

\usepackage{enumitem}
\usepackage{wrapfig}

\title{Metric as Transform: Exploring beyond \\
Affine Transform for Interpretable Neural Network}

\author{Suman Sapkota \thanks{Code available at \href{https://github.com/tsumansapkota/Metric-Local-Neuron}{github}.} \\
Independent Researcher
}

\iclrfinalcopy 
\begin{document}

\maketitle

\begin{abstract}

Artificial Neural Networks of varying architectures are generally paired with affine transformation at the core. However, we find dot product neurons with global influence less interpretable as compared to local influence of euclidean distance (as used in Radial Basis Function Network). In this work, we explore the generalization of dot product neurons to $l^p$-norm, metrics, and beyond. We find that metrics as transform performs similarly to affine transform when used in MultiLayer Perceptron or Convolutional Neural Network. 
Moreover, we explore various properties of Metrics, compare it with Affine, and present multiple cases where metrics seem to provide better interpretability.
We develop an interpretable local dictionary based Neural Networks and use it to understand and reject adversarial examples.

\end{abstract}

\input{01_introduction}

\section{Exploring Properties of Metrics and Beyond}
\label{sec:exploring_metrics_and_properties}

First, we analyze various properties of Metrics and how it could help us with interpretability as compared to linear transforms generally used in most neural network architectures.

\input{02_on_metrics_as_neuron}

\input{03_convolution_with_linear_vs_l2}
\input{04_voronoi_diagram_of_transforms}
\input{05_metric_layers_on_DNN}
\input{06_generalized_distance_as_neuron}
\input{07_on_embeddings_activation_viz}

\input{08_invertibility_of_metrics}
\input{09_invertibility_of_generalized_distances}

\input{10_properties_of_metrics}

\section{Application of Metrics for Locally Interpretable Neurons}

With amazing properties of metrics and ability to replace dense linear transform in neural network if normalized properly, we focus on use of metrics for interpretable MLP. We want to create architectures or mechanisms to easily modify and interpret MLPs. We mainly focus on a single layer of MLP for rigorous understanding and experimentation.

\paragraph{Motivation} The main motivation for building locally activating neuron is to steer away from the global effect of Linear+ReLU neurons. The problem with globally activating neuron, as already discussed in Section~\ref{sec:exploring_metrics_and_properties}, is that any addition or removal of neuron will have global consequences and the neuron does not have a unique finite maxima. With locally activating neuron we hope to overcome these problems.

In this section, we primarily focus on building locally interpretable neurons, and show its application.

\input{11_Dictionary_Learning_MLP}

\input{12_local_neuron_extension}
\input{13_uncertainty_estimation_and_adv_rejection}
\input{14_noisy_center_search_and_data_init}

\section{Discussion and Conclusion}
We find that various metrics as transform work in the modern Neural Network architectures with use of proper normalization. Moreover, we find multiple cases where metrics provide better interpretation of neurons in neural network and provide way to initialize neurons with data. This work also studies generalization of metrics and their properties. We build models dependent on data, including uncertainty prediction and guarantee maximization of particular neuron for particular input.

 We believe that metrics could increase intrepretability and allow for low-level modification of large models as well. 

\subsubsection*{Acknowledgements}
The preliminary work on using metrics as transform in neural network was supported by Dr. Binod Bhattarai while at NAAMII. We thank reviewers of Tiny Papers @ ICLR 2023 for constructive feedback and Mr. Ashish Paudel for his assistance in solving for inverse of distances and angles. This work was further supported by Ambika Sapkota, Bibek Bajagain, Manoj Sapkota, Shishir Acharya and Vibek Adhikari.


\bibliography{biblography}
\bibliographystyle{tinypaper}

\end{document}

%% file: math_commands.tex
\usepackage{amsmath,amsfonts,bm}

\def\eqref#1{equation~\ref{#1}}

\def\1{\bm{1}}

\def\va{{\bm{a}}}
\def\vb{{\bm{b}}}

\def\vd{{\bm{d}}}

\def\vg{{\bm{g}}}

\def\vi{{\bm{i}}}

\def\vk{{\bm{k}}}

\def\vq{{\bm{q}}}
\def\vr{{\bm{r}}}
\def\vs{{\bm{s}}}

\def\vv{{\bm{v}}}
\def\vw{{\bm{w}}}
\def\vx{{\bm{x}}}
\def\vy{{\bm{y}}}
\def\vz{{\bm{z}}}

\def\mA{{\bm{A}}}

\DeclareMathAlphabet{\mathsfit}{\encodingdefault}{\sfdefault}{m}{sl}
\SetMathAlphabet{\mathsfit}{bold}{\encodingdefault}{\sfdefault}{bx}{n}

\newcommand{\Var}{\mathrm{Var}}



%% file: 01_introduction.tex
\section{Introduction}

Artificial Neural Networks (ANN) are used end-to-end and generally as black-box function approximators. This is partly due to the vast number of parameters, the underlying function used, and the high dimension of input and hidden neurons. The backbone of Deep Networks including MLP, CNN~\cite{krizhevsky2012imagenet}, Transformers~\cite{vaswani2017attention}, and MLP-Mixers~\cite{tolstikhin2021mlp} has been a linear transform of form $\vy = \mathbf{W}\vx+\vb$ (or per neuron: $y_i = \vw_i.\vx+b$).

There have been explorations of other operations such as $l^1$-norm~\cite{chen2020addernet} and $l^2$-norm~\cite{li2022euclidnets} for transformations in Neural Networks much guided by the computational efficiency. However, matrix multiplication has won both software and hardware-lottery~\cite{hooker2021hardware}, is highly optimized and tightly integrated in current Deep Neural Networks. Since other transformations are not explored much, we explore a form of norms and generalized measure of distance as a neuron, which we call metrics - of the form $y_i = f_{metric}(\vx, \vw) + b$. Here, $f_{metrics}$ can be norm function (e.g. $l^p$), a proper metric function, or any generalization. It is possible to use $f_{metrics}$ for measuring similarity as a negative or inverse distance. 

Since metrics have a point of minima, the lower contour set of metrics is generally bounded and the minima represents that two points are similar. Norm space can be easily extended to metric space if we take $d(\vx, \vy) = f_{norm}(\vx-\vy)$. We can also generalize the metrics by relaxing different axioms (see Section~\ref{sec:metric_properties}).

In this paper, we explore various properties of metrics such as voronoi partitioning, metrics as transform, their generalization, invertibility, application in low dimensional embeddings, dictionary learning, adversarial rejection and noisy optimization.

%% file: 02_on_metrics_as_neuron.tex
\subsection{On Artificial Neurons Based on Linear and Radial function}

\begin{figure*}[h]
  \centering

\includegraphics[width=0.39\textwidth]{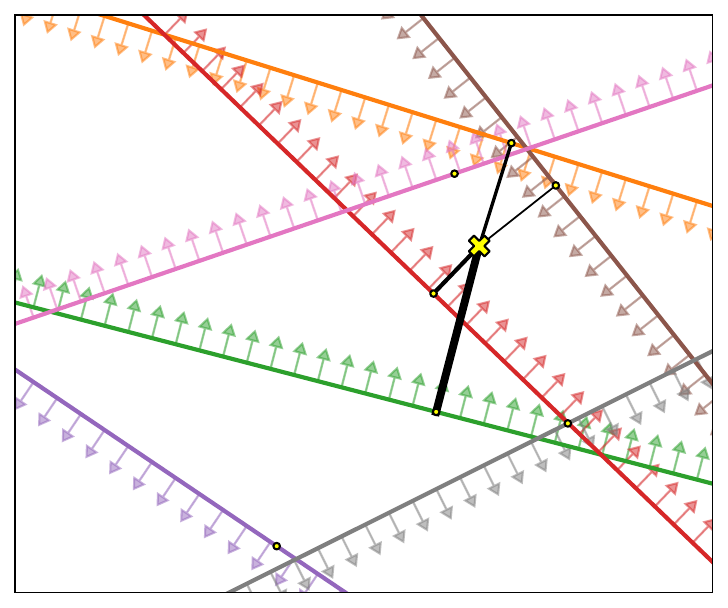}
\includegraphics[width=0.39\textwidth]{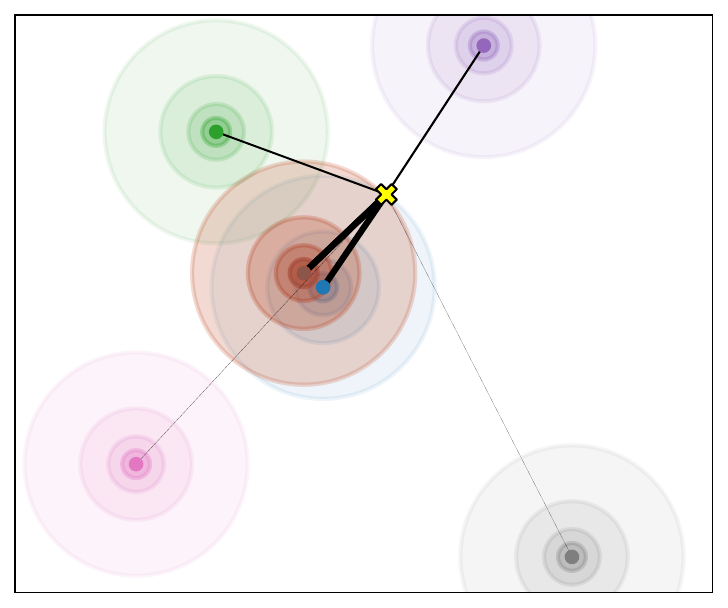}

  \caption{Neuron Interpretation: \textit{(LEFT)} 2D ReLU neuron at $\boldsymbol{w.x}+b \ge 0$,  \textit{(RIGHT)} 2D radial neuron at $\exp{(-\|\boldsymbol{x-w}\|^2)}$. The lines or circles show region of neuron firing, ($\times$) showing a data point and thickness of the lines indicating its activation magnitude.
  \textcolor{magenta}{Zoom in for details.}
  }
  \label{fig:viz_neuron_activation}
\end{figure*}

The dot product is simple to understand but we find the dot product neuron, with non-linear activation, is difficult to interpret. It represents a planar neuron rather than a local neuron as shown in Figure~\ref{fig:viz_neuron_activation}. Local neurons are generally found on Radial Basis Function (RBF) Network~\cite{broomhead1988multivariable} and generally consists of euclidean distance operation, which are not generally applied in ANNs. RBF has also been used as inspiration for interpretable manifold manipulation~\cite{olah2014neural}. 
All this motivates us to explore the area of metrics that give a sense of distance between input and weights (or centers) in neural networks.

Moreover, if we compare the properties of linear neuron and radial neuron, we find strikingly different properties.
The activation of linear neuron with relu activation is not bounded, and the maximum value is produced at infinity ($\infty$, or very high values of input). A certain value of activation corresponds to a plane (a line in 2D) of possible inputs, which are again unbounded.

Comparatively, the activation of radial neuron with gaussian activation is bounded, the maximum value is produced when center and input are same. A certain value of activation corresponds to a sphere (a circle in 2D) of possible inputs, which are bounded.

We can interpret the activations using similarity metric, which can be maximized by a single point only. We can visualize what a neuron represents by maximizing its activation. Linear neurons fail in that they do not have finite maxima. Previous works have also used neuron maximization to visualize what it represents~\cite{olah2017feature}.

Moreover, we can use any function with finite minima as a generalized measure of distance. Figure~\ref{fig:contour_plots_local_vs_global} shows variation of uniform, convex~\cite{amos2017input} and invex~\cite{sapkota2021input} function. We are concerned with functions with a single finite global minima, and can be used as a generalized distance function. The distance from minima can also be interpreted as similarity(e.g. $f(d) = e^{-d^2}$) and produces maximum value ($\simeq 1$) for a particular point or region. We discuss more on generalized metrics in Section~\ref{sec:generalized_metrics_mlp},\ref{sec:reverse_generalized_metrics} and \ref{sec:metric_properties}.

%% file: 03_convolution_with_linear_vs_l2.tex
\subsection{On use of Metrics for Convolution}

\begin{figure*}[h]
  \centering

\includegraphics[width=0.32\textwidth]{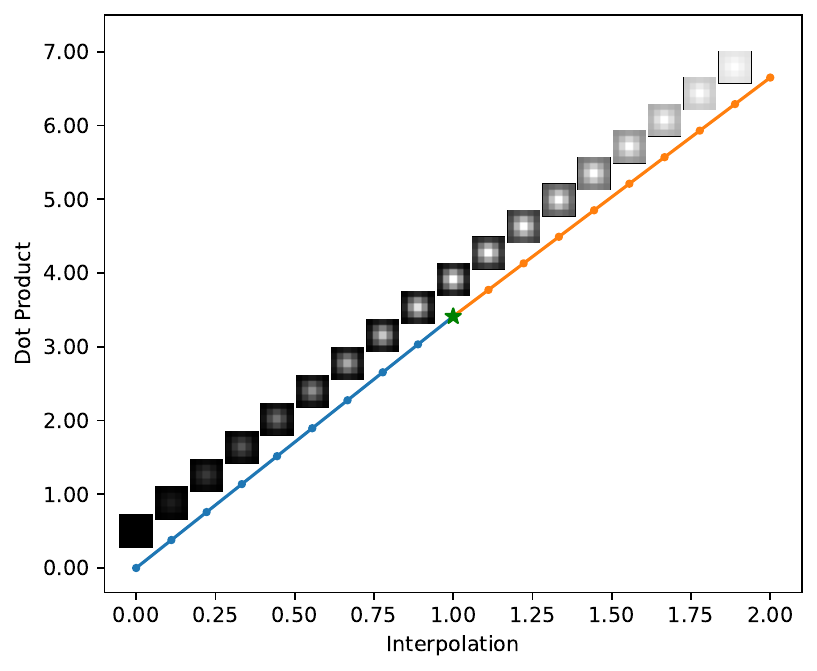}
\includegraphics[width=0.32\textwidth]{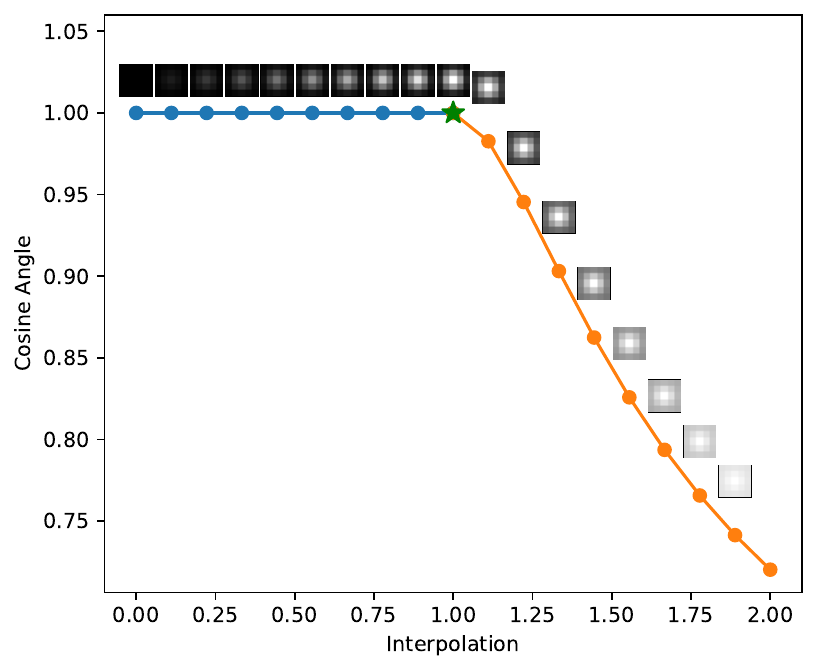}
\includegraphics[width=0.32\textwidth]{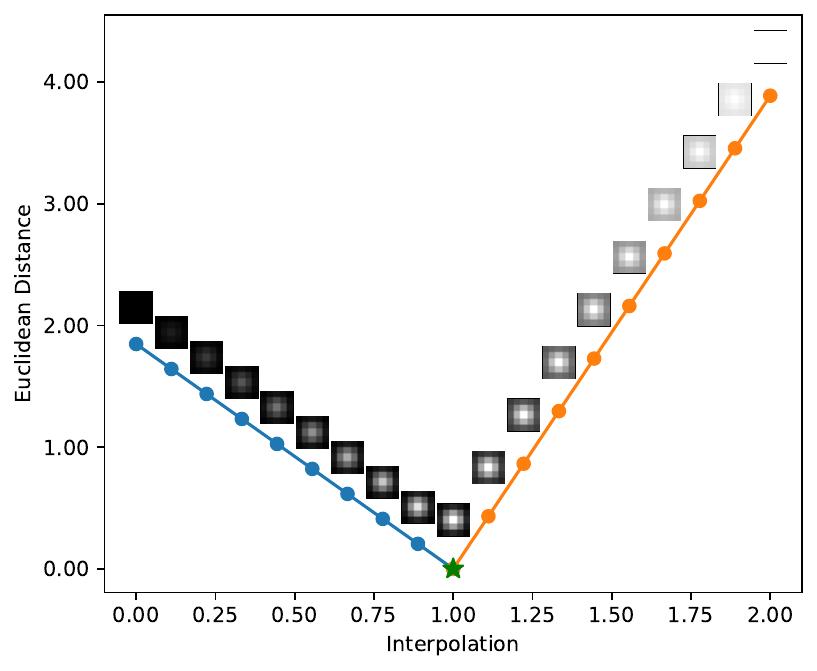}

  \caption{
  Dot product, cosine angle and distance between patches of image when the values are interpolated between black-gaussian-white. The images just above the scatter-points represents the interpolated image. The star ($\star$) symbol represents the reference point to which every interpolation is compared.
  \textcolor{magenta}{Zoom in for details.}
  }
  \label{fig:image_dot_vs_dist}
\end{figure*}

Similar to unstructured input space, convolution operation can be interpreted as a measure of similarity of reference patch (weight) and some input patch. We compare dot product and metrics, namely with $l^2$-norm and cosine angle, on Figure~\ref{fig:image_dot_vs_dist}. For individual activation, we find $l^2$ metric more interpretable as it gives minimum value if input and target match, and higher if they are farther away. Here, cosine angle does not differentiate between different intensity of the image patch. Dot product gives higher activation for larger intensity of input, more than dot product with reference input itself (i.e. alpha=1).

%% file: 04_voronoi_diagram_of_transforms.tex
\subsection{Voronoi Diagram of Transforms}

\begin{figure*}[h]
  \centering

\includegraphics[width=0.31\textwidth]{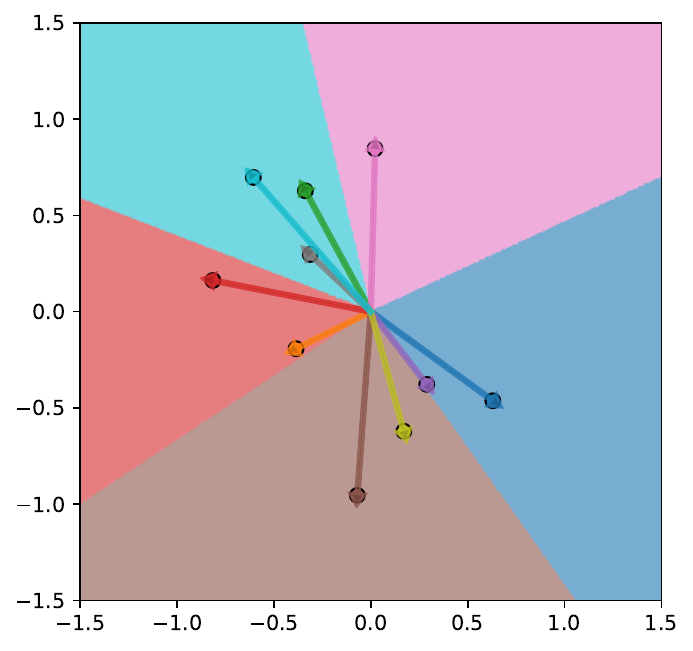}
\includegraphics[width=0.31\textwidth]{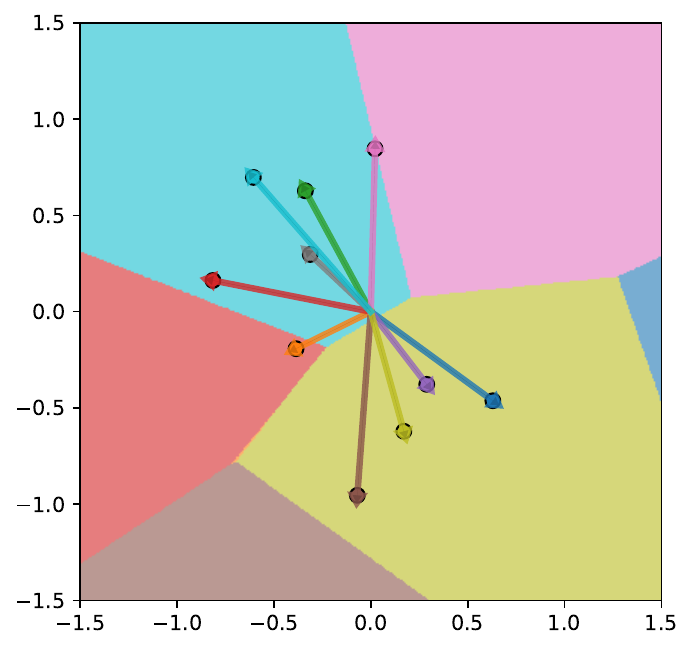}
\includegraphics[width=0.31\textwidth]{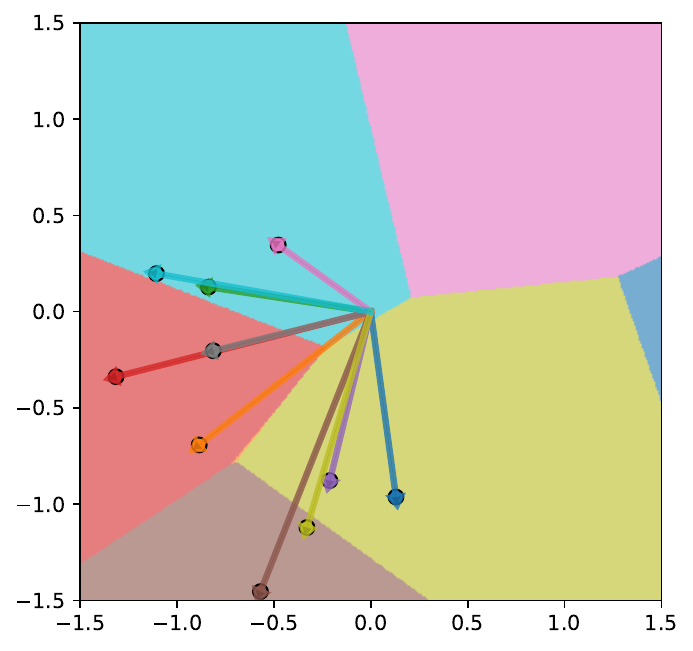}

\includegraphics[width=0.31\textwidth]{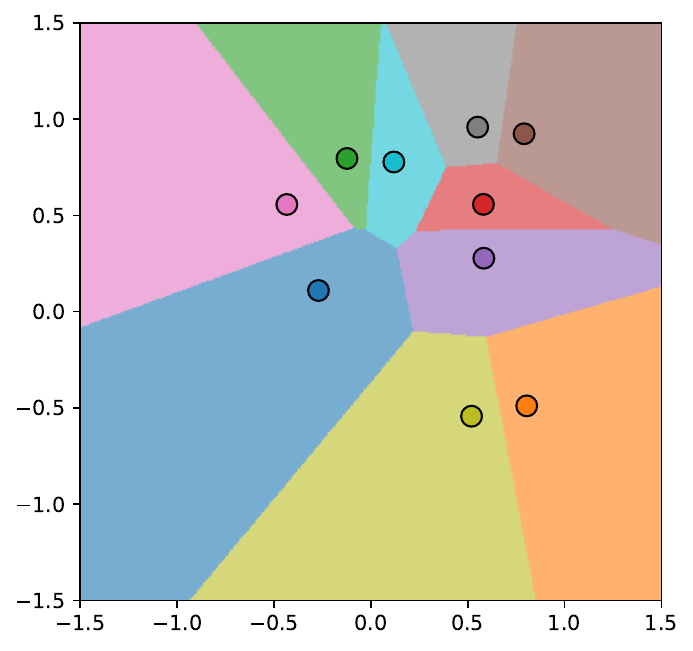}
\includegraphics[width=0.31\textwidth]{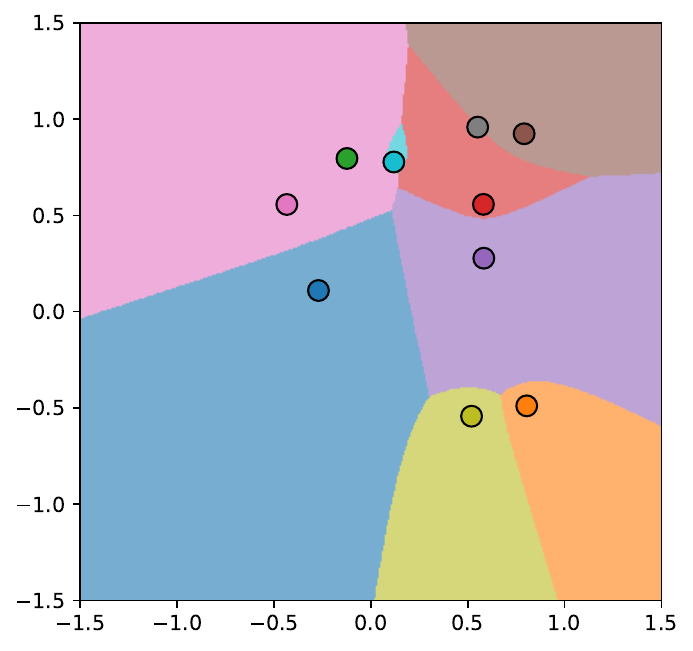}
\includegraphics[width=0.31\textwidth]{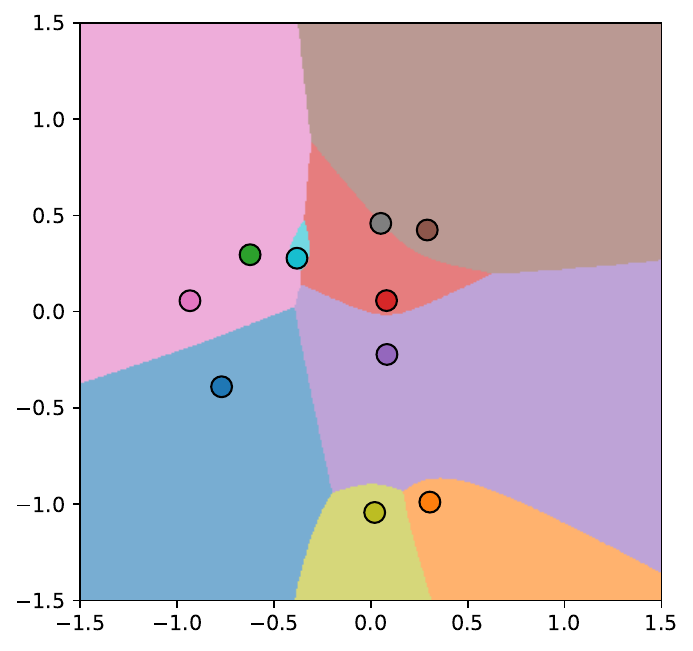}

\caption{
Voronoi Diagram of \textbf{(TOP)} Linear and \textbf{(BOT)} Distance for \textit{LEFT}: without using bias, \textit{MID}: using bias, and \textit{RIGHT}: using bias and shifting the center/weights by $[-0.5, -0.5]$
}
  \label{fig:voronoi_partitioning}
\end{figure*}

In clustering or classification, the decision boundary of sets and regions of input are assigned to a cluster or a class. Generally, regions are based on distance or linear transforms which produce Voronoi partitioning~\cite{voronoi1908nouvelles, fortune2017voronoi}. Each set or region is represented by unique neuron of the transform. In Figure~\ref{fig:voronoi_partitioning}, we compare Voronoi diagrams of linear transform and distance as transform, show the effect of bias use, and the effect of shift in parameters of the transformation. 

Although both linear and distance transforms produce general Voronoi partitioning, centers in distance based transform lie inside their respective sets, which is more interpretable than linear Voronoi sets.

%% file: 05_metric_layers_on_DNN.tex
\subsection{Application of Metrics as Transform in Deep Neural Network}
\label{sec:metric_layer}

\begin{table*}[h]
    \centering
    \caption{
    Training MLP with different types of transformations at first layer on 2-layered MLP using different number of hidden neurons ($H$). The given measurements are test accuracy in format \textit{mean$\pm$std}(\textcolor{blue}{max}).
    } 
    \vspace{2mm}
    \resizebox{0.99\linewidth}{!}{
\begin{tabular}{@{}rccccc@{}}

\toprule 

Method & $H=5$ & $H=10$ & $H=20$  & $H=100$ & $H=500$ \\
\midrule
$l^{0.5}$ & 72.92$\pm$1.07 (\textcolor{blue}{74.46}) & 78.06$\pm$0.41 (\textcolor{blue}{78.79}) & 80.13$\pm$0.34 (\textcolor{blue}{80.84}) & 83.16$\pm$0.05 (\textcolor{blue}{83.22}) & 84.74$\pm$0.20 (\textcolor{blue}{85.13})  \\
$l^{1}$ & 76.98$\pm$0.70 (\textcolor{blue}{77.89}) & 81.71$\pm$0.14 (\textcolor{blue}{81.92}) & 82.95$\pm$0.27 (\textcolor{blue}{83.43}) & 85.36$\pm$0.10 (\textcolor{blue}{85.52}) & 87.02$\pm$0.17 (\textcolor{blue}{87.35}) \\
$l^{2}$ & 77.84$\pm$0.76 (\textcolor{blue}{78.92}) & 82.86$\pm$0.18 (\textcolor{blue}{83.22}) & 83.72$\pm$0.26 (\textcolor{blue}{84.16}) & 86.16$\pm$0.09 (\textcolor{blue}{86.31}) & 87.67$\pm$0.08 (\textcolor{blue}{87.74}) \\
$l^{20}$ & 79.39$\pm$0.80 (\textcolor{blue}{80.45}) & 82.90$\pm$0.49 (\textcolor{blue}{83.62}) & 83.97$\pm$0.26 (\textcolor{blue}{84.34}) & 85.80$\pm$0.15 (\textcolor{blue}{86.02}) & 87.63$\pm$0.11 (\textcolor{blue}{87.79}) \\
i-stereo & 80.90$\pm$0.60 (\textcolor{blue}{81.79}) & 84.96$\pm$0.18 (\textcolor{blue}{85.25}) & 86.32$\pm$0.17 (\textcolor{blue}{86.59}) & 88.23$\pm$0.17 (\textcolor{blue}{88.43}) & 89.32$\pm$0.10 (\textcolor{blue}{89.52}) \\
\midrule
linear & 81.17$\pm$0.33 (\textcolor{blue}{81.70}) & 84.92$\pm$0.21 (\textcolor{blue}{85.28}) & 86.47$\pm$0.11 (\textcolor{blue}{86.67}) & 88.42$\pm$0.09 (\textcolor{blue}{88.55}) & 89.72$\pm$0.07 (\textcolor{blue}{89.80}) \\
\midrule
convex & 88.29$\pm$0.12 (\textcolor{blue}{88.48}) & 88.96$\pm$0.20 (\textcolor{blue}{89.30}) & 88.90$\pm$0.09 (\textcolor{blue}{88.99}) & 88.48$\pm$0.18 (\textcolor{blue}{88.75}) & 88.49$\pm$0.15 (\textcolor{blue}{88.67}) \\
invex & 88.54$\pm$0.26 (\textcolor{blue}{88.96}) & 89.29$\pm$0.12 (\textcolor{blue}{89.51}) & 88.18$\pm$0.81 (\textcolor{blue}{89.42}) & 88.19$\pm$0.26 (\textcolor{blue}{88.48}) & 89.11$\pm$0.29 (\textcolor{blue}{89.48}) \\
ordinary & 84.35$\pm$0.49 (\textcolor{blue}{84.88}) & 85.91$\pm$0.26 (\textcolor{blue}{86.26}) & 86.14$\pm$0.58 (\textcolor{blue}{87.24}) & 87.09$\pm$0.37 (\textcolor{blue}{87.55}) & 87.50$\pm$0.21 (\textcolor{blue}{87.92}) \\

\bottomrule
    \end{tabular}
    }
    \label{tab:metrics_in_fmnist}
\end{table*}

\begin{wraptable}{r}{0.2\linewidth}
    \centering
\vspace{-2em}
    \caption{
    Maximum test accuracy among 4 runs with various transforms. 
    }
    
\vspace{-0.5em}
\begin{tabular}{@{}rl@{}}
\toprule 
Transform & Accuracy \\
\midrule
$l^2$ & 93.06 \\
$i$-stereo & 92.71 \\
linear & 92.8 \\
\bottomrule
\end{tabular}
\label{tab:metrics_in_resnet}
\vspace{-1em}
\end{wraptable}
We try replacing linear layers in MLP with metrics and find that different types of metrics as transform simply work. We identify that normalization like BatchNorm~\cite{ioffe2015batch} and LayerNorm~\cite{ba2016layer} or Softmax~\cite{bishop2006pattern} helps in optimization, without which we could not efficiently optimize the network. Moreover, normalization of the activation with uniform scale and shift doesn't change the Voronoi set of the neurons. 

\textbf{$l^p$-norm:} We can use $l^p$-norm~\cite{lp-space} with any $p$ value as transform. The general equation of $l^p$-norm is : $\|\vx\|_p = (|x_1|^p+|x_2|^p+ \dots + |x_n|^p)^{1/p}$. The norm is convex and valid norm induced metric for $p \ge 1$, however for $p < 1$, it is not convex. Regardless, we can use it as general a measure of distance.

\textbf{$i$-stereo:} We can use inverse stereographic transform~\cite{stereographic_projection} to project N dimensional euclidean space into N+1 dimension Sphere. The angular measure on the Sphere is a metric and invertible as shown in Section~\ref{sec:reverse_metrics}.

The experiments on Fashion-MNIST dataset in Table~\ref{tab:metrics_in_fmnist} shows that various metrics perform similar to dot product. Moreover, other metrics have not been explored and optimized much. With optimization (like AdderNet~\cite{chen2020addernet}), we believe that the performance can increase. Here, we test on 2 layered MLP (\textcolor{gray}{with configuration as:} \textit{Layer1}$\to$\textit{BatchNorm}$\to$\textit{LayerNorm}$\to$\textit{ELU activation}~\cite{clevert2015fast}$\to$\textit{Linear}) with \textit{Layer1} replaced with various metrics.

We were also able to use some of these metrics as direct replacement of linear kernels in CNN (ResNet-20) trained on on CIFAR-10~\cite{krizhevsky2009learning} dataset for 200 epochs with batch size of 128. The accuracy is reported in Table~\ref{tab:metrics_in_resnet}. Here as well, we find metric based convolution performing similar to the dot product.

%% file: 06_generalized_distance_as_neuron.tex
\subsection{Generalized notion of distance beyond Metrics}
\label{sec:generalized_metrics_mlp}

The generalized distances such as learnable convex function~\cite{amos2017input}, invex function~\cite{sapkota2021input, nesterov2022learning}, or any function produce satisfactory results when used as transform in Neural Network. Although these functions are not truly metrics, they serve as generalized notion of distance. We discuss this on Section~\ref{sec:metric_properties}

\begin{figure*}[h]
  \centering

\scalebox{0.9}[1.0]{\includegraphics[width=0.31\textwidth]{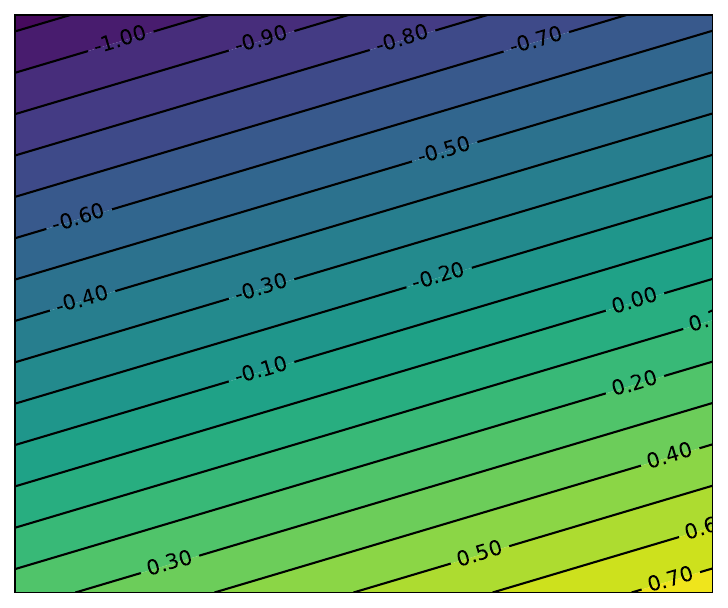}}
\scalebox{0.9}[1.0]{\includegraphics[width=0.31\textwidth]{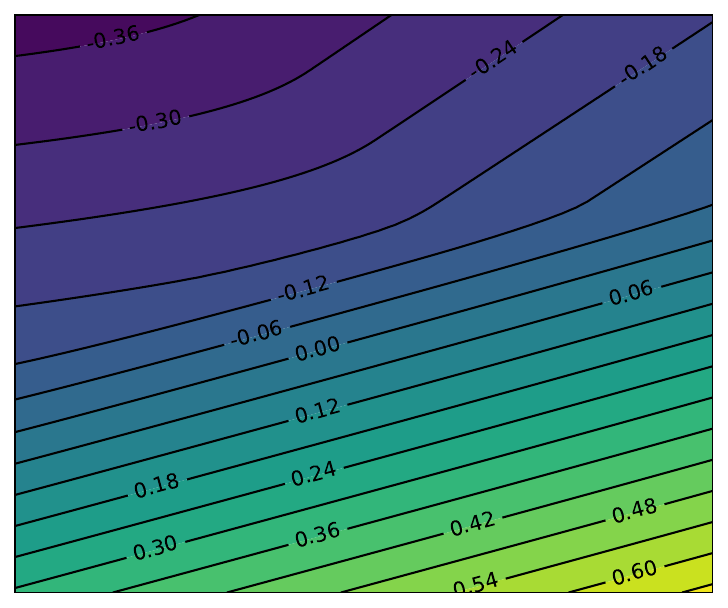}}
\scalebox{0.9}[1.0]{\includegraphics[width=0.31\textwidth]{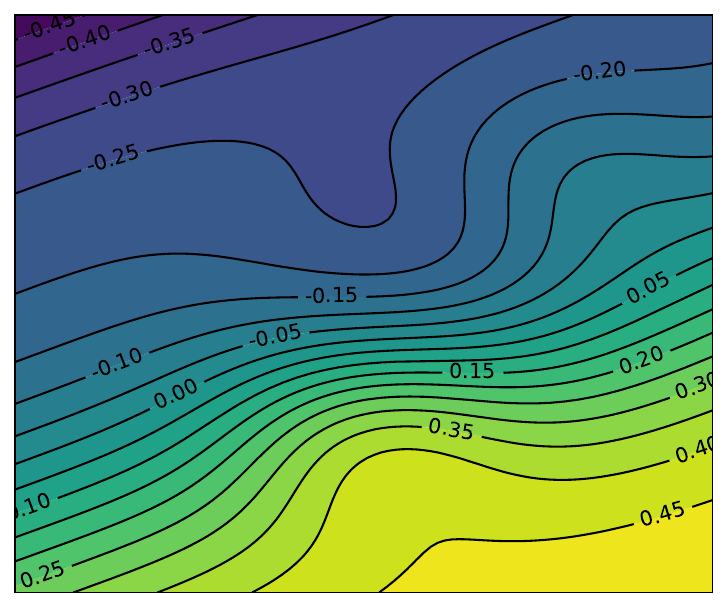}}

\scalebox{0.9}[1.0]{\includegraphics[width=0.31\textwidth]{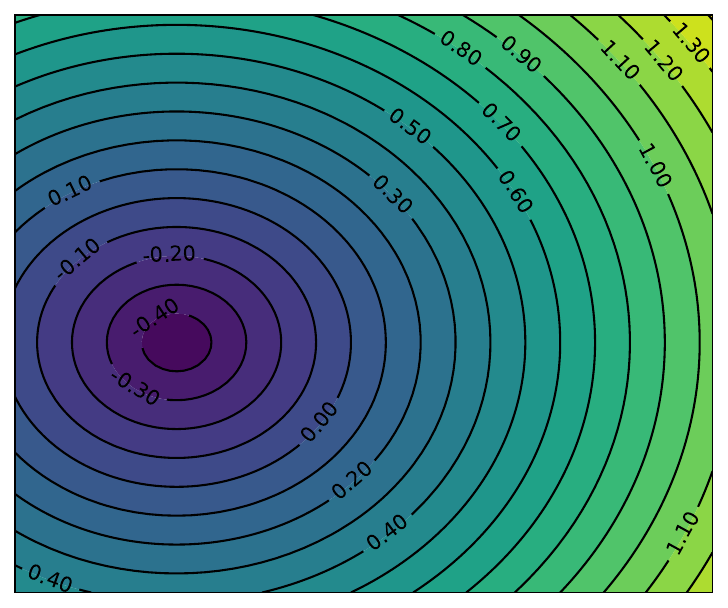}}
\scalebox{0.9}[1.0]{\includegraphics[width=0.31\textwidth]{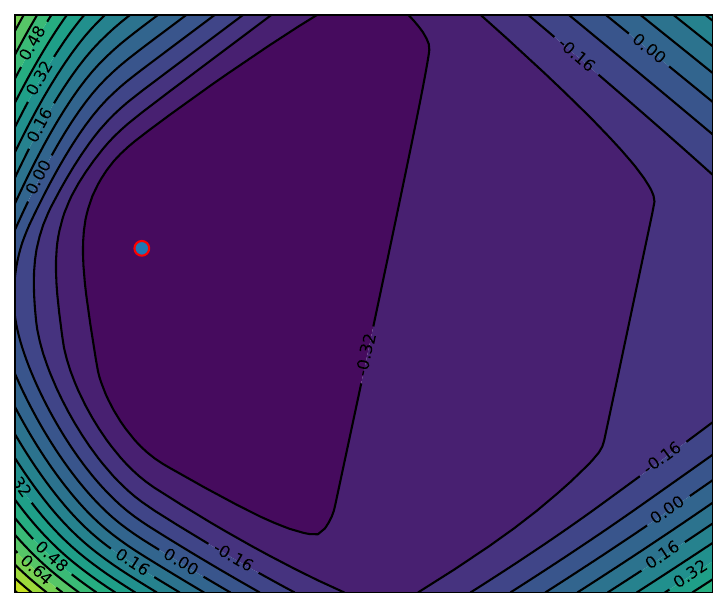}}
\scalebox{0.9}[1.0]{\includegraphics[width=0.31\textwidth]{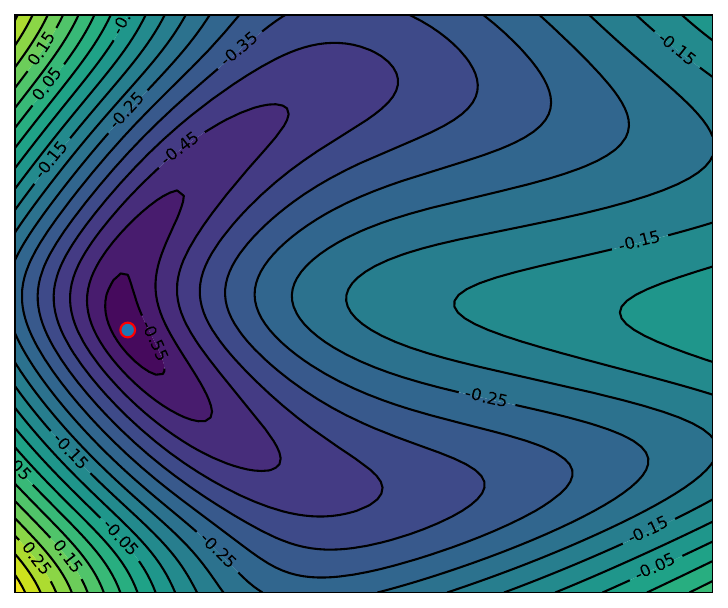}}

  \caption{ 
  We compare bounded functions with their unbounded counterparts (\textit{LEFT}) Uniform functions (\textit{MID}) convex function and (\textit{RIGHT}) invex function. For measure of distance, we are concerned with bounded functions (\textit{BOT}) with minimum at a point. Here, we can use function with bounded contour sets as measure of distance from the minima ($x^*$). To use the function itself as norm, we need to shift the minima to origin and the output of origin to zero.
  }
  \label{fig:contour_plots_local_vs_global}
\end{figure*}

In table~\ref{tab:metrics_in_fmnist}, we compare these learnable function as distances in the form: $y = f_{metric}(\vx - \vw) + b$, where $f_{metric}$ is itself specific type of neural network (depicted by Figure~\ref{fig:contour_plots_local_vs_global}) with similar number of parameters for comparison. We use same settings as Section~\ref{sec:metric_layer}, replacing \textit{Layer1} with generalized distance. The results show that these notion of learnable generalized distances work. Moreover, we may credit the low accuracy of ordinary neural network to it not being a generalized form of metric; we need more experiments to rule out other factors like regularization.

%% file: 07_on_embeddings_activation_viz.tex
\subsection{Activation visualization on Embedding Space}

\begin{figure*}[h]
  \centering

\includegraphics[width=0.34\textwidth]{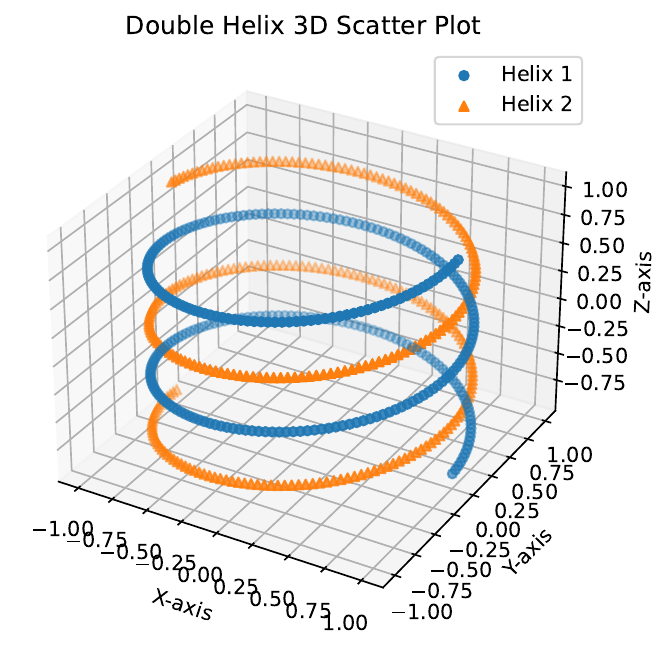}
\includegraphics[width=0.38\textwidth]{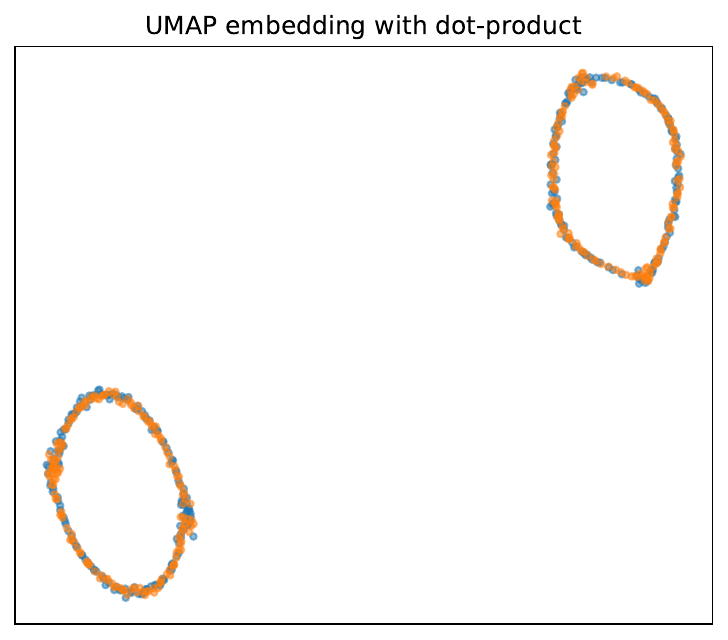}
\includegraphics[width=0.38\textwidth]{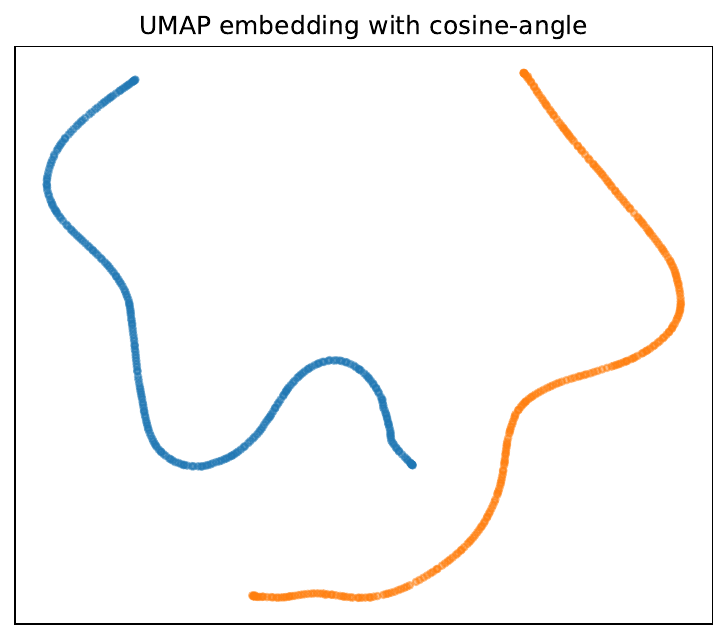}
\includegraphics[width=0.38\textwidth]{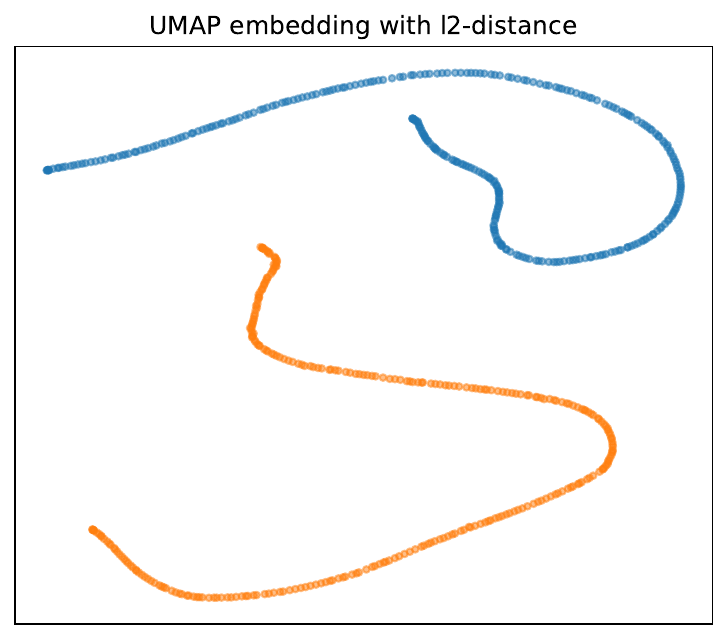}

\caption{
Mapping of double hellix using different measure of distance.
}
  \label{fig:double_helix_embedding}
\end{figure*}

\textbf{Case Study on Dimensionality Reduction of Double-Helix:}
We compare UMAP~\cite{mcinnes2018umap} embeddings based on various metrics: angle and $l^2$-distance as compared to dot-product. Dot-product produces poor seperation in edge case of double-hellix as shown by Figure~\ref{fig:double_helix_embedding}. We believe that it has to be a proper metric to produce good embedding. Dot-product on the other hand, does not follow the properties of metrics, as well as have both magnitude and angular component to it, making it difficult to disentangle the measurement. 

\begin{figure*}[h]
  \centering

\includegraphics[width=0.45\textwidth]{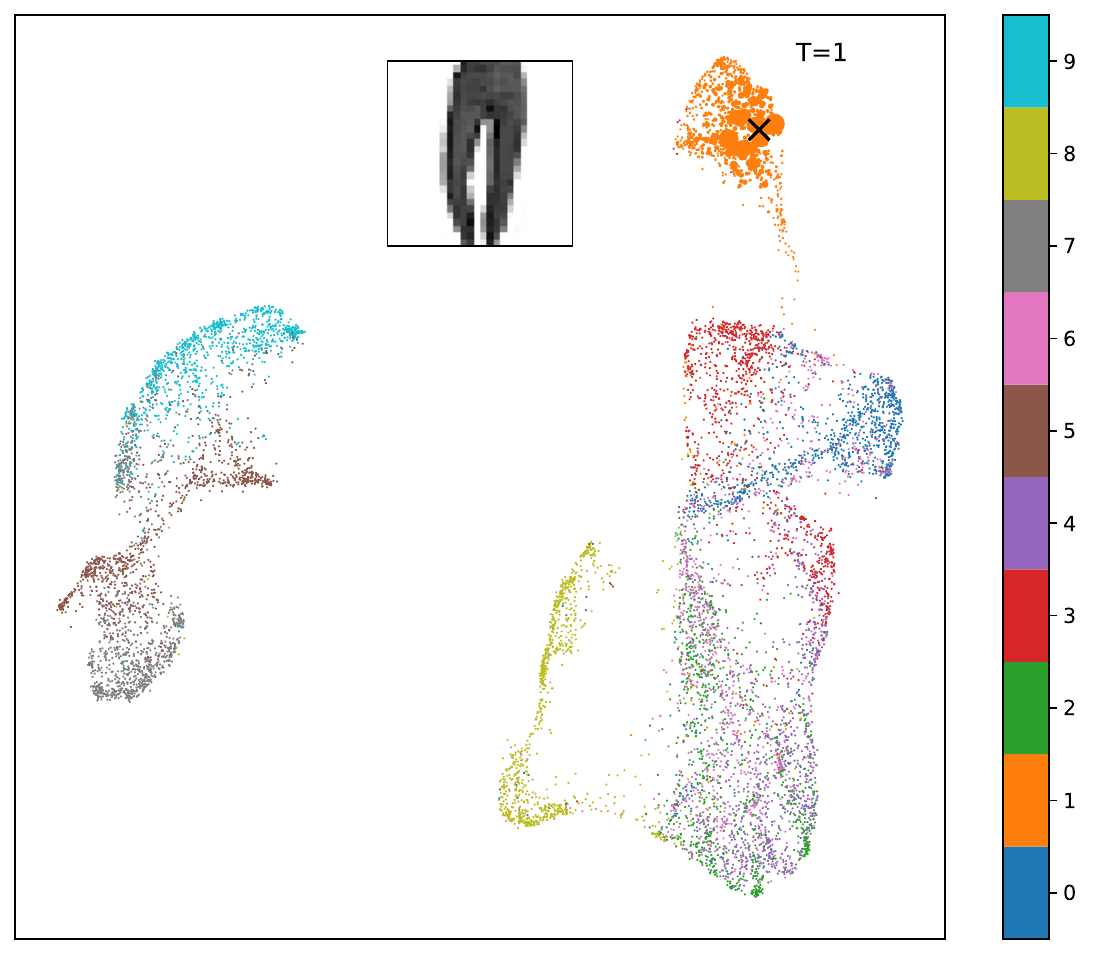}
\includegraphics[width=0.45\textwidth]{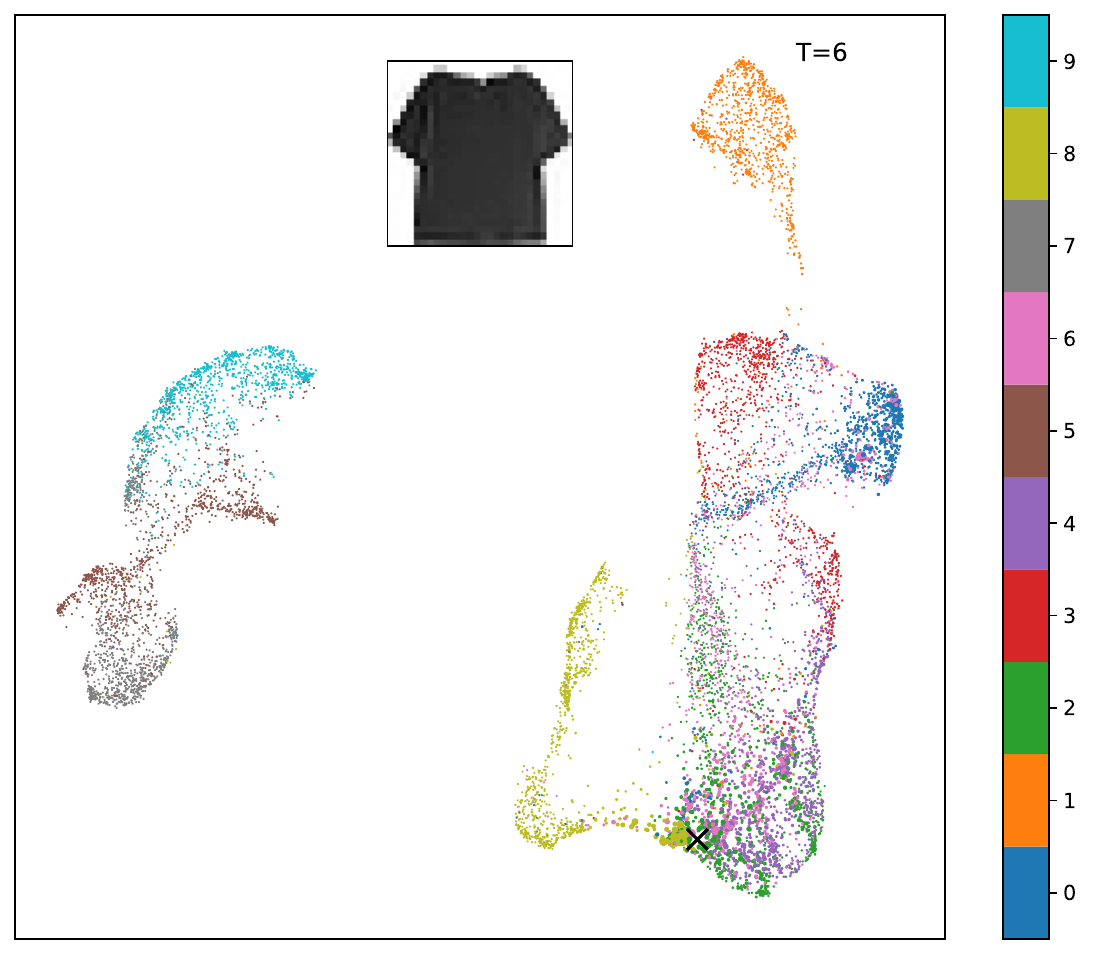}

\includegraphics[width=0.45\textwidth]{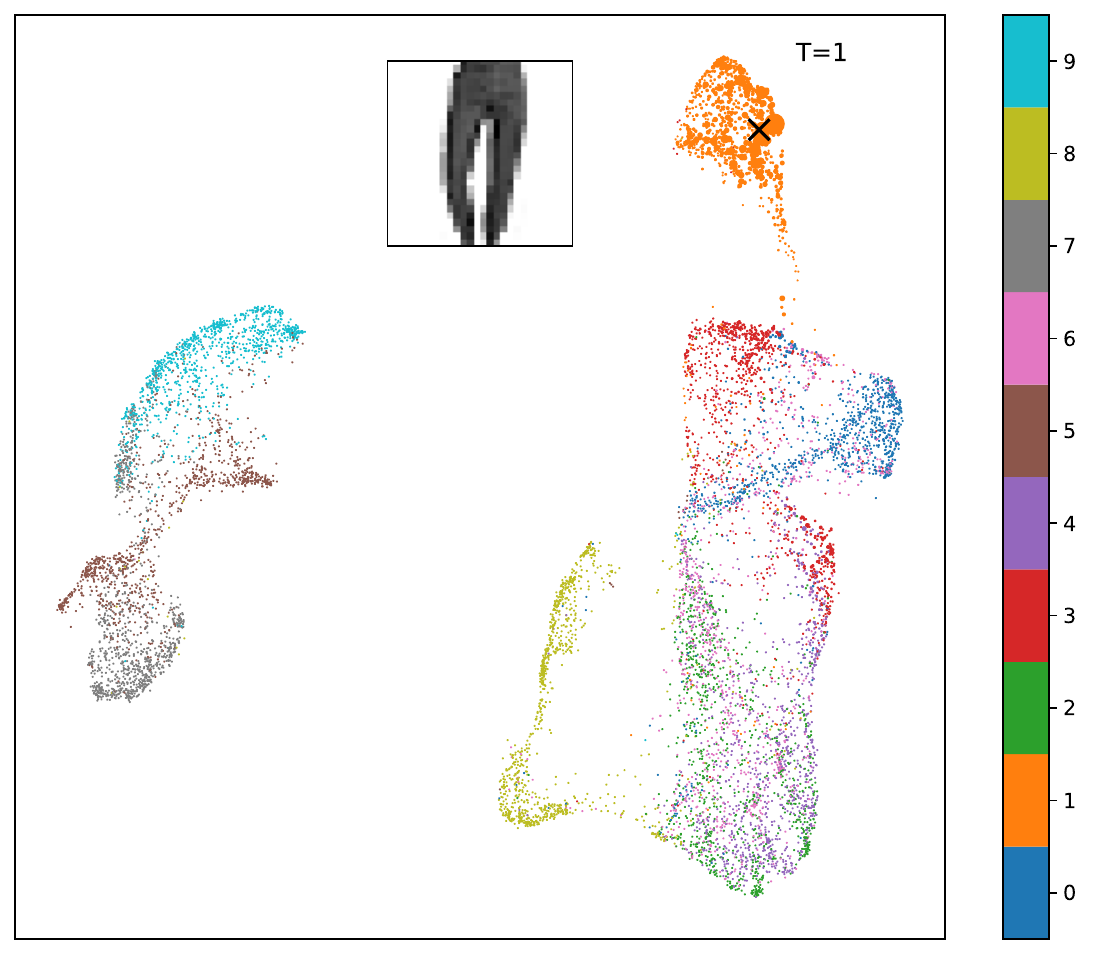}
\includegraphics[width=0.45\textwidth]{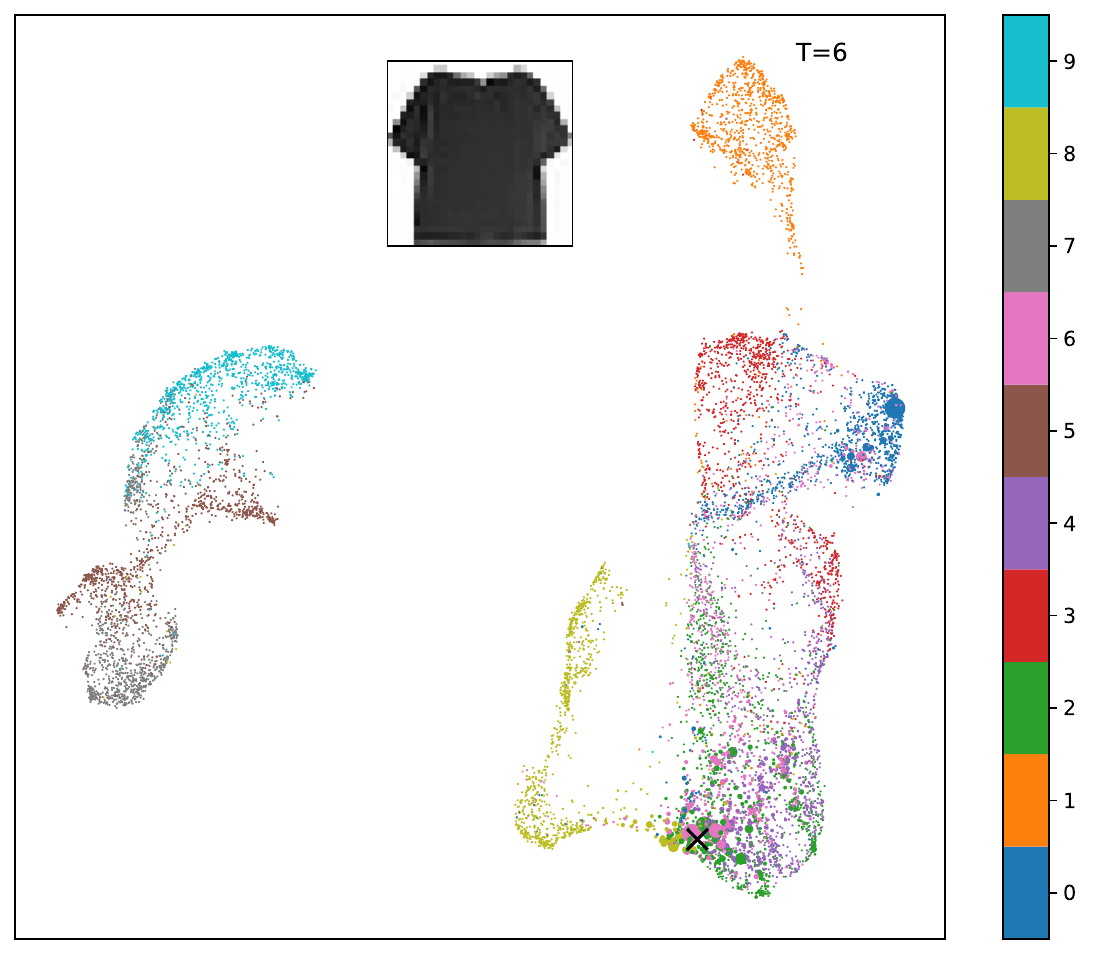}

\includegraphics[width=0.45\textwidth]{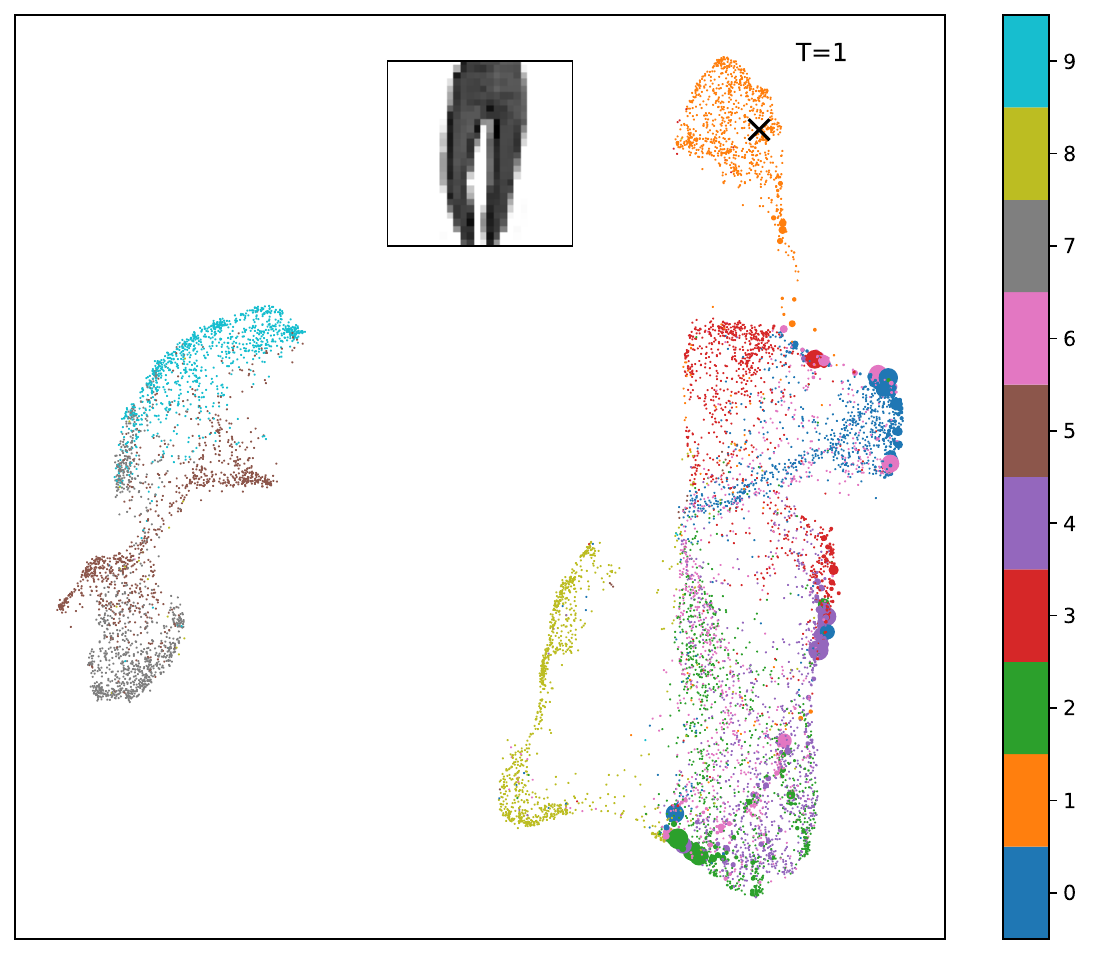}
\includegraphics[width=0.45\textwidth]{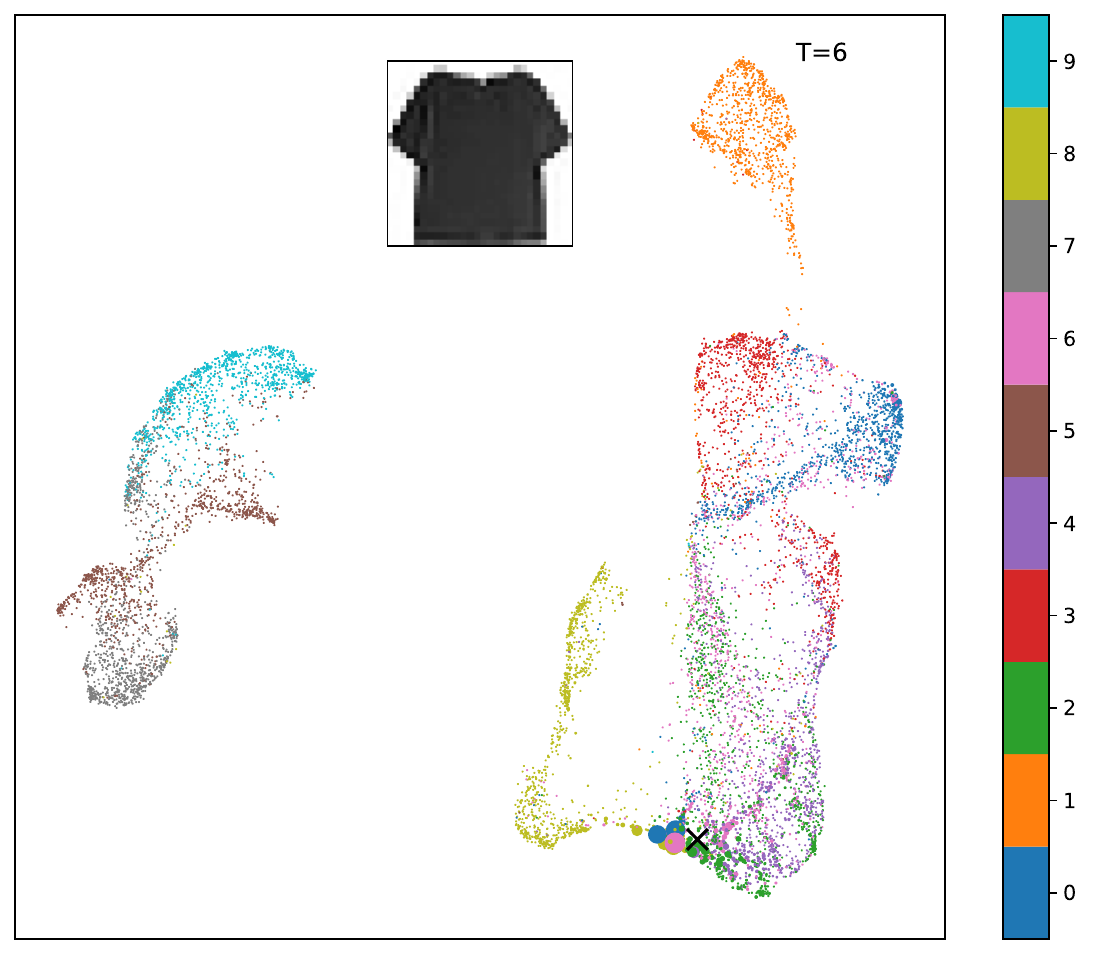}

  \caption{
  Data Interpretation using local activation visualization. Points represent the UMAP of weights/center and the size represents similarity to the test sample ($\times$), $T$ represents the target class of the sample. \textit{(TOP)} using l2-norm \textit{(MID)} using cosine angle \textit{(BOT)} using dot-product.
  We choose the plot for when dot-product does poor(left) and when distance does poor(right) among 32 activation samples. \textcolor{magenta}{Zoom in for details.}
  }
  \label{fig:viz_activation_in_embedding}
\end{figure*}

\clearpage

\textbf{Interpretation of High Dimensional Data:} UMAP~\cite{mcinnes2018umap} has been used widely to visualize high dimensional dataset. However, it projects input to low dimension with not enough information to infer about the inputs from embeddings only. To increase information about input, we add extra dimension to the embedding points with local activation as shown in Figure~\ref{fig:viz_activation_in_embedding}. Here, we use negative exponential to get local similarity measure, a continuous version of k-nearest neighbour as used on UMAP. The mapping of centers in low dimension along with magnitude of activation allows us to interpret high-dimensional input space in embedding space. The dimension of local activation is higher than the embedding space, providing more information to infer about input. 
The activations are always local for metrics, i.e. angle and $l^2$-norm.
For dot product activation, the embedding activations are not always local suggesting non-local activations on the high dimension itself.

%% file: 08_invertibility_of_metrics.tex
\subsection{Reversibility of Metrics as Transform}
\label{sec:reverse_metrics}

Metrics are reversible, i.e., we can reconstruct inputs given metrics from a set of points. Generally, N+1 metric values from unique centers intersect at unique point. There exists certain values of measures/distances, that do not intersect uniquely, such as scaled down $l^2-$distances, or when centers are collinear. 

\textbf{Invertibility of Linear Transform:}
Linear Transforms $f:\mathbf{R}^M\to \mathbf{R}^N$, where $N \ge M$, can be inverted using Moore–Penrose inverse (or Pseudoinverse). Scientific computing libraries provide functions for this operation. The bias can be simply subtracted during the inversion.
$$\vy = \mA \vx + \vb$$
This can be inverted as follows, where $A^+$ is the pseudoinverse.
$$\vx = \mA^+ (\vy - \vb)$$

\textbf{Invertibility of ``Inverse Stereographic Transform'':}
We use inverse stereographic transform ($i-$stereo) to project nD space to (n+1)D space with nD sphere as data manifold. This transform is reversible, called as stereographic transform~\cite{stereographic_projection}. Moreover, parametric linear transform or cosine angle on top of $i-$stereo is also reversible. Here, angle($i-$stereo($\vx$), $\vw$) is a metric on sphere but not on euclidean space of $\vx$.

\textbf{Invertibility of Angles:}
When we compute cosine angles, we need to normalize the vectors to unit magnitude of 1, and hence, the magnitude information is lost. Since, the cosine angles are the linear transform with unit vectors, the unit vectors are invertible. We can use extra angle, not from the origin but from extra known point to be able to reconstruct the magnitude information.

\begin{figure*}[h]
  \centering

\frame{\includegraphics[width=0.335\linewidth]{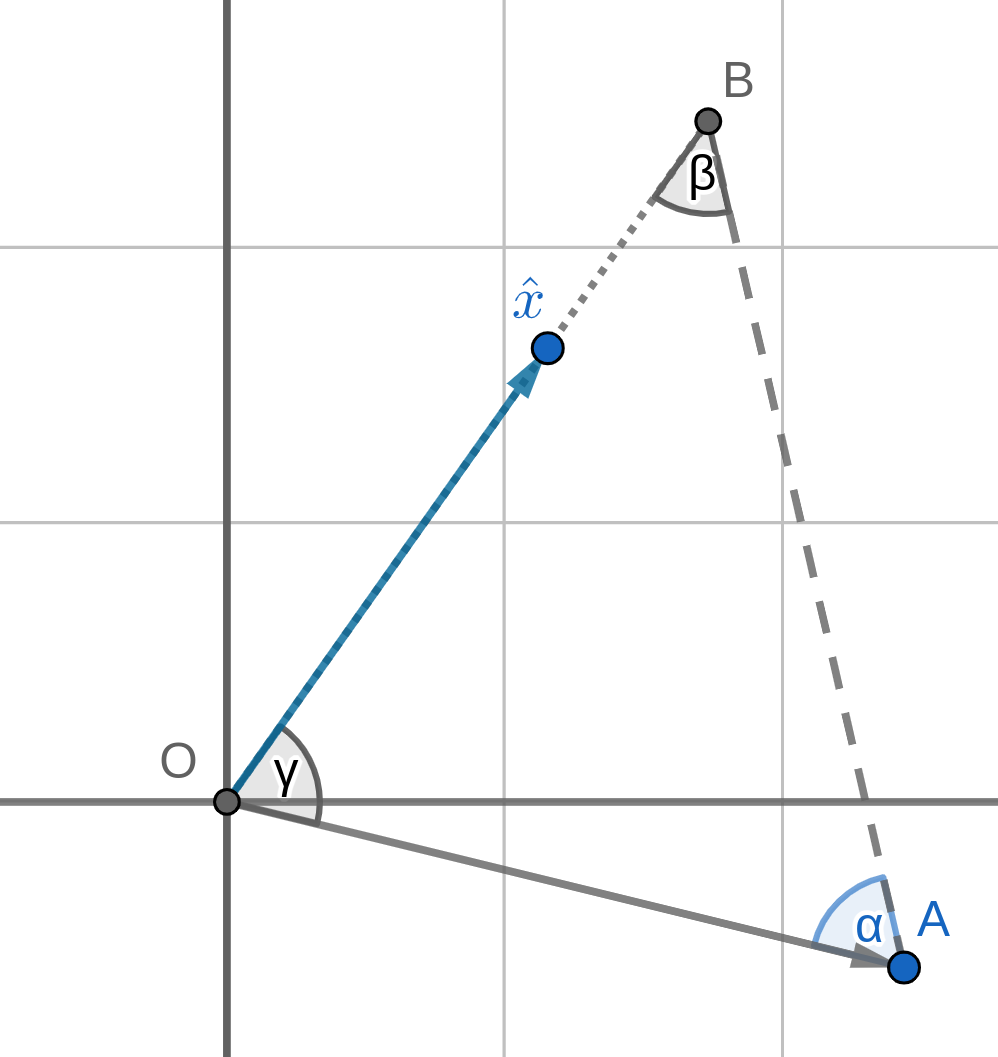}}
\frame{\includegraphics[width=0.36\linewidth]{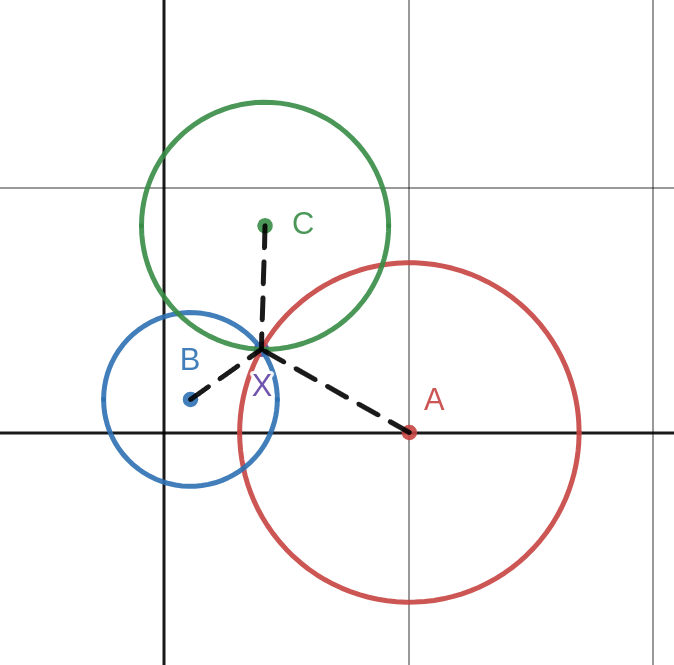} }
\caption{Inversion of input from angles (left) and distances(right).}
\label{fig:angle_inversion}
\end{figure*}

From the figure~\ref{fig:angle_inversion}, we try to derive the inverse ($\vx$, or length $OB$), with known $A$, $\alpha$, $\hat{\vx}$. Here, $\hat{\vx}$ is calculated by pseudo-inverse of cosine-angles with the weights.  

Here, we can calculate angles: $\gamma = \cos^{-1}(\hat{\vx} \cdot \va / \|a\| )$ and $\beta = \pi - \gamma - \alpha$.
Now, using triangle law,
$$\frac{OB}{\sin \alpha} = \frac{OA}{\sin \beta}$$
$$OB = \frac{OA \sin \alpha}{\sin \beta}$$
Hence, with the magnitude $OB$ known, reconstruction $\vx = OB \hat{\vx}$.

We consider this to be special case of triangulation, where N angles are calculated from input-origin-weight and, 1 extra angle calculated from origin-weight-input. 

\textbf{Invertibility of Euclidean Distance:}
Inverting a measure of euclidean distance from centers is inspired by the Global Positioning System (GPS)~\cite{bancroft1985algebraic, norrdine2012algebraic}. These system use distance from 4 known centers to locate a point in 3D space. We find euclidean distance can be inverted in any dimensions. If the dimension of point is $N$, then we need $N+1$ distances from known centers to be able to reconstruct it. This could be done by solving the equation of n-Sphere for intersection.

\texttt{Proof:}
Consider that we have $N+1$ centers for $N$ Dimensional space. Let the centers be $\mathbf{x_1}, \mathbf{x_2}, \cdots , \mathbf{x_{(N+1)}}$ where a single point $\mathbf{x_i}$ is a coordinate of form $(\mathbf{x_i}_{;1}, \mathbf{x_i}_{;2}, \cdots, \mathbf{x_i}_{;N} )$. The corresponding $l^2$-distance ($d$) from an input point $\mathbf{x}$ (without subscript) for $i^{th}$ center is given as:
$${d_i}^2 = \sum_{j=1}^N (\mathbf{x}_{;j} - \mathbf{x_i}_{;j})$$
$${d_i}^2 = \sum_{j=1}^N (\mathbf{x}_{;j}^2 - 2\mathbf{x}_{;j}\mathbf{x_i}_{;j} + \mathbf{x_i}_{;j}^2)$$
Here, $;j$ corresponds to index of coordinate dimension. Since we have N+1 equations to solve for N values of $\mathbf{x}$, we equate ${d_1}^2 - {d_2}^2$ as follows:
$${d_1}^2 - {d_2}^2 = \sum_{j=1}^N 2\mathbf{x}_{;j}(\mathbf{x_2}_{;j} - \mathbf{x_1}_{;j})+ \mathbf{x_1}_{;j}^2 - \mathbf{x_2}_{;j}^2$$
We can construct $N$ such linear equations: ${d_i}^2 - {d_{i+1}}^2$ for $N+1$ number of centers.

For simplicity, we choose 2-dimensional input space with 3 distances. The equation can be written in matrix form as:
$$
\begin{bmatrix}
2(\mathbf{x_2}_{;1} - \mathbf{x_1}_{;1}) & 2(\mathbf{x_2}_{;2} - \mathbf{x_1}_{;2})\\
2(\mathbf{x_3}_{;1} - \mathbf{x_2}_{;1}) & 2(\mathbf{x_3}_{;2} - \mathbf{x_2}_{;2}) 
\end{bmatrix}
\cdot
\begin{bmatrix}
\mathbf{x}_{;1}\\
\mathbf{x}_{;2}
\end{bmatrix}
=
\begin{bmatrix}
{d_1}^2 - {d_2}^2\\
{d_2}^2 - {d_3}^2
\end{bmatrix}
-
\begin{bmatrix}
(\mathbf{x_1}_{;1}^2 - \mathbf{x_2}_{;1}^2) + (\mathbf{x_1}_{;2}^2 - \mathbf{x_2}_{;2}^2)\\
(\mathbf{x_2}_{;1}^2 - \mathbf{x_3}_{;1}^2) + (\mathbf{x_2}_{;2}^2 - \mathbf{x_3}_{;2}^2)
\end{bmatrix}
$$
To find the inversion in x-space, we need to solve for $\mathbf{x} = \begin{bmatrix}
\mathbf{x}_{;1}\\
\mathbf{x}_{;2}
\end{bmatrix}$ in above equation.

Similarly, we can solve the above equation for any dimensions. We provide the general implementation for the inversion of euclidean distances in Algorithm~\ref{algo:euclidean_inversion}. If the exact distance is not known, then scaled distances from N+2 points is needed for reconstruction, we find this with Gradient Descent based input reconstruction.

\begin{algorithm*}[h]
\caption{Multi-lateration: Euclidean Distance Inversion}
\label{algo:euclidean_inversion}

\begin{minted}[fontsize=\footnotesize,baselinestretch=1]{python}
def inverse_euclidean(C, d):
    ''' To find -> x: shape [b, N] -> given C: shape [N+1, N], d:[b, N+1]''' 
    A = 2*(C[1:]-C[:-1])
    c2 = C**2
    Z = (c2[:-1]-c2[1:]).sum(dim=1, keepdim=True)
    invA = torch.pinverse(A)
    
    d2 = d**2
    D = d2[:, :-1]-d2[:, 1:]
    
    x = torch.matmul(invA, D.t()-Z).t()
    return x
\end{minted}
\end{algorithm*}

%% file: 09_invertibility_of_generalized_distances.tex
\subsection{On Reversibility of Generalized Metrics as Transform}
\label{sec:reverse_generalized_metrics}

Generalized distances such as convex and invex function does not have unique contour sets. The intersection of contours of different distances may intersect at non-unique points, i.e. the intersection of different contours of general convex function may not be unique. 
\begin{figure}[h]
  \centering

\includegraphics[width=0.31\textwidth]{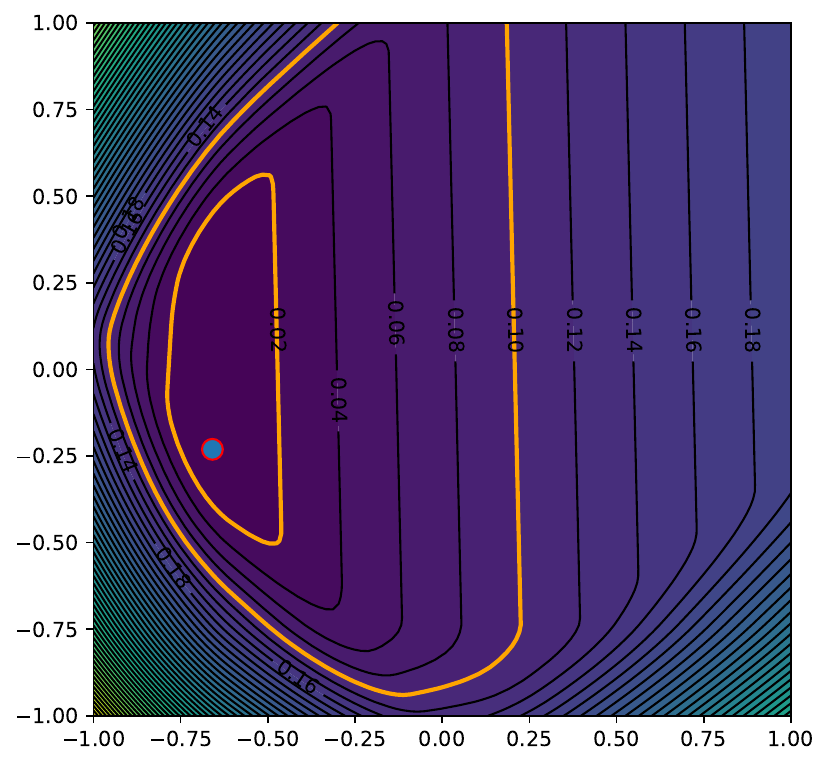}
\includegraphics[width=0.31\textwidth]{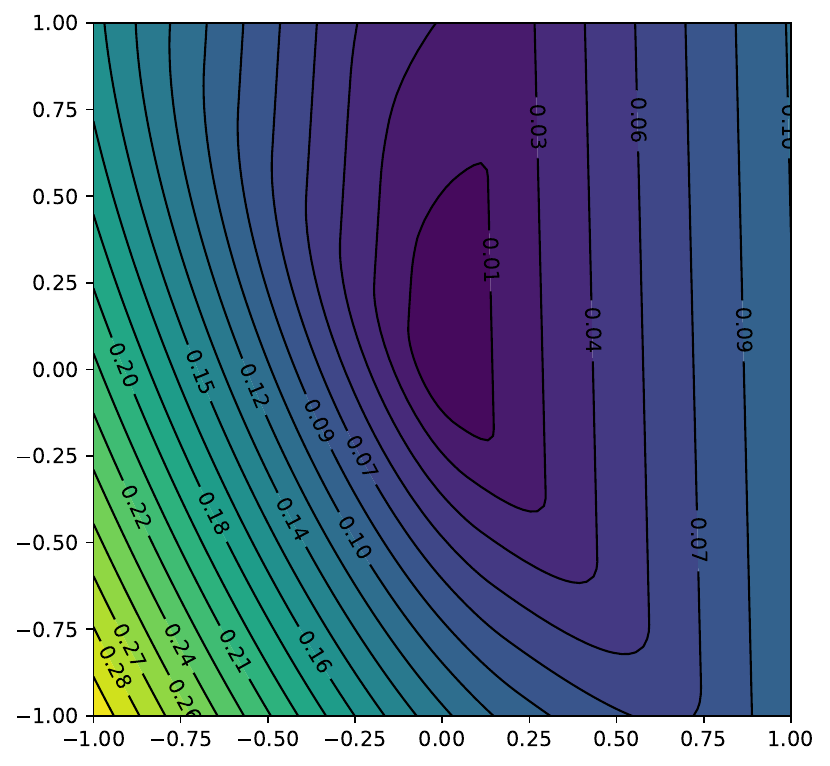}
\includegraphics[width=0.31\textwidth]{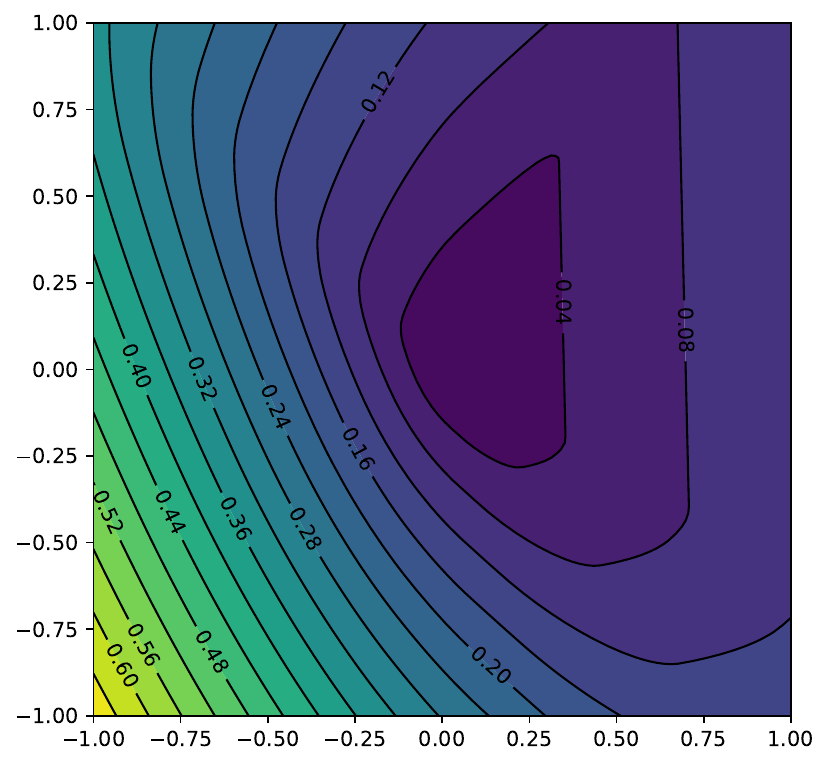}

\includegraphics[width=0.31\textwidth]{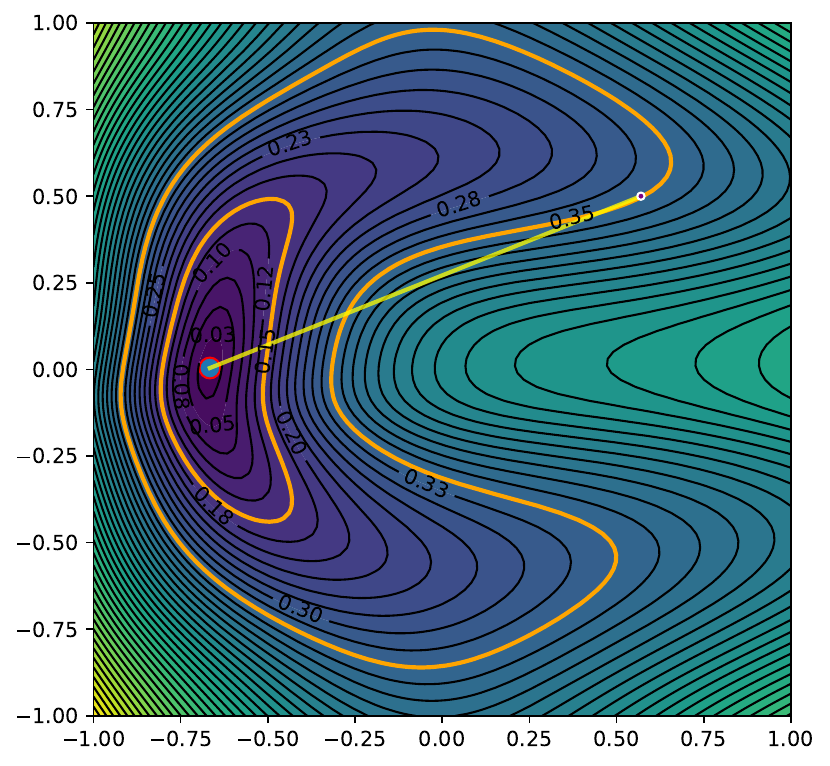}
\includegraphics[width=0.31\textwidth]{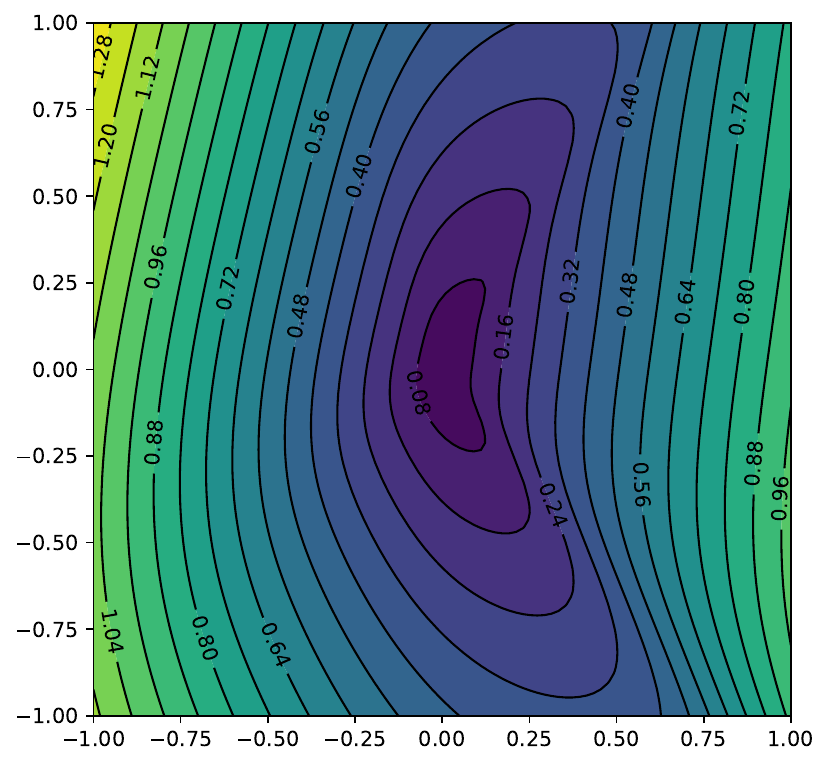}
\includegraphics[width=0.31\textwidth]{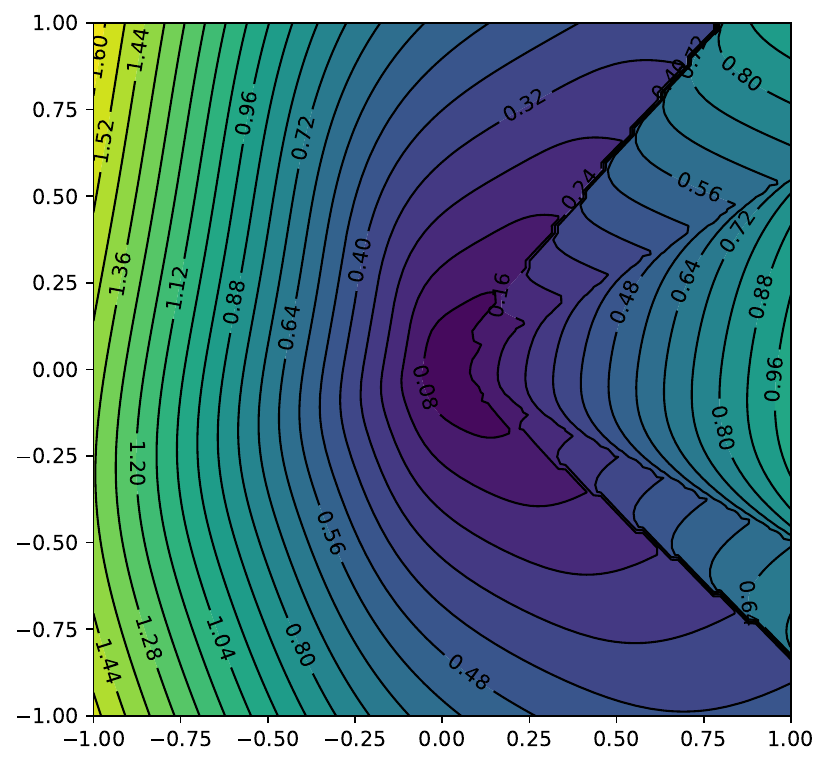}

\caption{
Contour Distance of \textbf{(TOP)} Convex and \textbf{(BOT)} Invex function. \textit{LEFT}: General convex and invex functions with respective contour sets to create distance from, \textit{MID}: contour distance from smaller set and \textit{RIGHT}: contour distance from larger set. Here, invex contour distance does not create proper function as two contours from same set intersect with a straight line from center.
}
  \label{fig:generalized_contour_distance}
\end{figure}

\begin{figure}[h]
  \centering

\frame{\includegraphics[width=0.32\textwidth, trim={3cm 1cm 12cm 1cm},clip]{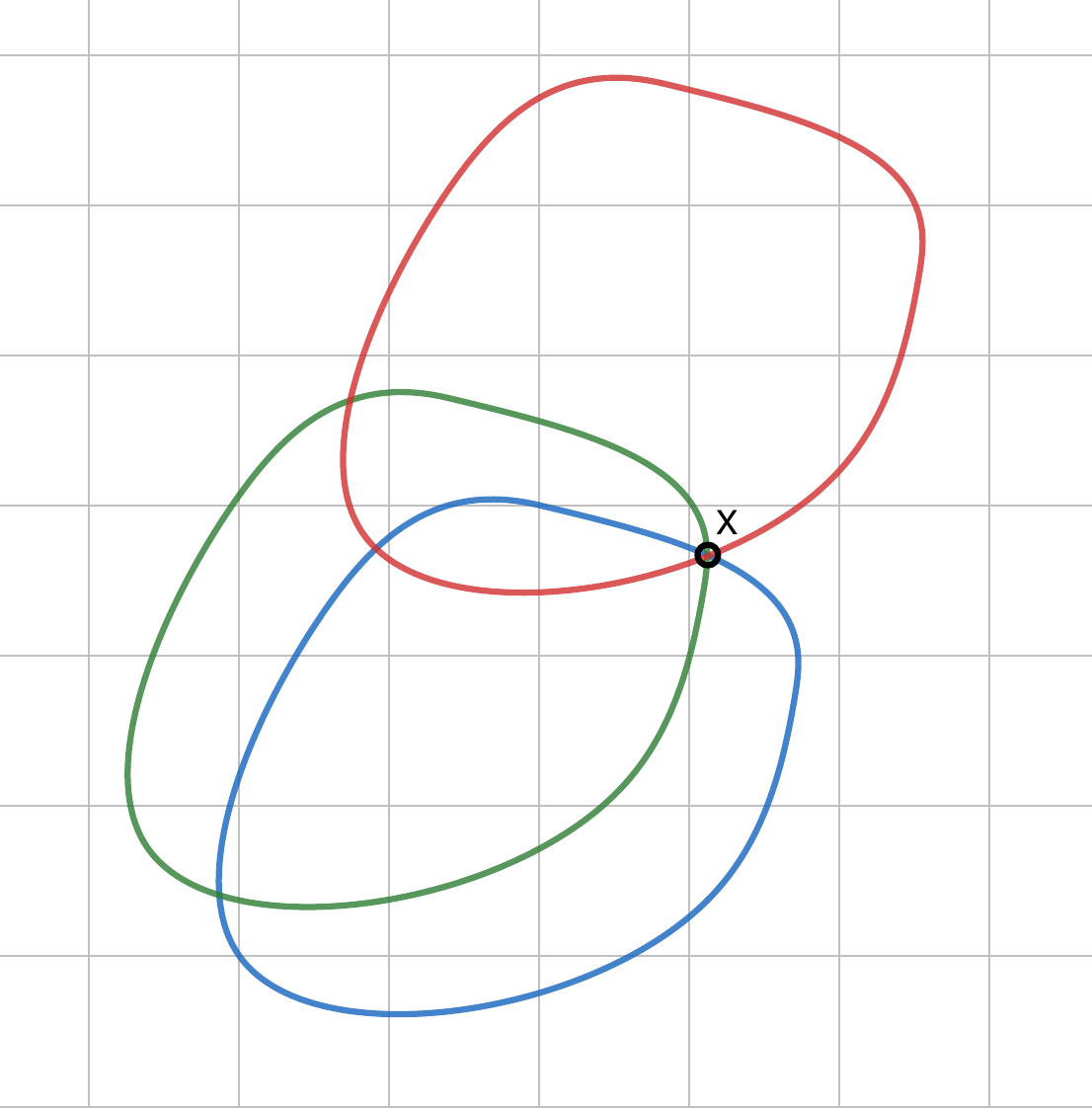}}
\frame{\includegraphics[width=0.355\textwidth, trim={1cm 1cm 12cm 1cm},clip]{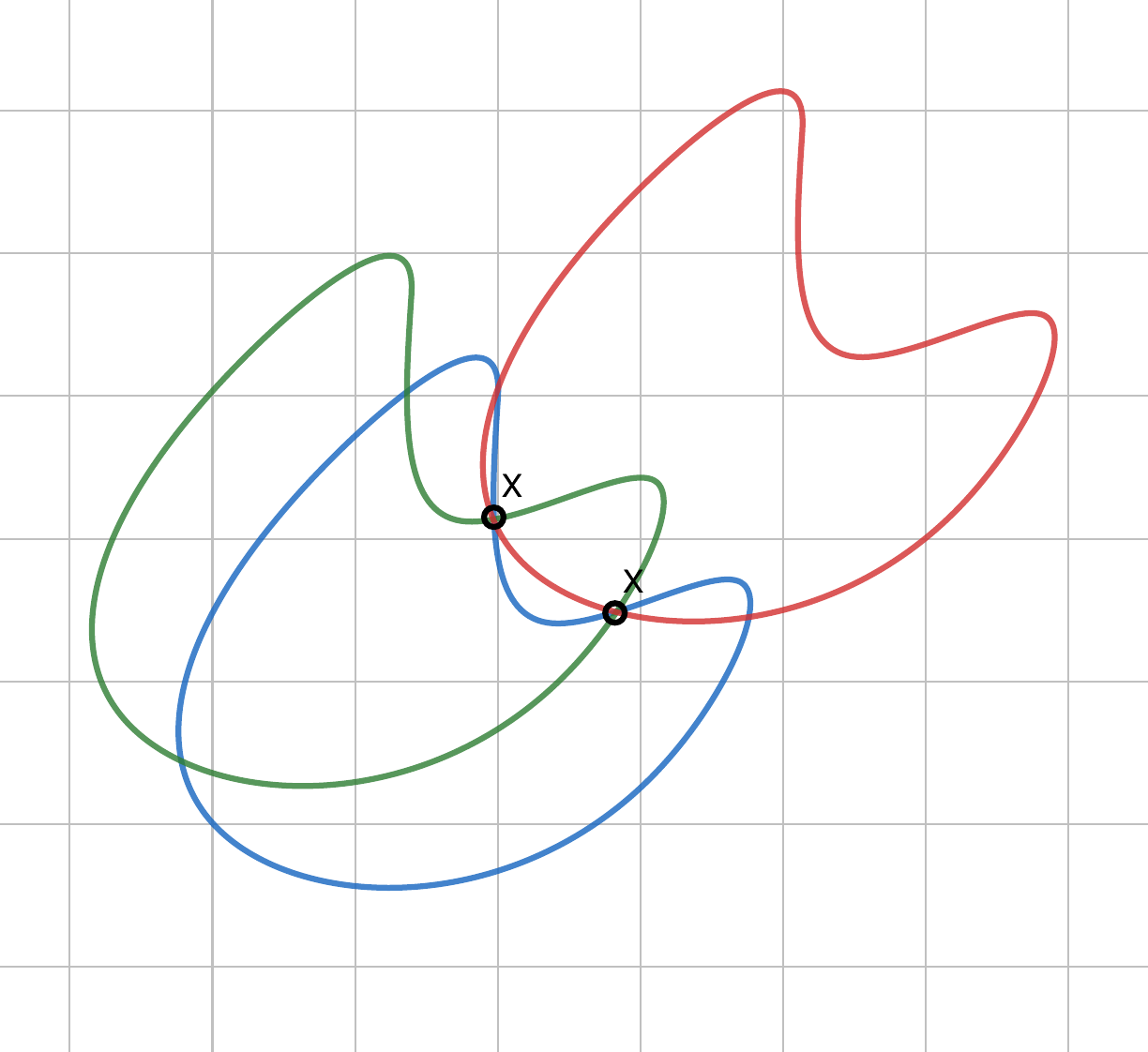}}

\caption{
Intersection of convex-sets(\textit{LEFT}) and invex-sets(\textit{RIGHT}).
}
  \label{fig:intersection_of_contour_sets}
\end{figure}

\textbf{Invertibility of Convex Contour Distance:}
We specialize the convex function to have same contour sets, which we call convex contour distance, i.e. the shape of the convex contour does not change~\cite{ma2000bisectors} as shown in Figure~\ref{fig:generalized_contour_distance}. Such convex distance function have unique solution.

Similar to euclidean distance, convex distance produce 1-1 function. Even euclidean distance produce 1-1 function, however, any possible set of distances might not even intersect, and not have any input associated with it. Hence, euclidean along with convex contour distance produce 1-1 into function as shown in Figure~\ref{fig:intersection_of_contour_sets}, excluding the case when centers are collinear. 

\textbf{Reversibility of 1-Invex Contour Distance:}
Moreover, we can specialize the invex function as well to have same contour sets called invex contour distance. We find that if the contours can be joined with straight line from the center without intersection, then it has sensible contour sets (Figure~\ref{fig:generalized_contour_distance} MID) which we call 1-Invex contour distance. If contours intersect with straight line from centers, the contour distance of such sets intersect with each other (Figure~\ref{fig:generalized_contour_distance} RIGHT), it is unsuitable to use as general distance. 

1-Invex contour distance can intersect at multiple points. Hence, the function produced is many-1 into as shown in Figure~\ref{fig:intersection_of_contour_sets}.

%% file: 10_properties_of_metrics.tex
\subsection{Properties of Metrics}
\label{sec:metric_properties}

\textbf{Metrics:} A metric space $(\mathbf{M}, d)$ is a set of input-space ($\mathbf{M}$) and distance function $d(\vx, \vy)$ between any two elements in the space. The distance is measured by a function called a metric or distance function and satisfies following axioms for all $\vx, \vy, \vz \in \mathbf{M}$~\cite{metric-space}.  
\begin{enumerate}
  \item The distance from a point to itself is zero: $$d(\vx, \vx) = 0$$
  \item (Positivity) The distance between two distinct points is always positive: $$\text{If } \vx \neq \vy \text{ , then } d(\vx, \vy) > 0$$
  \item (Symmetry) The distance from $x$ to $y$ is always the same as the distance from $y$ to $x$: $$d(\vx, \vy) = d(\vy, \vx)$$
  \item The triangle inequality holds: $$d(\vx, \vz) \leq d(\vx, \vy) + d(\vy, \vz)$$
\end{enumerate}

Such metrics include $l^p$ norm induced metrics and angular distance. However, we can generalize metrics by relaxing axioms $1-4$ with different types of measures.

\textbf{Angular Distance:} Angle($\theta$) with respect to origin is a metric for inputs on nSphere manifold. The angle can be modified like in Section~\ref{sec:reverse_metrics} to behave as metric in euclidean space. Although $1-\cos{\theta}$ behaves like a distance and is invertible, it does not follow triangle inequality~\cite{cosine-sim}.

\begin{figure}[h]
  \centering

\includegraphics[width=0.32\textwidth]{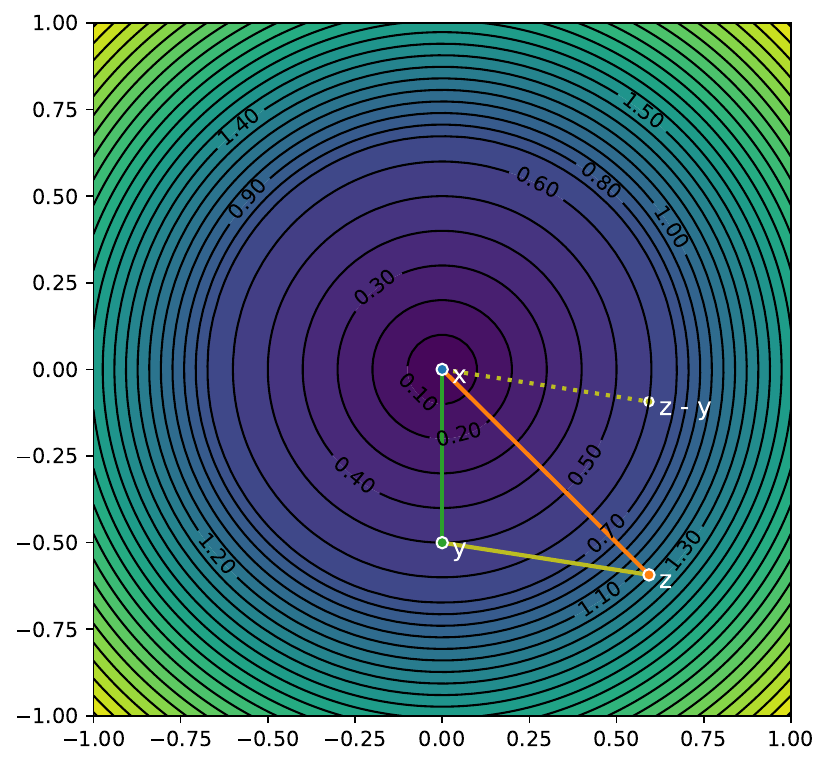}
\includegraphics[width=0.32\textwidth]{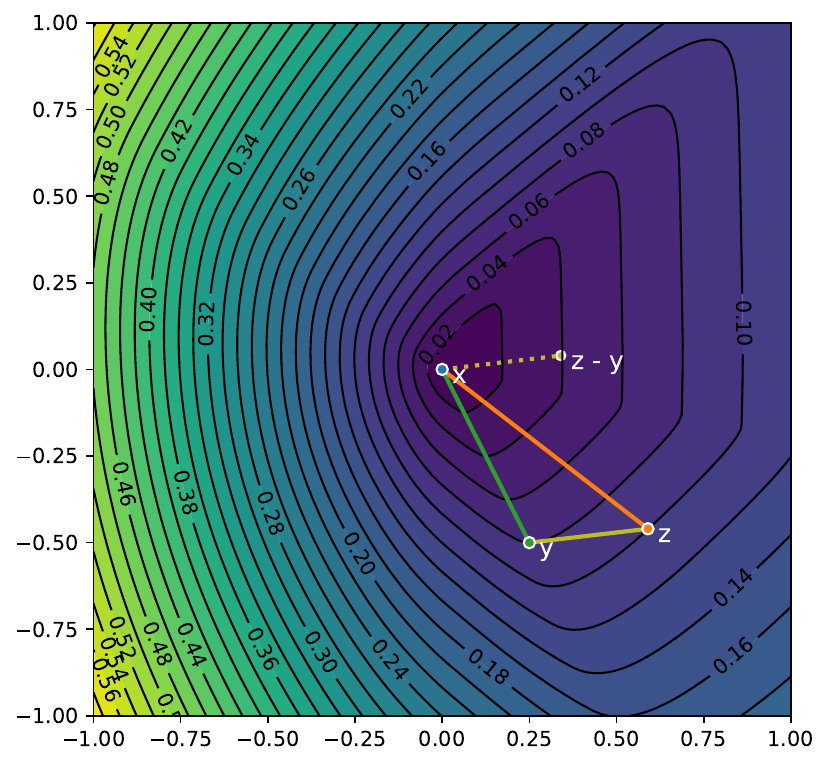}
\includegraphics[width=0.32\textwidth]{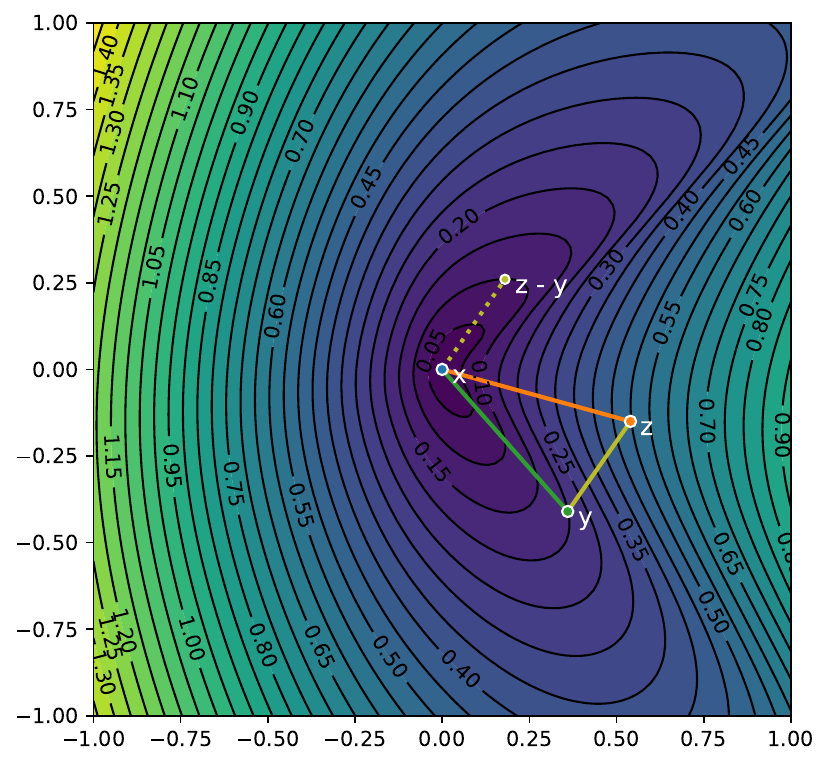}

  \caption{
  Triangular inequality in various types of metrics. We test if $d(\vx, \vz) \leq d(\vx, \vy) + d(\vy, \vz)$ is $true$ or $false$  \textit{(LEFT)}: Modified $l^2-$norm; $1.2 \leq 0.5 + 0.6$ is $false$, \textit{(MID)}: Convex contour distance; $0.1 \leq 0.08 + 0.04$ is $true$ and \textit{(RIGHT)}: 1-Invex contour distance; $0.5 \leq 0.2 + 0.1$ is $false$.
  }
  \label{fig:triangle_inequality_metrics}
\end{figure}

\textbf{Modified $l^2$-distance:} Consider a convex piecewise function $f(x) = max(x, s*(x-b)+b)$ where, $s>1$ is the slope of a piece and $b$ is the location of intersection of the pieces. Then the overall distance function given by $f(l^2(x, y))$ is not a proper metric. It follows axiom $1,2,3$ except the triangle inequality (shown in Figure~\ref{fig:triangle_inequality_metrics} LEFT).
This generalization follows properties of Semimetrics.

\textbf{General convex function:} A general convex function like portrayed in Figure~\ref{fig:generalized_contour_distance} does not have symmetric distance (axiom-$3$), and similar to previous convex piecewise modification, it also does not follow triangle inequality. It only follows axiom $1$ and $2$.
This generalization follows properties of Premetrics.

\textbf{General invex function:} A general invex function also follows the same axioms of a general convex function.

\textbf{Convex contour distance:} A convex function with linear increase in distance from the minima is depicted by Figure~\ref{fig:triangle_inequality_metrics} MID. Here, the convex contour distance follows the triangle inequality. However, it does not follow symmetry property, following axiom $1,2,$ and $4$ only.  
This generalization follows properties of Quasimetrics.

\textbf{1-invex contour distance:} A 1-invex contour distance function with linear increase in distance from minima is dipicted by Figure~\ref{fig:triangle_inequality_metrics} RIGHT. Such generalized metrics do not follow symmetry and the triangle inequality, following only axiom $1$ and $2$. This generalization also follows the properties of Premetrics.

For example, the function $f(\vx, \vy) = 0.9+0.1\cos(2d)-e^{-d^2} \hspace{0.5em}; \hspace{0.5em} d = \|\vx-\vy\| \hspace{0.5em}$ is symmetric but does not follow triangle inequality. The function is Semimetric, despite not being always increasing function of distance. 

Overall, we find multiple generalizations of metrics: Quasimetrics, Premetrics and Semimetrics unique in their construction. The generalized metrics have very small constrain that many function with different properties satisfy the axioms.

%% file: 11_Dictionary_Learning_MLP.tex
\subsection{Dictionary Learning as 2 layered Neural Network}
\label{subsec:dictionary_learning}

One of the ways to create mapping of input to output can be done using Dictionary. 
$$\vy = f_{similarity}(\vq, \textbf{K}).\textbf{V}$$
$\textbf{K}$ey - The reference that stores the Value\\
$\textbf{V}$alue - The respective mapping value of some Key\\
$\vq$uery - The input ($\vx$) that is matched with Key to get the interpolated Value. 

We use $\vq$ and $\vx$ interchangeably. $f_{similarity}$ is in the range $[0, 1]$. We focus on $l^2$-norm based distance to measure similarity for experiments. 

\subsubsection{Un-Normalized Similarity}

To simplify our network interpretation, we have to normalize the activation to produce maximum value of 1 for neurons similar to input. This way, the value associated with neuron(key) can propagate as output for similar input(query). One way to compute similarity from distances is to use Gaussian function similar to RBF Network. However, the work on UMAP~\cite{mcinnes2018umap} shows that the distribution of distances shift to higher values and with less variance when the dimension of the $\mathbf{K}, \vq$(or $\textbf{Q}$) increases. The work normalizes the distances to be used as similarity as shown below.
$$f_{similarity}(\vx, \mathbf{K}, \tau) = \exp(\frac{-(\vd- \min(\vd))}{\tau\sqrt{\Var(\vd)}}) ; \hspace{2em} \vd = \{ f_{metric}(\vx, \vk_i) \}$$ for $i$ represents index of all the keys (or centers), $\tau$ represents temperature to change scale of exponential.

The second layer of MLP: $\vy = f_{similarity}(\vx, \mathbf{K}) \cdot \mathbf{V}$ is a linear map, corresponding to the value ($\vv_i$) when similarity of $i^{th}$ neuron is $1$. 
The initialization is done using input target pair, where, matrix $\mathbf{K}$ is initialized with input samples and $\mathbf{V}$ is initialized with the respective target values.

We can overfit a metric-transform based 2-layer neural network by memorizing M data points with M neurons, similar to $1$-Nearest Neighbour. However, this is not feasible for a large dataset. The smooth similarity helps to predict output robustly due to output being predicted by multiple neurons. We experiment with center initialization in $l^2-norm$ based ANN to some random training samples in Table~\ref{tab:random_data_as_neuron}. We find that we can gain significant accuracy in MNIST and F-MNIST dataset without even training. We can initialize 2 layered MLP with data and gain some test performance higher than random accuracy.

\begin{table}[h]
    \centering
    \caption{
    Initializing centers and labels with given $H=n$ hidden neurons which are initialized to random samples from the dataset. The accuracy is measured on test dataset with 20 different seeds.
    } 
    \resizebox{0.8\linewidth}{!}{
\begin{tabular}{@{}rlcccccc@{}}
\toprule 
Dataset & Acc & $H=10$ & $H=50$ & $H=200$  & $H=1000$ & $H=5000$ & $H=20000$ \\
\midrule
\multirow{3}*{MNIST} & mean & 36.48 & 60.76 & 76.40 & 85.61  & 88.56 & 89.43 \\
& std & 3.29 & 3.49 & 2.21 & 0.83  & 0.39 & 0.20 \\
& max & 42.69 & 67.86 & 81.0 & 87.03 & 89.57 & 89.79 \\
\midrule
\multirow{3}*{F-MNIST} & mean & 37.75 & 58.08 & 67.75 & 72.60  & 73.96 & 74.43 \\
& std & 6.63 & 3.47 & 1.84 & 1.43  & 0.45 & 0.21 \\
& max & 50.23 & 62.10 & 70.20 & 75.08 & 74.74 & 74.94 \\
\bottomrule
\end{tabular}
    }
    \label{tab:random_data_as_neuron}
\end{table}

Despite being sufficient for function approximation, this similarity has multiple problems that need to be solved for locally activating neuron. First, we find that this type of neuron may activate the highest when the $\vq$ and $\vk$ are not same, but if away from other $\textbf{K}$eys. This is not suitable if we want certainty that the similarity is maximum only when key and query match. Second, the activation of all the neurons, i.e. sum of all similarity is not equal to 1. This makes the problem of disentangling multiple neurons harder. 2 neurons might fire $\simeq 1$ for a given input. This problem can be observed in Figure~\ref{fig:epsilon-sm-neurons} as well.

\subsubsection{SoftMax (soft-arg-max) Normalized Similarity}

\begin{figure*}
     \begin{subfigure}{0.3\linewidth}
         \centering
         \includegraphics[width=\textwidth]{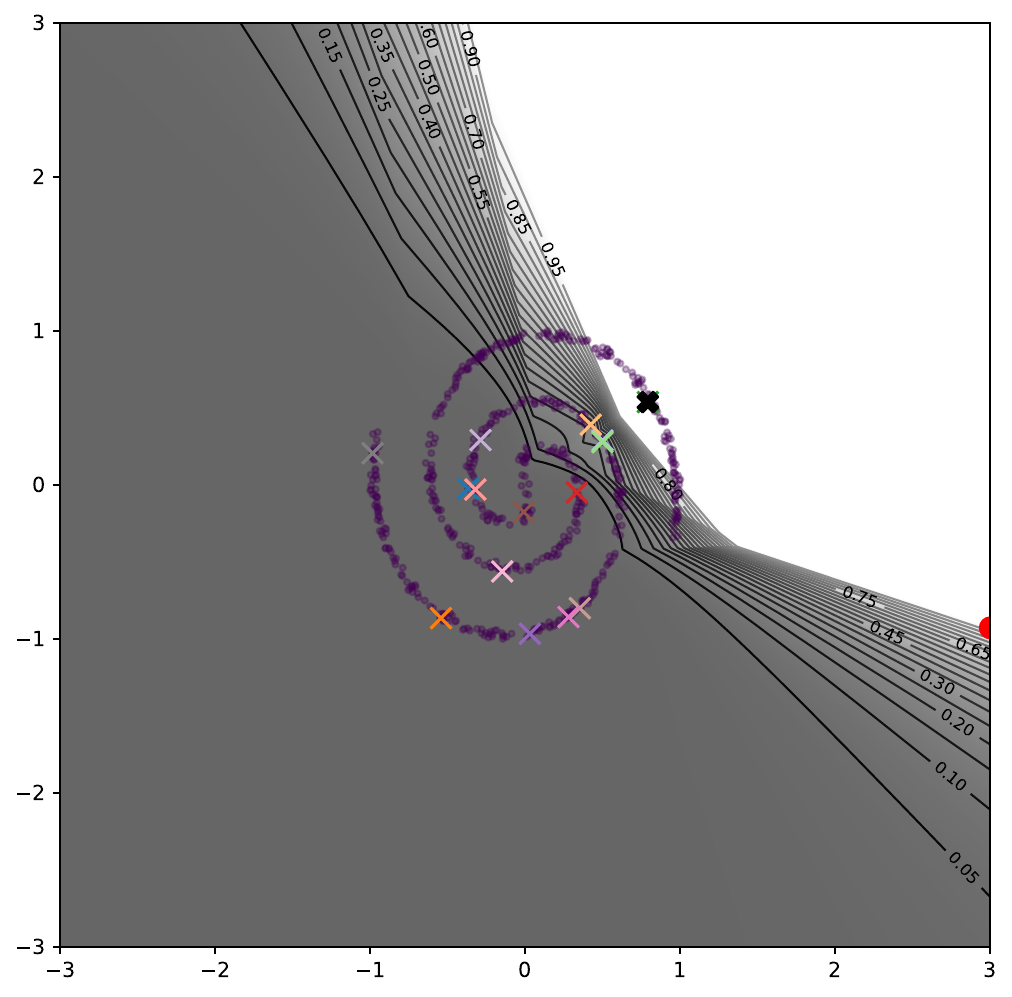}
         \caption{neuron 1: un-normalized}
     \end{subfigure}
     \begin{subfigure}{0.3\textwidth}
         \centering
         \includegraphics[width=\textwidth]{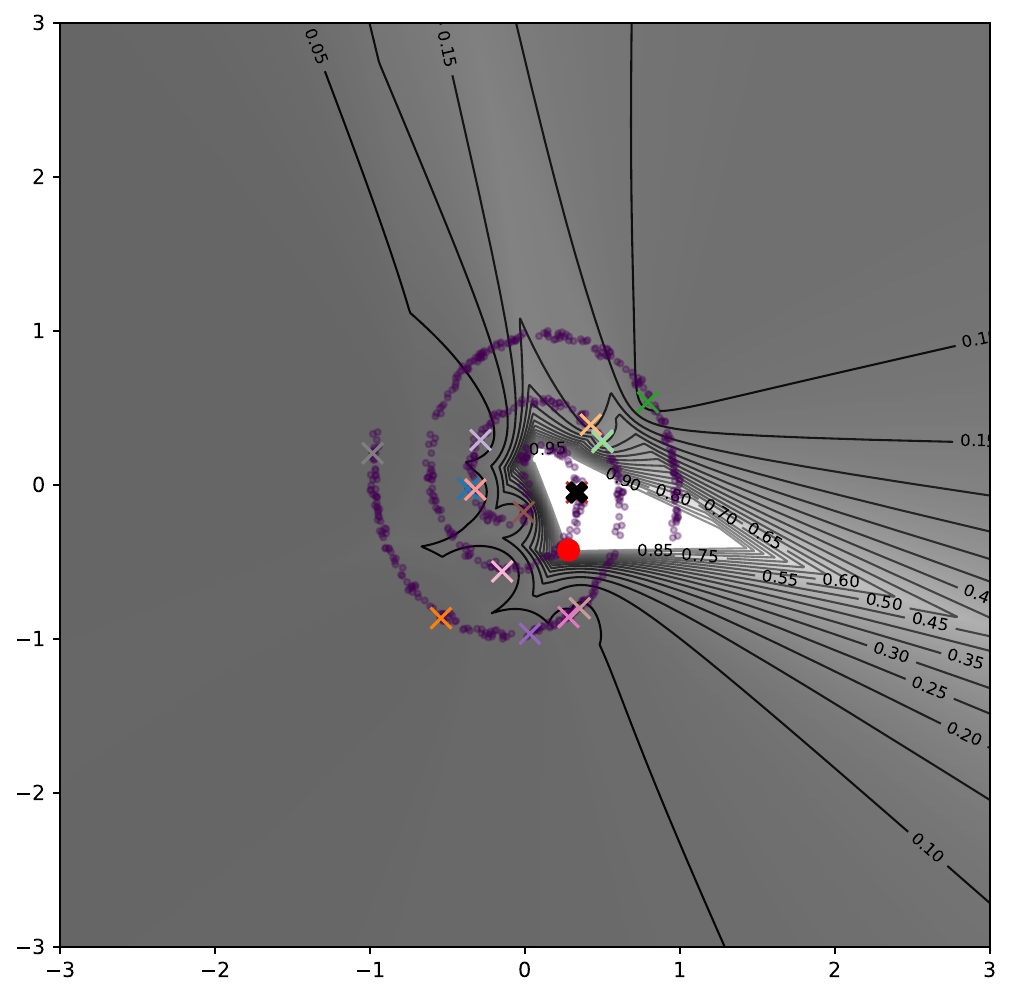}
         \caption{neuron 2: un-normalized}
     \end{subfigure} 
     \newline
      \begin{subfigure}{0.3\linewidth}
         \centering
         \includegraphics[width=\textwidth]{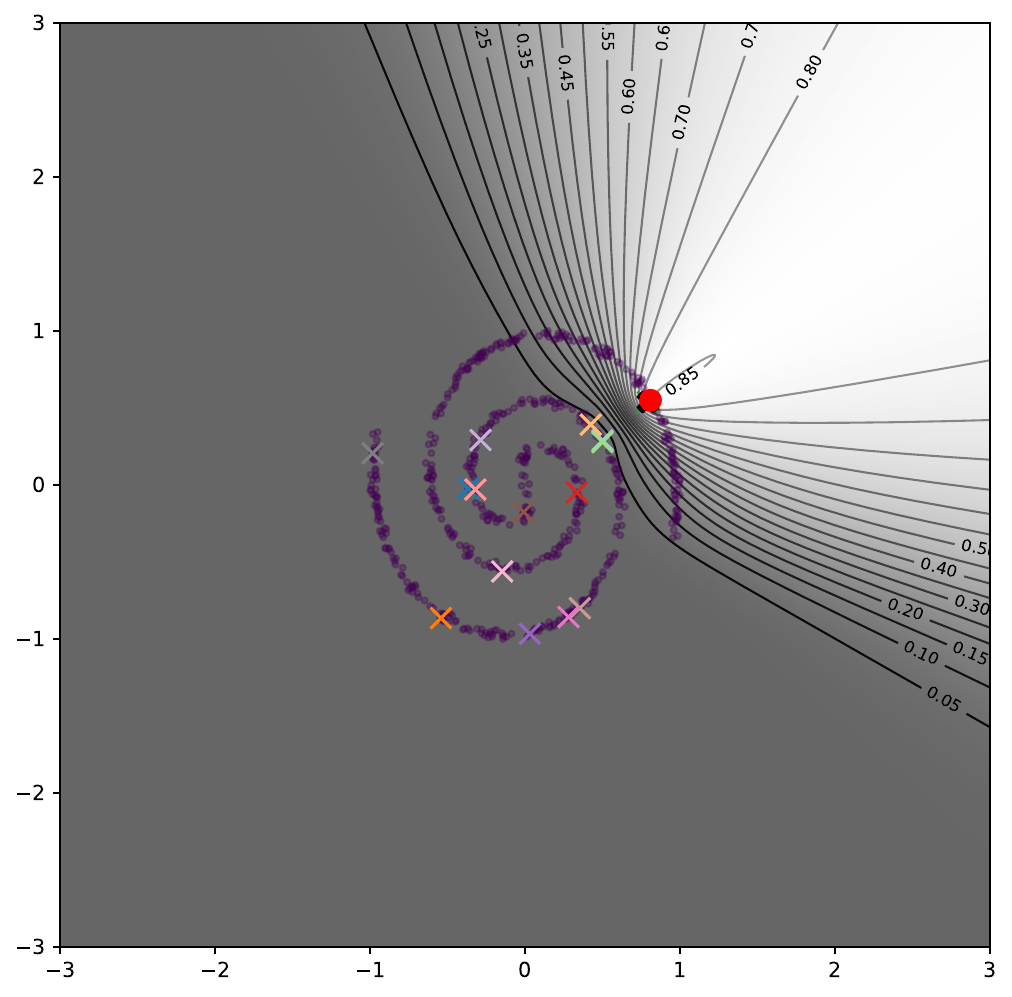}
         \caption{neuron 1: $\tau = e^{-2}, \epsilon = \textrm{None}$}
     \end{subfigure}
     \begin{subfigure}{0.3\textwidth}
         \centering
         \includegraphics[width=\textwidth]{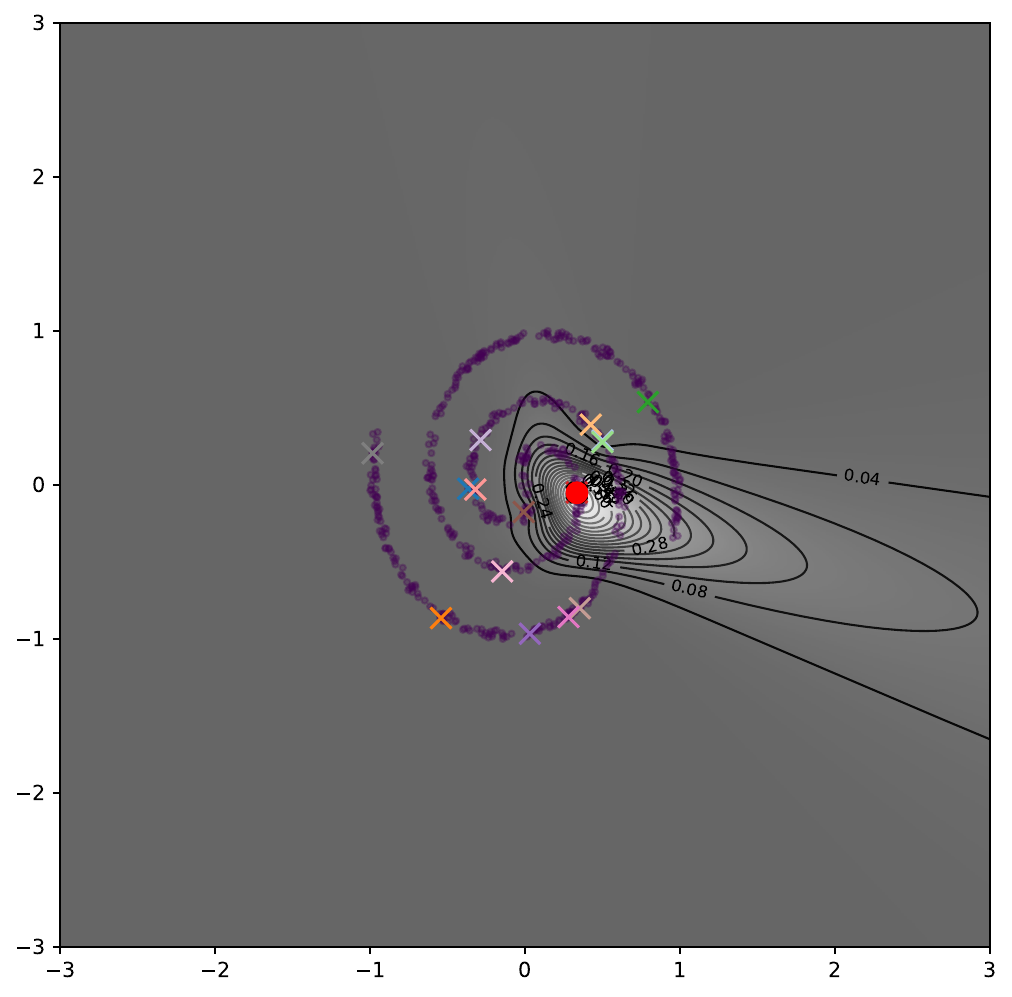}
         \caption{neuron 2: $\tau = e^{-2}, \epsilon = \textrm{None}$}
     \end{subfigure} 
     \newline
     \begin{subfigure}{0.3\textwidth}
         \centering
         \includegraphics[width=\textwidth]{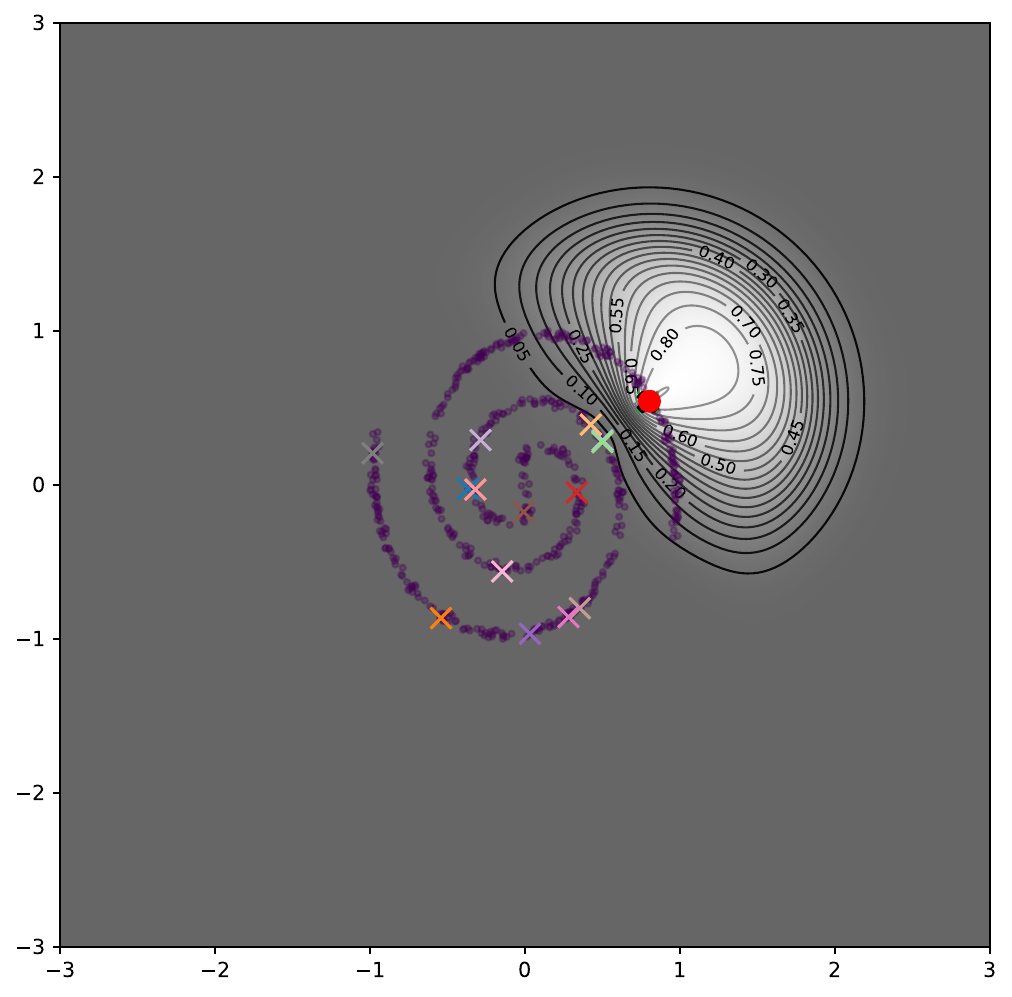}
         \caption{neuron 1: $\tau = e^{-2}, \epsilon = 1.0$}
     \end{subfigure}
     \begin{subfigure}{0.3\textwidth}
         \centering
         \includegraphics[width=\textwidth]{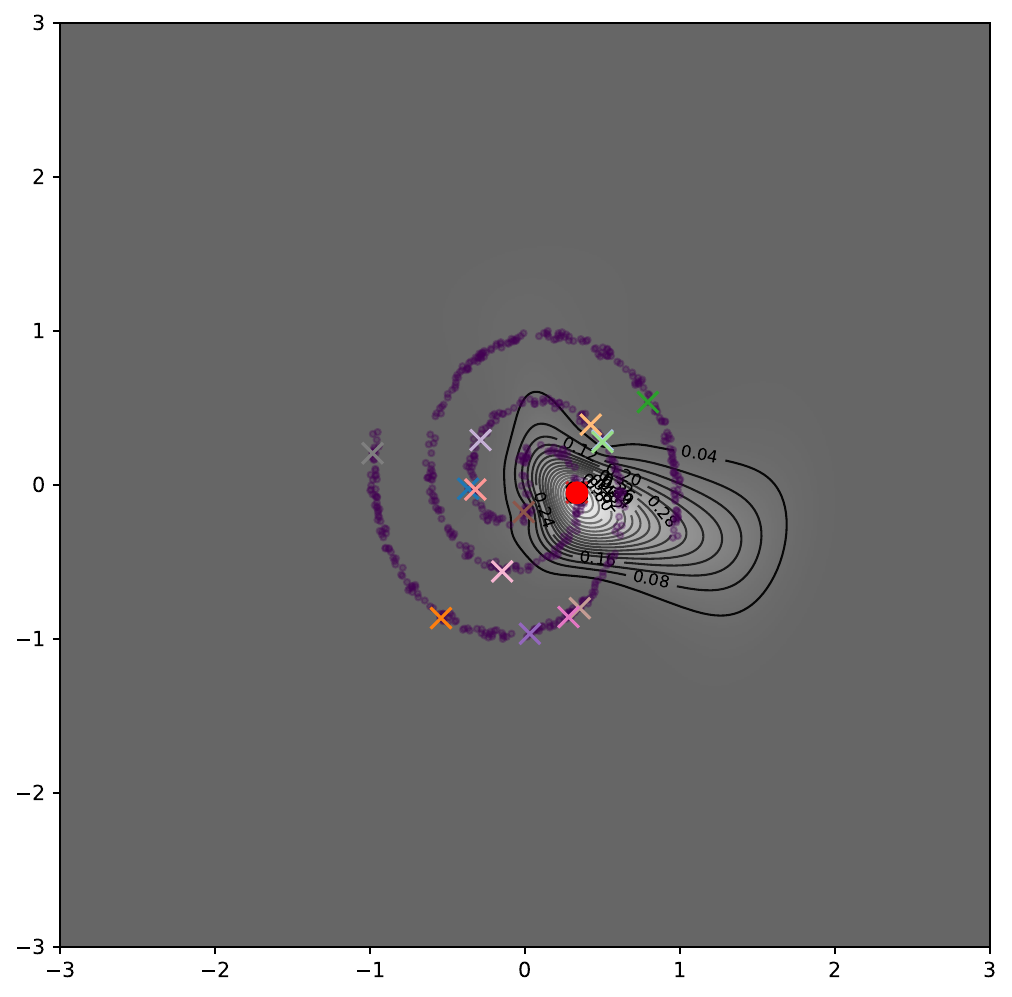}
         \caption{neuron 2: $\tau = e^{-2}, \epsilon = 1.0$}
     \end{subfigure}
     \begin{subfigure}{0.3\textwidth}
         \centering
         \includegraphics[width=\textwidth]{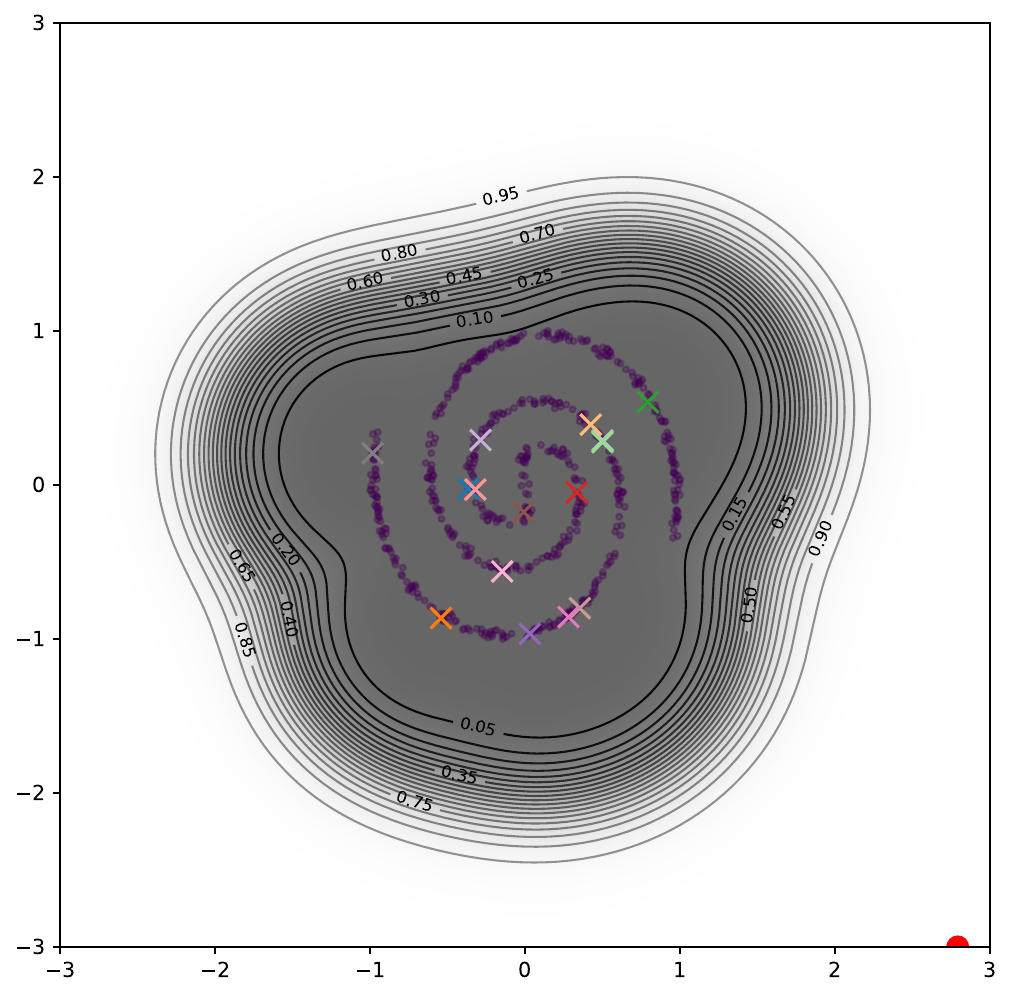}
         \caption{neuron $\epsilon$: $\tau = e^{-2}, \epsilon = 1.0$}
     \end{subfigure}

     \begin{subfigure}[b]{0.3\textwidth}
         \centering
         \includegraphics[width=\textwidth]{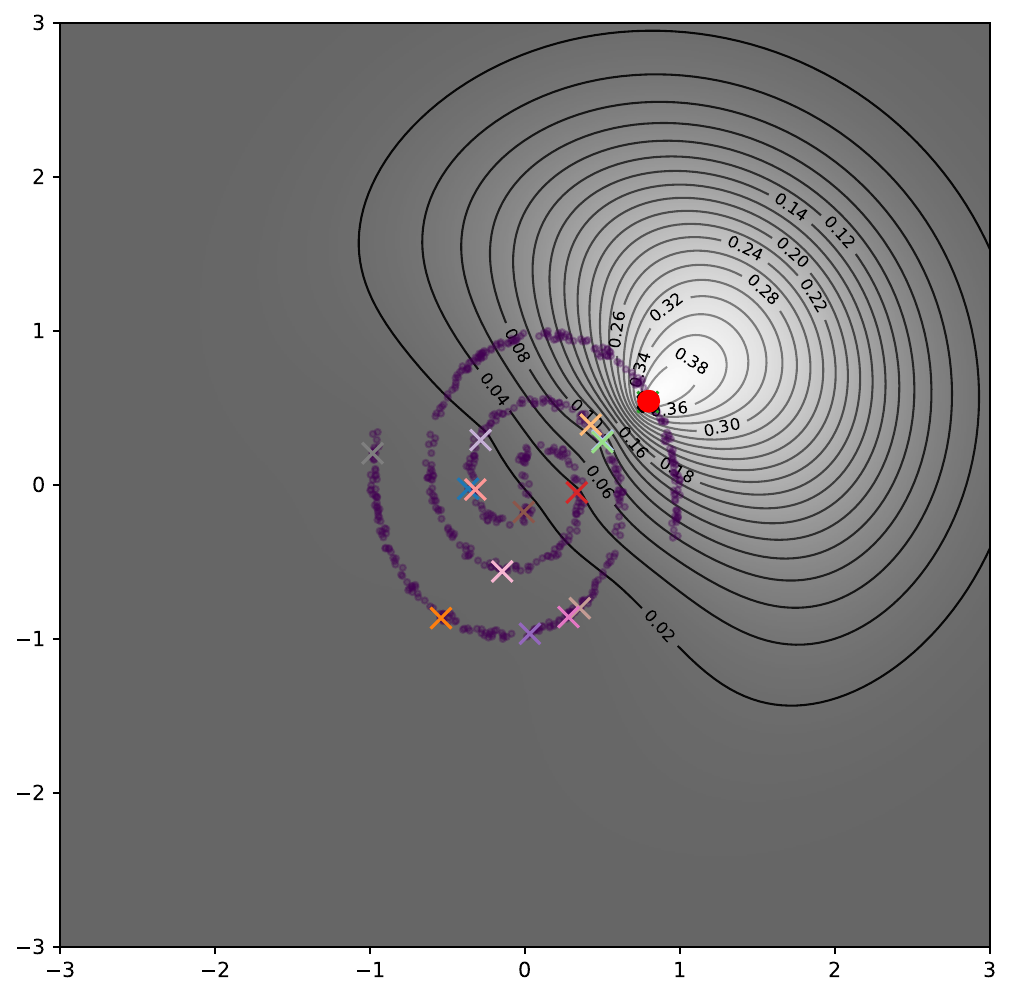}
         \caption{neuron 1: $\tau = e^{-1}, \epsilon = 1.0$}
     \end{subfigure}
     \begin{subfigure}[b]{0.3\textwidth}
         \centering
         \includegraphics[width=\textwidth]{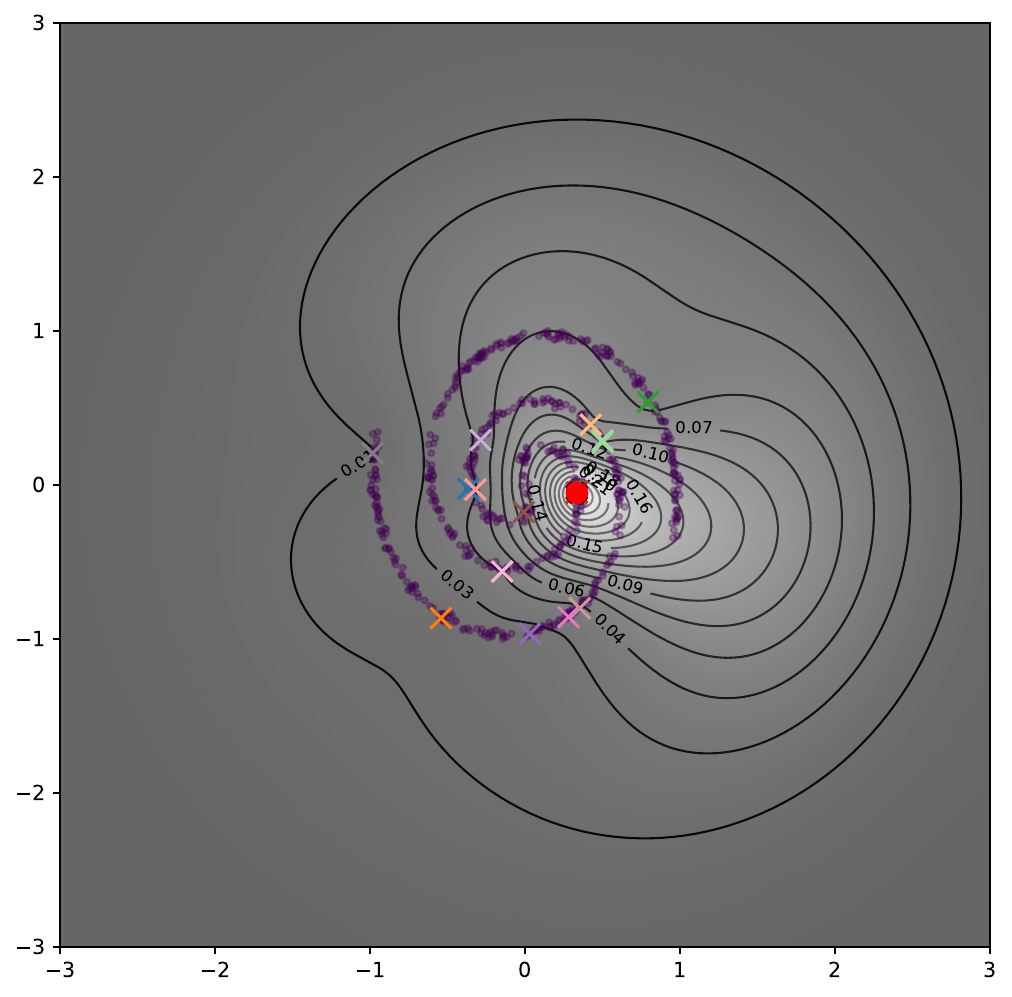}
         \caption{neuron 2: $\tau = e^{-1}, \epsilon = 1.0$}
     \end{subfigure}
     \begin{subfigure}[b]{0.3\textwidth}
         \centering
         \includegraphics[width=\textwidth]{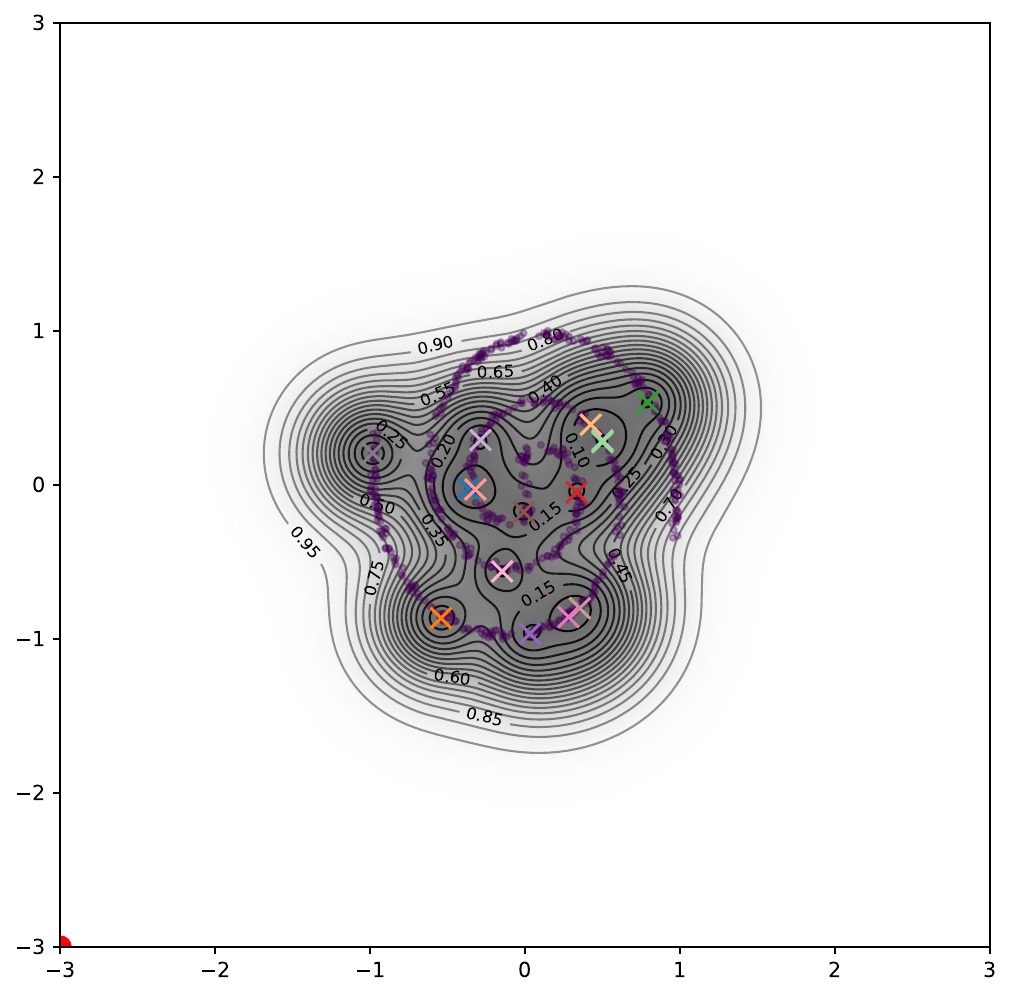}
         \caption{neuron $\epsilon$: $\tau = e^{-2}, \epsilon = 0.3$}
     \end{subfigure}
        \caption{Visualization of region of neuron firing in toy 2D setting with various parameters $\tau$ and $\epsilon$ ($\epsilon =$None is $\epsilon = \infty$ or without $\epsilon$).}
        \label{fig:epsilon-sm-neurons}
\end{figure*}

The softmax function has been used widely to create probability distribution given un-normalized inputs. Use of softmax for interpretable features is not new. It has been used in MLP layers for interpretability~\cite{sukhbaatar2015end}, and also in the Attention layer itself where, the $K,Q,V$ are dependent on data. Moreover, SoLU~\cite{elhage2022solu} also uses softmax to create interpretable neurons. These previous works does not take into consideration distances or metrics for measuring similarity.

$$f_{softmax-sim}(\vx, K, \tau) = \frac{\exp(-\vd/\tau)}{\sum_{i} \exp(-\vd_{[i]}/\tau)} ; \hspace{2em} \vd = \{ f_{metric}(\vx, \vk_i) \}$$ for $i$ represents index of all the keys (or centers). Here, $\tau$ is a temperature to change the hardness of softmax activation. 
Softmax, as used in Boltzman distributon~\cite{softmax}, has a temperature parameter controlling the hardness of the distribution. The effect of temperature can also be visualized easily as shown in Figure~\ref{fig:epsilon-sm-neurons}. This parameter has useful interpretation. If $\tau\to0$, then we get $argmax$, where only 1 winning neuron fires. This is useful to create activations that are unimodal, disentangled or activate very sharply for certain Keys.

With few experiments we find that metrics($\vd$) transformed to similarity by function $f: \mathbf{R} \to \mathbf{R}$; $f$ is convex and monotonically decreasing function, like $e^{-x}$ and $-log(x)$ or $1/x$ for $x>0$ or even $-x$, preserves the property that maxima is at the zero distance from key. Functions like $\cos{\theta}; 0<\theta<\pi$ or $e^{-x^2}$ does not preserve the property.

Despite having normalized similarity, the second problem remains. Neurons can produce high activation in region very far away from the reference centers ($\textbf{K}$) under certain condition of temperature and position of other centers (see Figure~\ref{fig:epsilon-sm-neurons}). The activation grows radially in direction of a center from other centers, which is undesirable to create a neuron with local influence.


\subsubsection{Distance Transform based $\epsilon$-softmax Neurons}

\paragraph{$0$-softmax:}
We were facing the above mentioned problem when we found about $0$-softmax~\cite{0softmax}, where the intuition is that $0$ entry is added to the softmax probability calculation so that when similarity measure $(Q.K)$  do not match, it does not normalize attention score to $1$. It allows the attention to attend nothing as well, instead of ordinary softmax attention that sum to $1$.

\paragraph{$\epsilon$-softmax:}
This is our interpretation and modification of 0-softmax applied to distance based neuron. We found that we can set a value of epsilon when calculating neuron activations (or $f_{similarity}$). The epsilon($\epsilon$) value sets a threshold on similarity, and if the similarities goes below the threshold, the epsilon neuron produces the highest activation. Hence we can use the epsilon neuron to represent region where the similarity to any $\vk$ey is low. 

In practice, we move the epsilon value from softmax to the calculation of distance itself. It can be interpreted as the distance in the input space above which it activates the most. Since softmax is independent on the shift we only add the temperature parameter to scale the distances. The similarity is calculated as follows.

$$f_{\epsilon-softmax-sim}(\vx, K, \tau) = \frac{\exp(-\vd/\tau)}{\sum_{i} \exp(-\vd_{[i]}/\tau)} ;  \hspace{2em} \vd = \{ f_{metric}(\vx, \vk_i) \} \cup \{\epsilon\} $$ where symbols are same as previous. The value of epsilon($\epsilon$) behaves like a threshold value on distances. We may also use bias with distances($\vd$) to increase or decrease influence of individual keys (or neurons) on the similarity score. For experiments, we skip the bias term to make interpretation simple.

This function seems to have truely local activation and has capacity to disentangle the activation region of closely located neurons as well. The activation of such neuron is shown in Figure~\ref{fig:epsilon-sm-neurons} along with change in temperature as well. This similarity is equivalent to softmax-similarity when $\epsilon = \infty$. Hence, we work on top of this more general neuron to create basic interpretable architectures.

%% file: 12_local_neuron_extension.tex
\subsubsection{Local Residual MLP}

We interpret the residual network~\cite{he2016deep} of form $\mathbf{y} = \mathbf{x} + f_{similarity} (\mathbf{x}, \mathbf{K}).\mathbf{S}$.
The similarity function $f_{similarity}$ produces peak activation of $\approx \mathbf{1}$ when the center and data are similar, like previous section. For low temperature ($t$), the softmax decision boundary is hard, and Voronoi diagram is produced. The higher temperature ($t$) is useful for training due to its smooth gradients.

\begin{figure}[h]
  \centering

\includegraphics[width=0.350\textwidth]{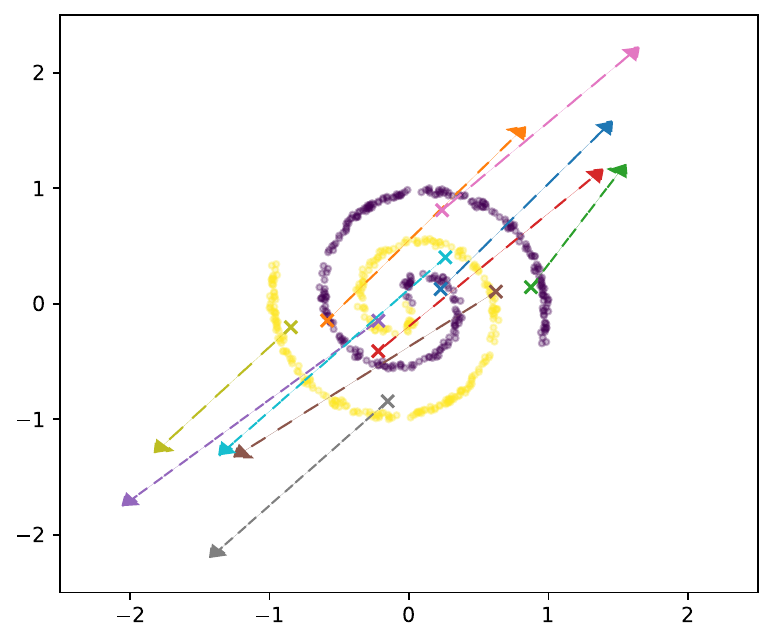}
\includegraphics[width=0.350\textwidth]{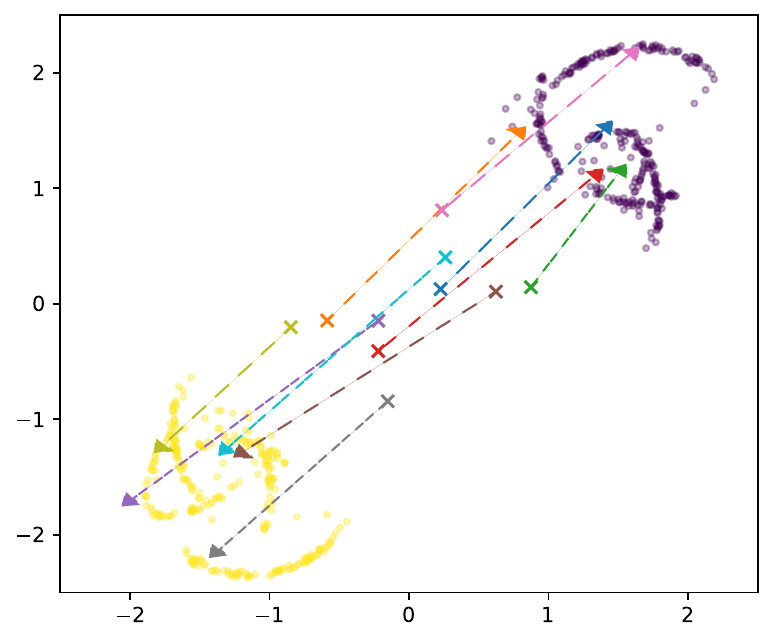}

  \caption{
Local Residual Layer interpretation; visualization of the dataset ($\cdot$) along with centers ($\times$); Residual MLP moves the centers as shown by the arrow. \textit{(LEFT)} $\textbf{x}$-space \textit{(RIGHT)} $\textbf{x}+f_{res}(\textbf{x}) = \textbf{y}$ space. We make the shift of $\epsilon$-neuron to be $\textbf{0}$.}
  \label{fig:local_residual}
\end{figure}

The vector $\vs_i$ for each $i^{th}$ neuron represents the shift of $\mathbf{x}$-space after residual as shown in Figure(~\ref{fig:local_residual}). The shift vector $\vs_i$ and centers $\vk_i$ can be initialized with data or trained with backpropagation~\cite{rumelhart1986learning}.

Figure~\ref{fig:local_residual} shows local residual MLP with $\epsilon = 0.5$ neuron. Due to small temperature, the highest activation of a neuron $<<1$, i.e. the activation is smooth and hence overall shift is also smooth. Importantly, we are able to interpret the non-linear classification of 2-spirals dataset using the concept of centers, similarity and shift.

%% file: 13_uncertainty_estimation_and_adv_rejection.tex
\subsection{$\epsilon$-softmax for Uncertainty Estimation and Adversarial Rejection}

To start at the problem of adversarial example rejection, we want to tackle uncertainty estimation using the epsilon neuron. The higher the epsilon activation, the farther the input is from known centers ($K$) and vice-versa. The prediction is maximally certain when the data point $x$ is the same as one of the $K$. Random input or adversarial example has higher chance to activate epsilon($\epsilon$) neuron.

\begin{wrapfigure}{r}{0.4\textwidth}
  \vspace{-1em}
  \begin{center}
    \includegraphics[width=0.39\textwidth]{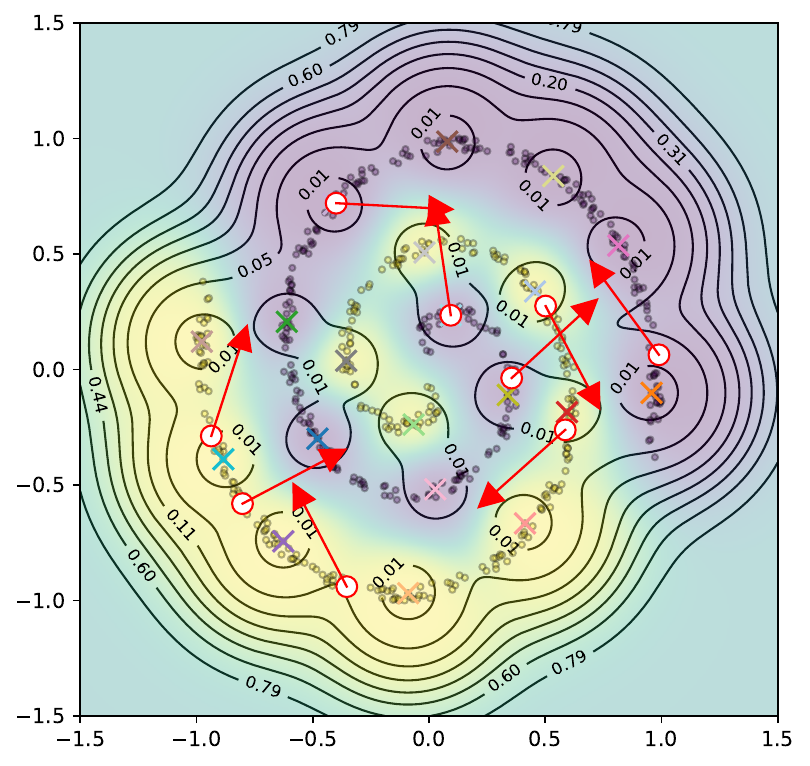}
  \end{center}
  \vspace{-1em}
  \caption{Negative adversarial gradient direction on a 2D dataset. Contour showing $\epsilon$-neuron.}
  \label{fig:toy_adversarial_grads}
\vspace{-4em}
\end{wrapfigure}

In a simple 2D spiral dataset we can understand adversarial examples as shown in Figure~\ref{fig:toy_adversarial_grads}. We can see that the adversarial examples are well behaved, pointing towards region of wrong class. We can interpret the adversarial gradient as sum of two objectives: $1)$ it tries to steer input away from current correct neuron activation, and $2)$ towards neuron predicting wrong class. Combined, we get the adversarial gradient.  
Next, we aim to make use of “Locally Activating Neurons” achieved with $\epsilon$-Softmax based dictionary MLP to detect adversarial examples~\cite{szegedy2013intriguing, goodfellow2014explaining}.

\textbf{Experiment on fMNIST with 2 layered dictionary MLP}\\
With experiments on 2 layer Dictionary MLP as developed in Section~\ref{subsec:dictionary_learning}, we tried to reject random samples using epsilon($\epsilon$) neuron. We were able to do it easily by checking if the epsilon($\epsilon$) neuron was firing the most.

Moreover, we tested if the $\epsilon$-neuron would fire highest for adversarial examples, and after tuning value of epsilon, we could reject many adversarial examples. We study extensively to understand the adversarial rejection property of local neurons.

\begin{figure}[t]
     \begin{subfigure}{0.45\linewidth}
         \centering
         \includegraphics[width=\textwidth]{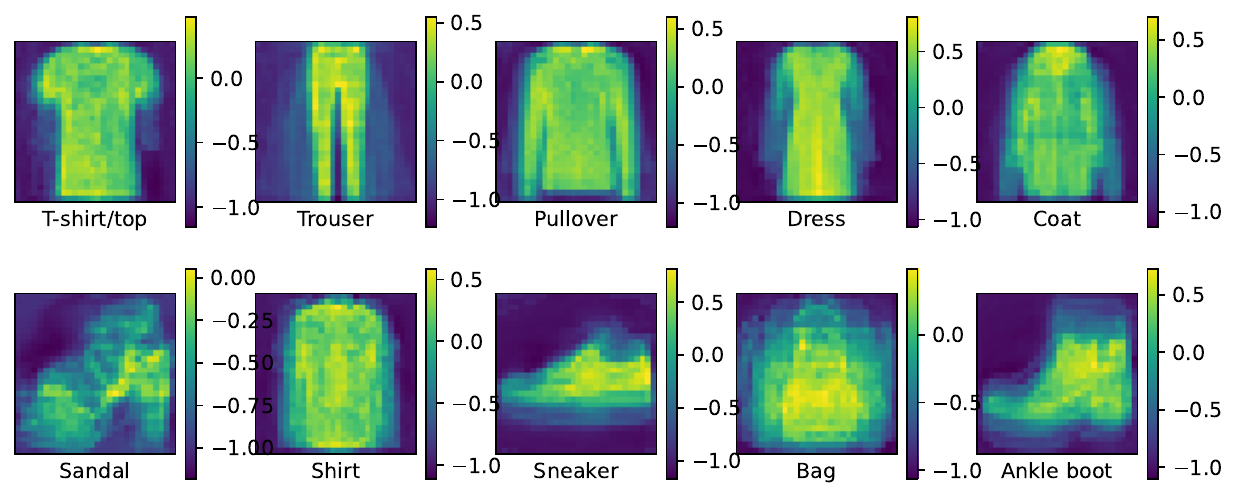}
         \caption{(mean class centers) init:data, clr:0.01}
     \end{subfigure}\hfill
     \begin{subfigure}{0.45\linewidth}
         \centering
         \includegraphics[width=\textwidth]{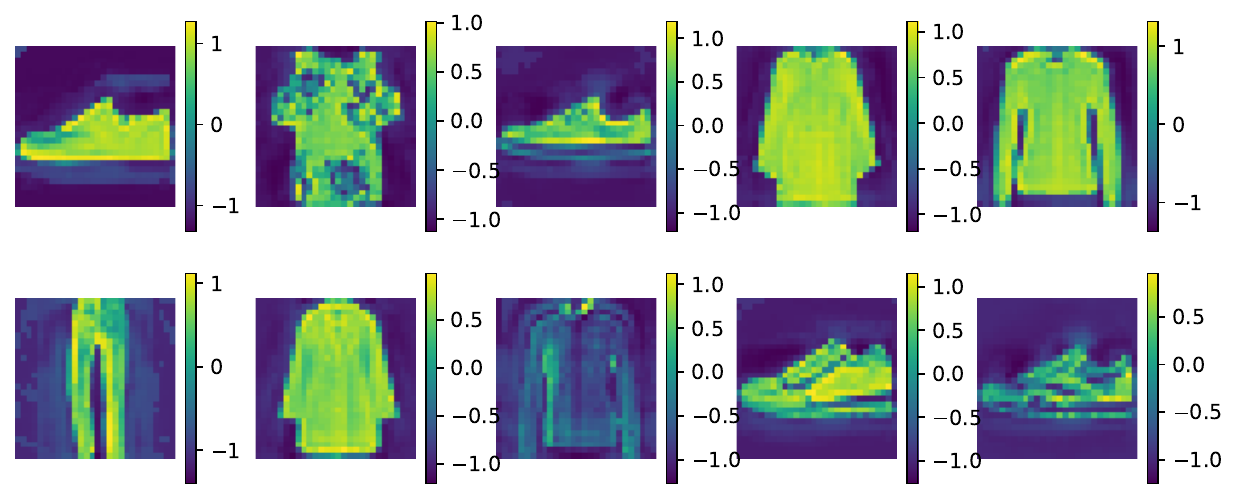}
         \caption{(sample centers) init:data, clr:0.01}
     \end{subfigure}\\
               \begin{subfigure}{0.45\linewidth}
         \centering
         \includegraphics[width=\textwidth]{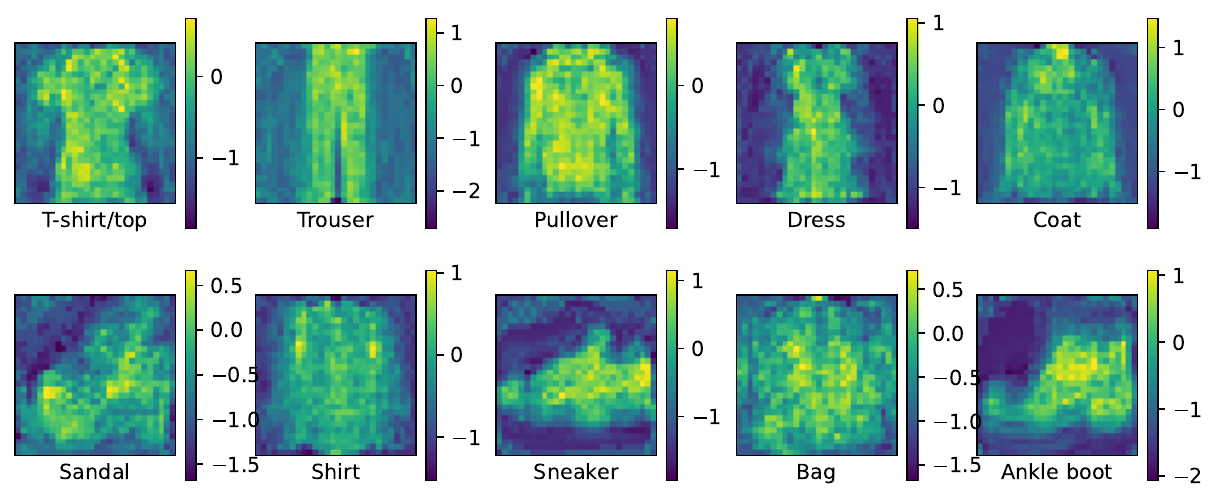}
         \caption{(mean class centers) init:data, clr:1.0}
     \end{subfigure}\hfill
     \begin{subfigure}{0.45\linewidth}
         \centering
         \includegraphics[width=\textwidth]{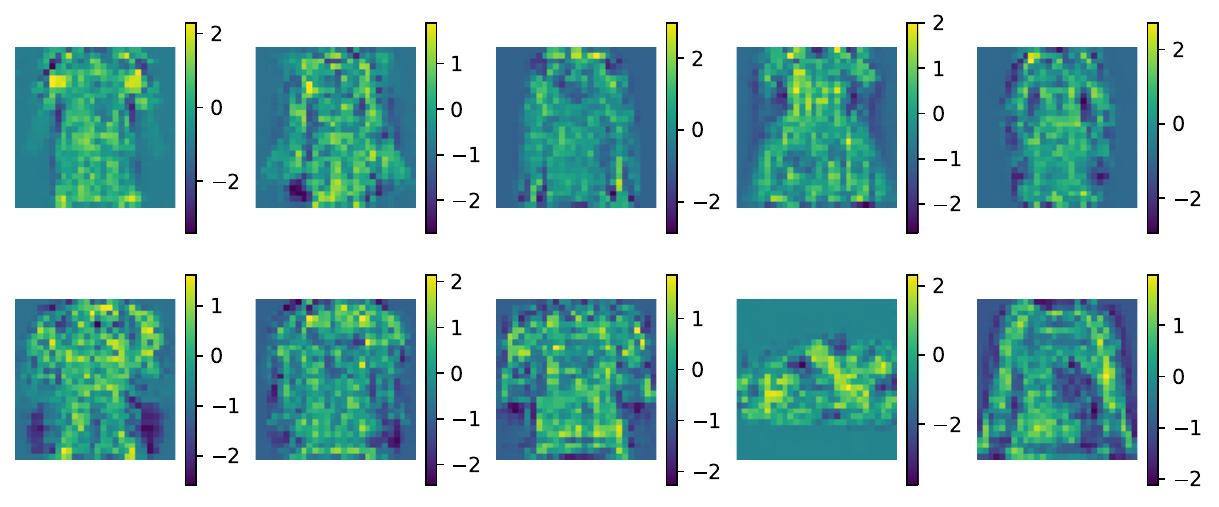}
         \caption{(sample centers) init:data, clr:1.0}
     \end{subfigure}\\
     \begin{subfigure}{0.45\textwidth}
         \centering
         \includegraphics[width=\textwidth]{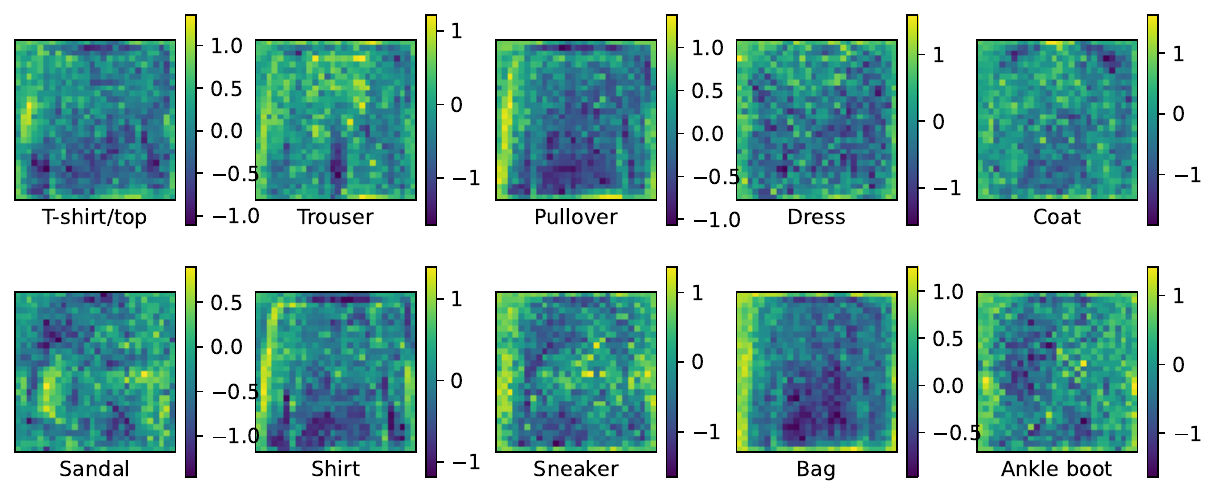}
         \caption{(mean class centers) init:rand, clr:1.0}
     \end{subfigure} \hfill
     \begin{subfigure}{0.45\textwidth}
         \centering
         \includegraphics[width=\textwidth]{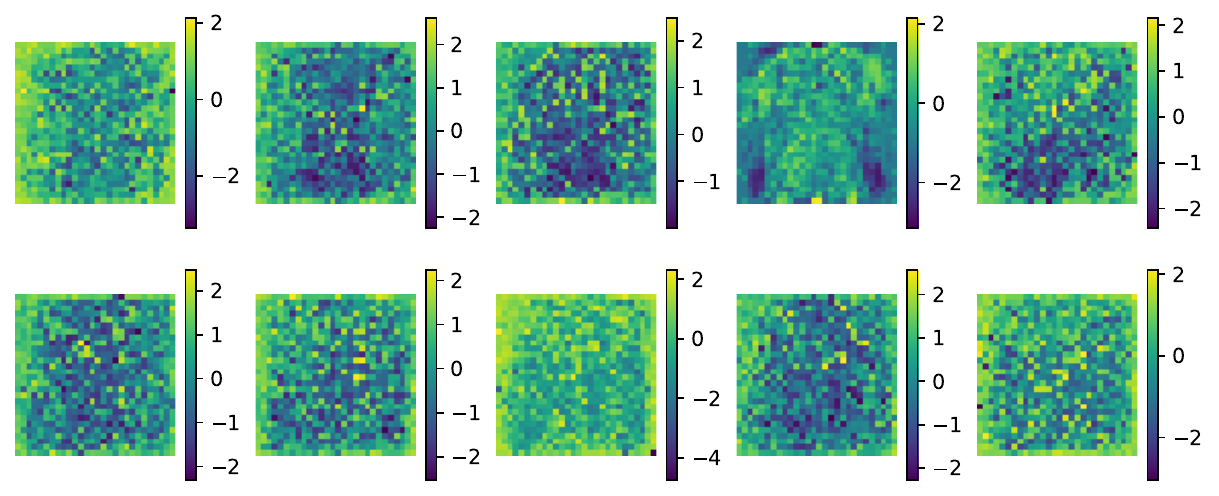}
         \caption{(sample centers) init:rand, clr:1.0}
     \end{subfigure}\\
     \begin{subfigure}{0.45\textwidth}
         \centering
         \includegraphics[width=\textwidth]{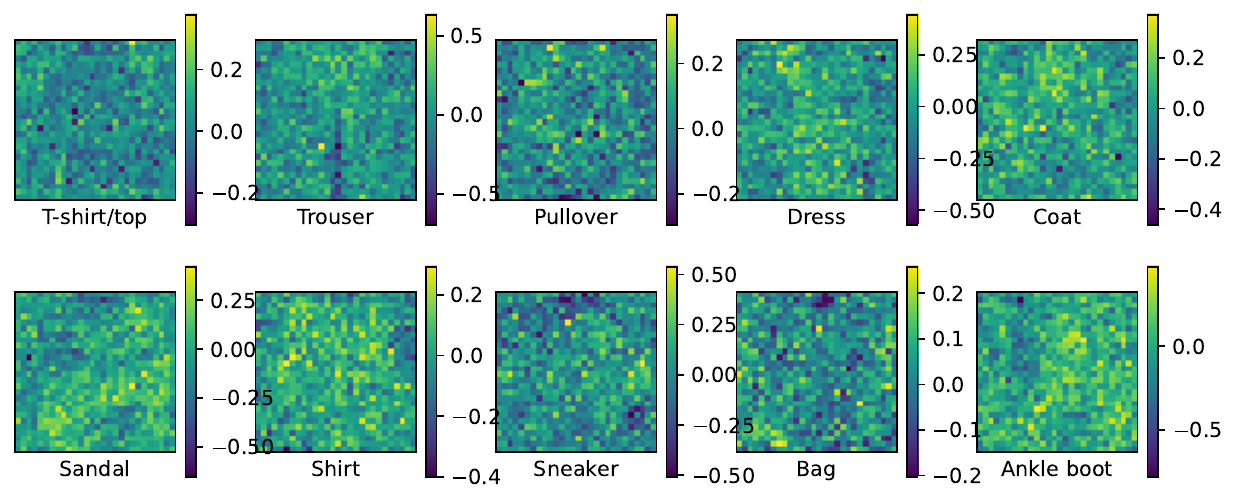}
         \caption{(mean class centers) init:rand, clr:0.01}
     \end{subfigure}\hfill
     \begin{subfigure}{0.45\textwidth}
         \centering
         \includegraphics[width=\textwidth]{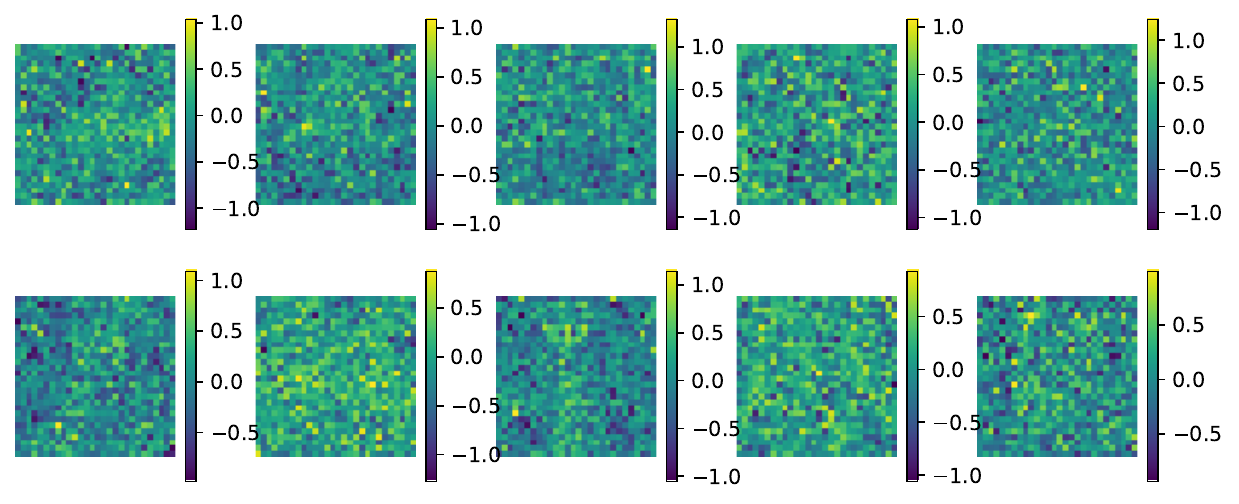}
         \caption{(sample centers) init:rand, clr:0.01}
     \end{subfigure}
     
        \caption{Analyzing the centers of model trained with different initialization($init$) and key/center learning rate ($clr$) }
        \label{fig:adv_model_centers}
\end{figure}

\emph{\textbf{Setting up Attack:}}
We search for the value of epsilon ($\epsilon$) for which the adversarial attacks were rejected. We found that for certain value of epsilon, the model rejects adversarial examples, but not the true data samples. 

For low epsilon, all data and adversarial examples would be rejected, and for high value of epsilon, the model would not reject either adversarial examples or input samples. We look for middle ground.\\
\textit{x-rejected}: fraction of inputs that were rejected.
\textit{rejected}: adversarial examples that were rejected.\\
\textit{failed}: adversarial attack success, and not rejected.
\textit{measure}: \textit{x-rejected}+\textit{failed} is the quantity we want to minimize.

While training, we use epsilon($\epsilon$) $= \texttt{EMA}(d)$, where $d =$ average distance between $\vx$ and centers($\textbf{K}$). Here, $\texttt{EMA}$ stands for exponentially moving average.

\begin{figure}[h]
     \begin{subfigure}{0.48\linewidth}
         \centering
         \fbox{\includegraphics[width=\textwidth]{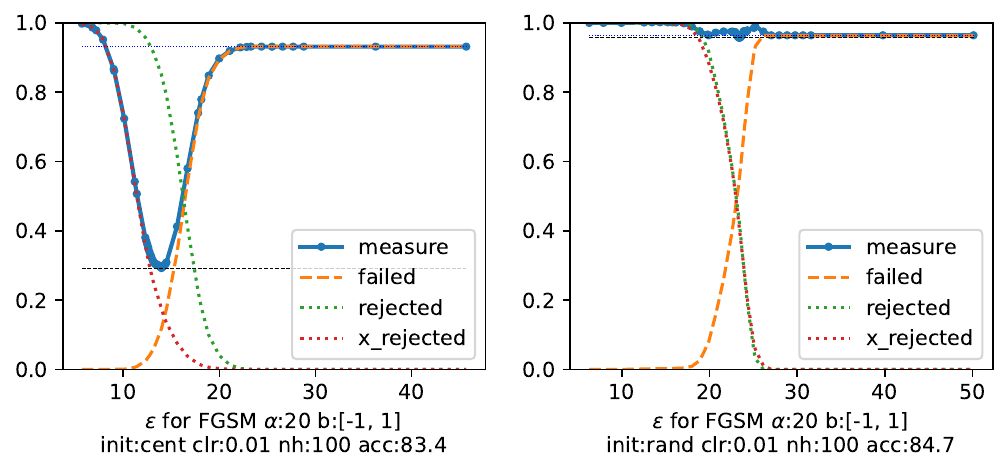}}
     \end{subfigure}\hfill
     \begin{subfigure}{0.48\textwidth}
         \centering
         \fbox{\includegraphics[width=\textwidth]{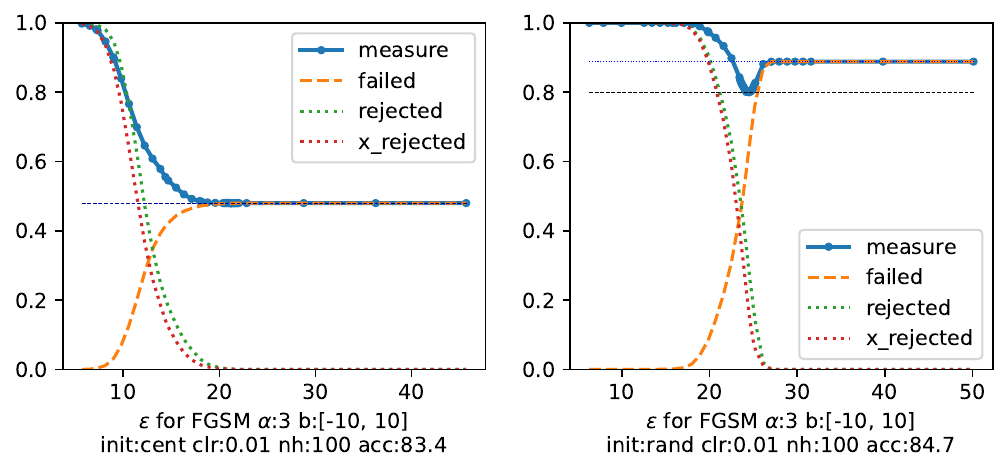}}
     \end{subfigure}\vspace{1em}
      \begin{subfigure}{0.48\linewidth}
         \centering
         \fbox{\includegraphics[width=\textwidth]{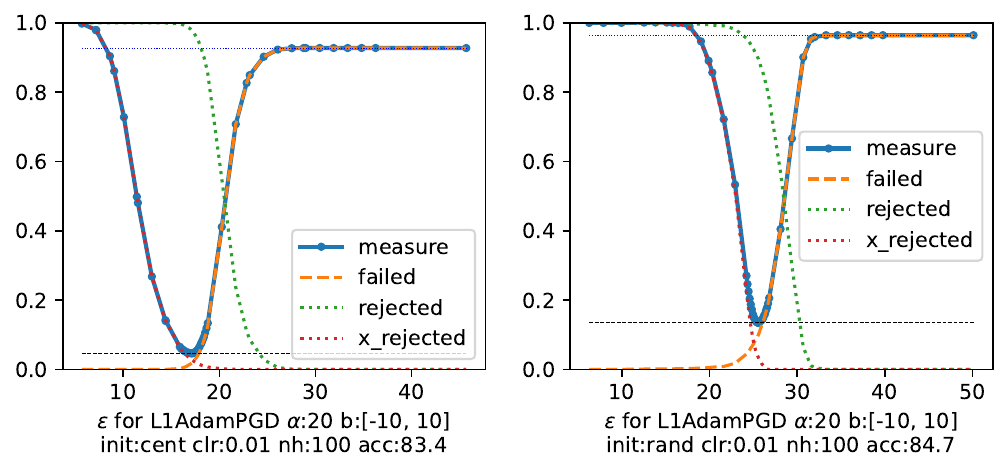}}
     \end{subfigure}\hfill
     \begin{subfigure}{0.48\textwidth}
         \centering
         \fbox{\includegraphics[width=\textwidth]{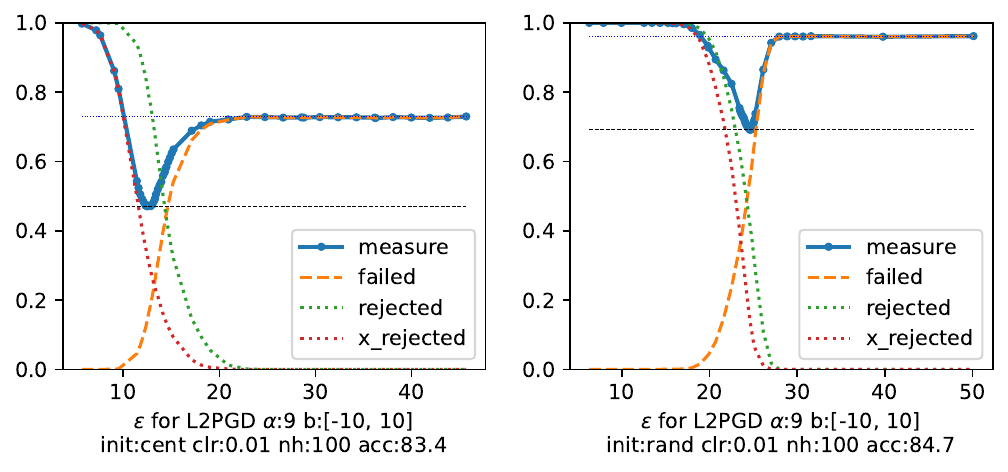}}
     \end{subfigure}\vspace{1em}
          \begin{subfigure}{0.48\linewidth}
         \centering
         \fbox{\includegraphics[width=\textwidth]{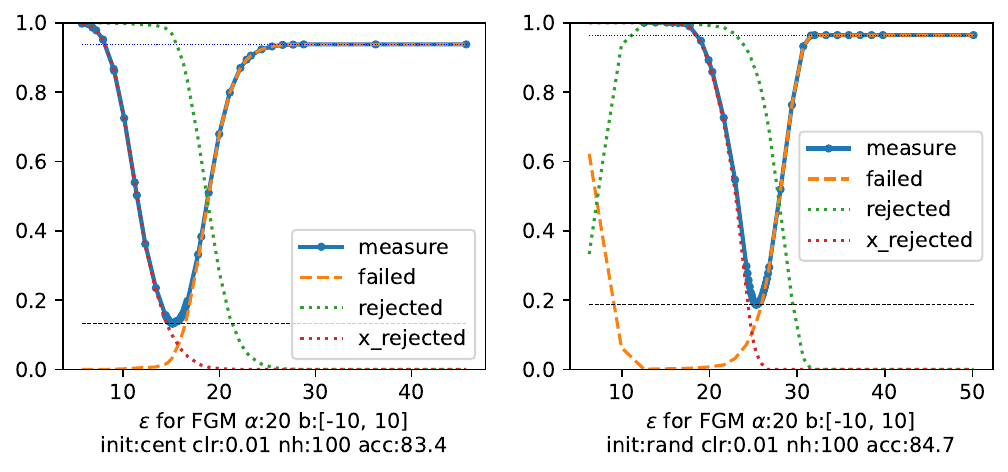}}
     \end{subfigure}\hfill
     \begin{subfigure}{0.48\textwidth}
         \centering
         \fbox{\includegraphics[width=\textwidth]{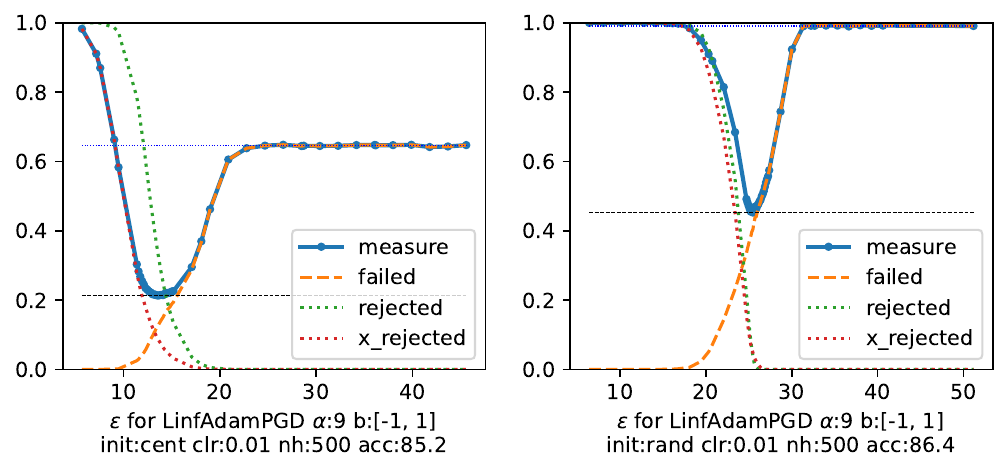}}
     \end{subfigure}
     \newline
        \caption{Adversarial Rejection with center initialized with data versus center initialized with random sample. The center-learning-rates($clr$) is low to not allow centers to change much. All pair of comparison is done with different attacks and adversarial-learning-rate($\alpha$).}
        \label{fig:adv_rejection_init_compare}
\end{figure}

\begin{figure*}
     \begin{subfigure}{0.48\linewidth}
         \centering
         \fbox{\includegraphics[width=\textwidth]{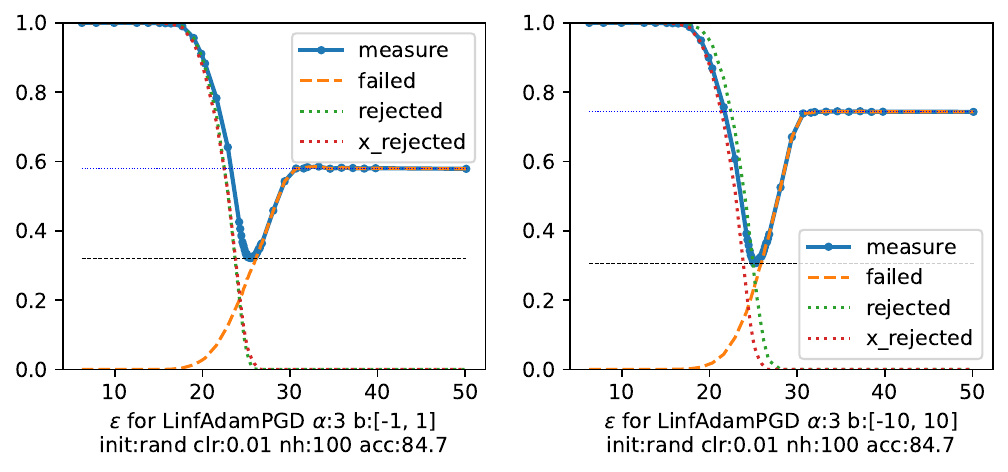}}
     \end{subfigure}\hfill
     \begin{subfigure}{0.48\textwidth}
         \centering
         \fbox{\includegraphics[width=\textwidth]{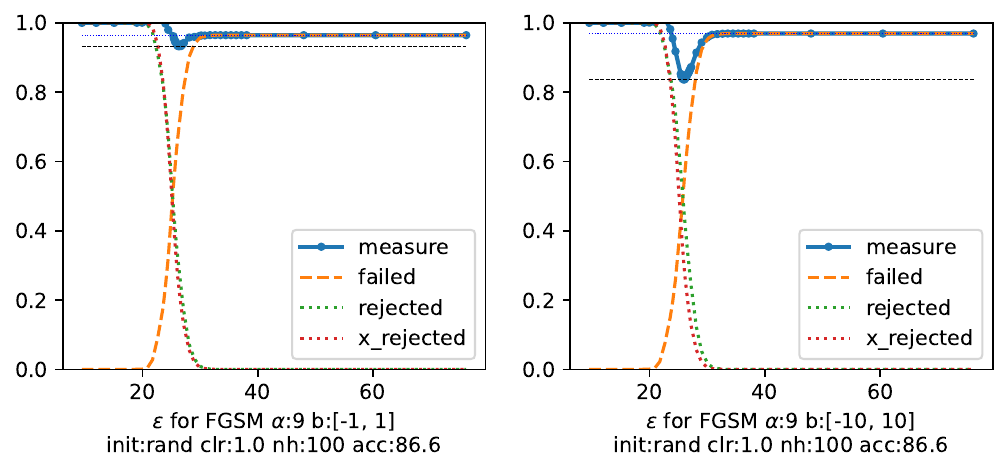}}
     \end{subfigure}\vspace{1em}
      \begin{subfigure}{0.48\linewidth}
         \centering
         \fbox{\includegraphics[width=\textwidth]{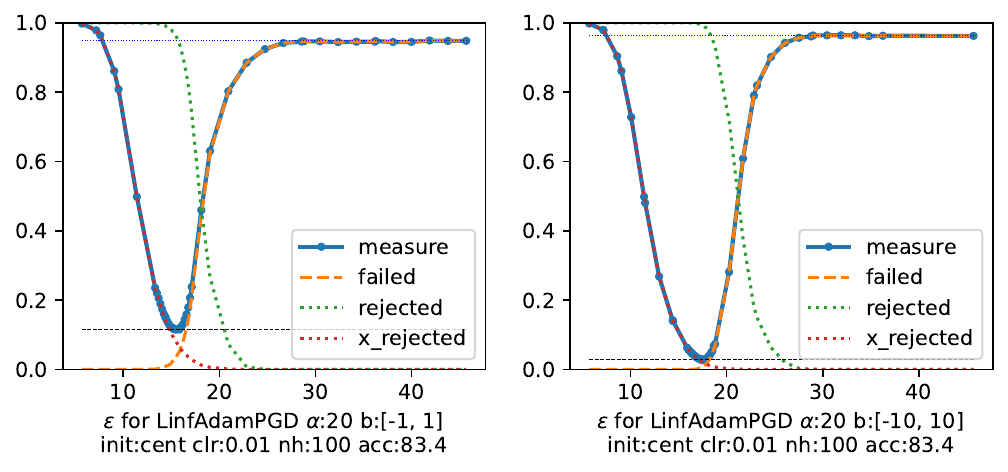}}
     \end{subfigure}\hfill
     \begin{subfigure}{0.48\textwidth}
         \centering
         \fbox{\includegraphics[width=\textwidth]{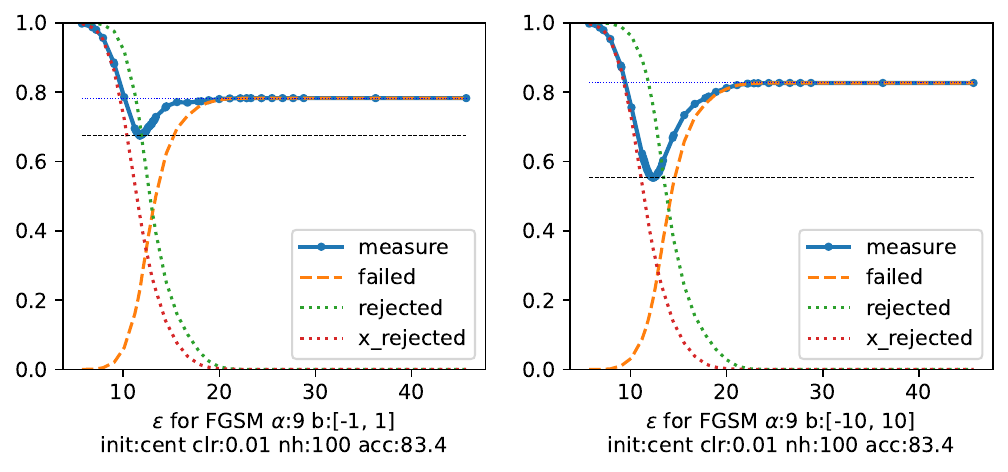}}
     \end{subfigure}
     \newline
        \caption{Adversarial Rejection with bounds in [-1,1] versus [-10, 10]. We try to sample various configurations.}
        \label{fig:adv_rejection_bound_compare}
\end{figure*}

To search sweet spot for rejecting adversarial examples, we evaluate within some range of learned epsilon as shown in Figure~\ref{fig:adv_rejection_init_compare} and \ref{fig:adv_rejection_bound_compare}. With experiments using various adversarial attack methods, and with various learning rate, we analyze the \textit{measure} of adversarial rejection.

\emph{\textbf{Adversarial Attacks:}} For generating adversarial attacks, we use multiple methods from foolbox~\cite{rauber2017foolboxnative}: FGSM, FGM, L2PGD, LinfPGD, L1AdamPGD, L2AdamPGD, LinfAdamPGD and L2AdamBasic. Moreover, we believe that white box adversarial attacks are the hardest as the model gives up the information to fool itself. 

\emph{\textbf{Normalizing Attack Magnitude:}} We find that some adversarial attacks were unsuccessful due to small magnitude of the attack. Hence, we attack the model with normalized gradient.
$$\vx_{adv} = \vx - \alpha \frac{\vg_{x}}{\|\vg_{x}\|}$$
Where, $\alpha$ is the learning-rate or attack intensity, $\vx_{adv}$ is the adversarial example, $\vx$ is the input, and $\vg_{x}$ is the adversarial gradient. We experiment with increasing value of learning rate.

\emph{\textbf{Setting up Model:}}
We test adversarial attack rejection on model 1) trained with random initialization of parameters and 2) with random training samples for initialization. The models have 100 or 500 hidden units and are trained for 30 epochs. We additionally test both models with low and high learning rate for the centers(or keys).

\emph{\textbf{Observation of the model:}}
Model with data initialization have centers that look like data, but with larger learning-rate, the centers develop some artifacts.
Model trained with random initialization have its centers learned to be like inputs itself. The average/mean of the centers representing a same class looks more like input samples even trained with random initialization. The model trained with low learning-rate on randomly initialized centers have its mean class resembling structure of inputs as well. This is shown by Figure~\ref{fig:adv_model_centers}.

\emph{\textbf{Observation of adversarial rejection:}}
We find that the model trained with low learning-rate for center(key) show that \textit{the centers/keys that look like the samples itself is better to reject adversarial examples}. With all attack parameters, the model with center similar to data reject the adversarial examples the most. We show some of the attacks in Figure~\ref{fig:adv_rejection_init_compare}.

Furthermore, we also find that adversarial examples with pixel boundary of adversarial samples limited in range of [-1, 1] similar to input has lower adversarial rejection success than bounded in range of [-10, 10] as shown in Figure~\ref{fig:adv_rejection_bound_compare}. This is primarily because large bound allows adversarial examples to occur very far away from centers, making it easier for epsilon neuron to activate and to reject.

\emph{\textbf{Observation of adversarial samples:}}
We generate adversarial examples using epsilon that is lower than best, best, and higher than best to analyze the adversarial examples itself.
We find adversarial examples very much related to centers. Firstly, using the random centers, we encounter adversarial examples that look like random noise. When we tested with data based centers, we find adversarial examples that look like real samples but are able to fool the model.

\begin{figure*}
     \begin{subfigure}{0.32\linewidth}
         \centering
         \fbox{\includegraphics[width=\textwidth]{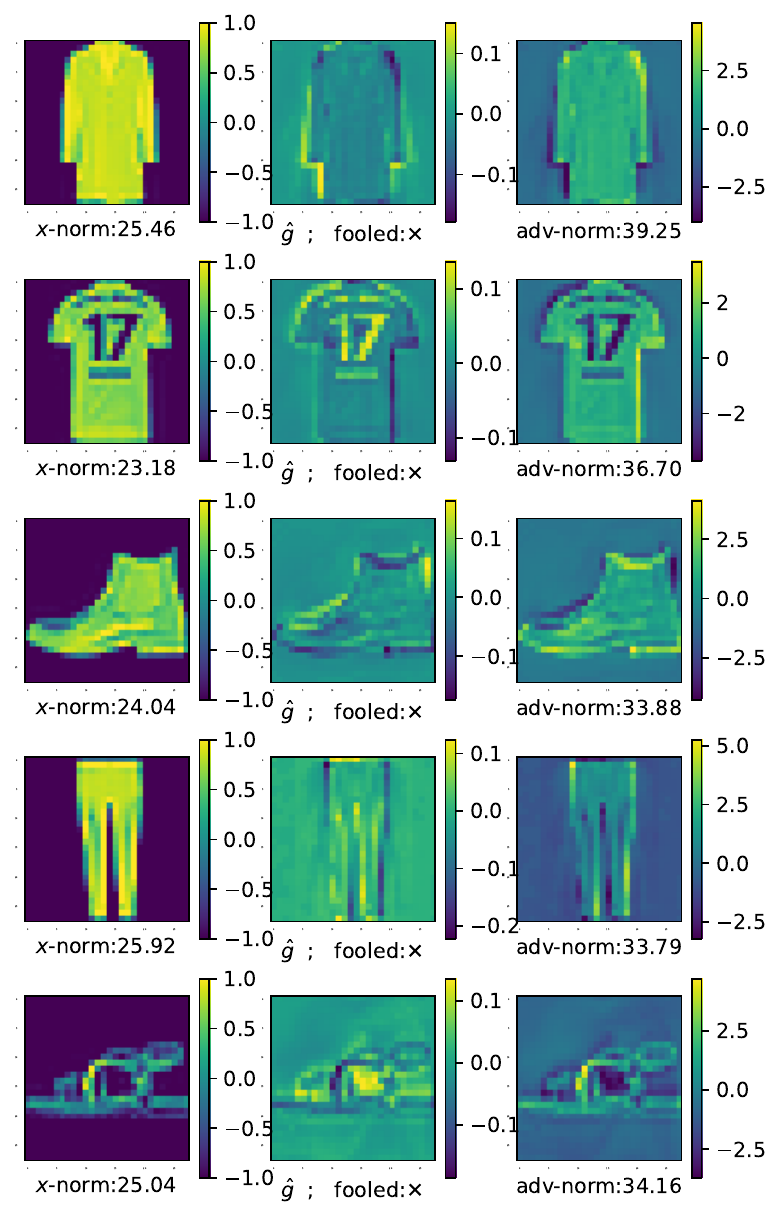}}
     \end{subfigure}\hfill
     \begin{subfigure}{0.32\textwidth}
         \centering
         \fbox{\includegraphics[width=\textwidth]{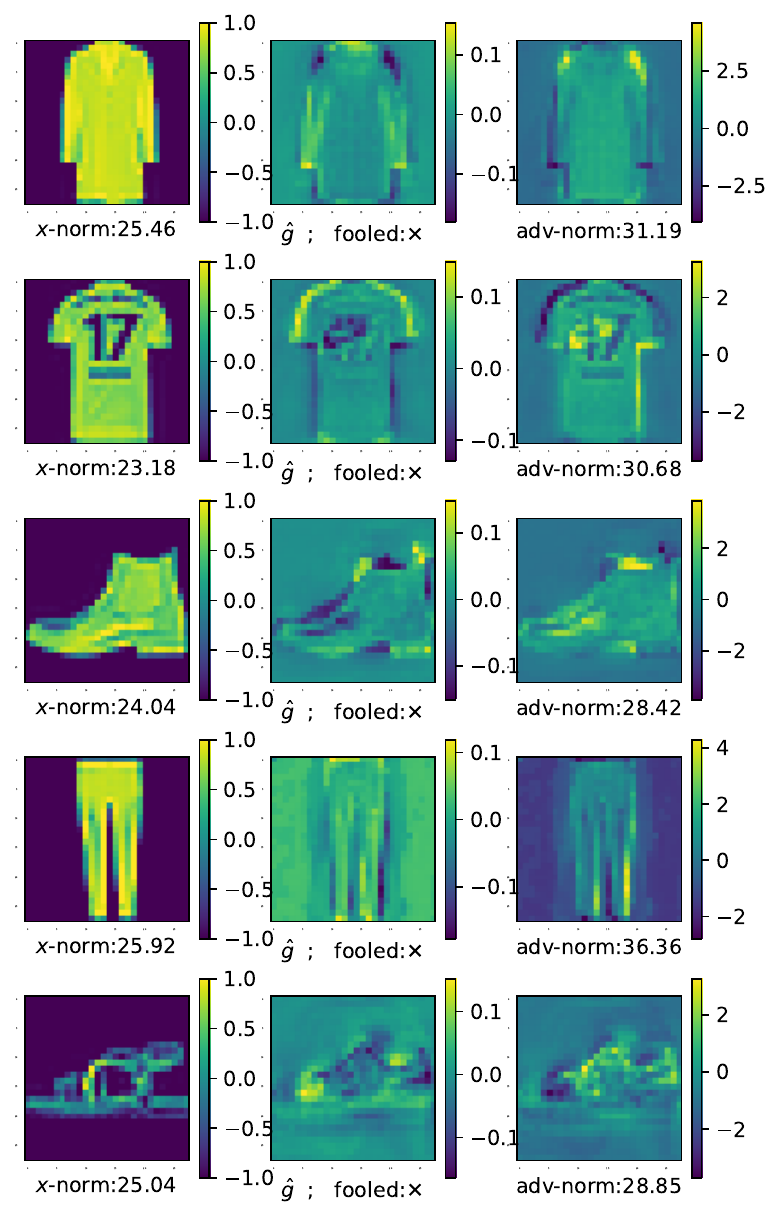}}
     \end{subfigure}\hfill
      \begin{subfigure}{0.32\linewidth}
         \centering
         \fbox{\includegraphics[width=\textwidth]{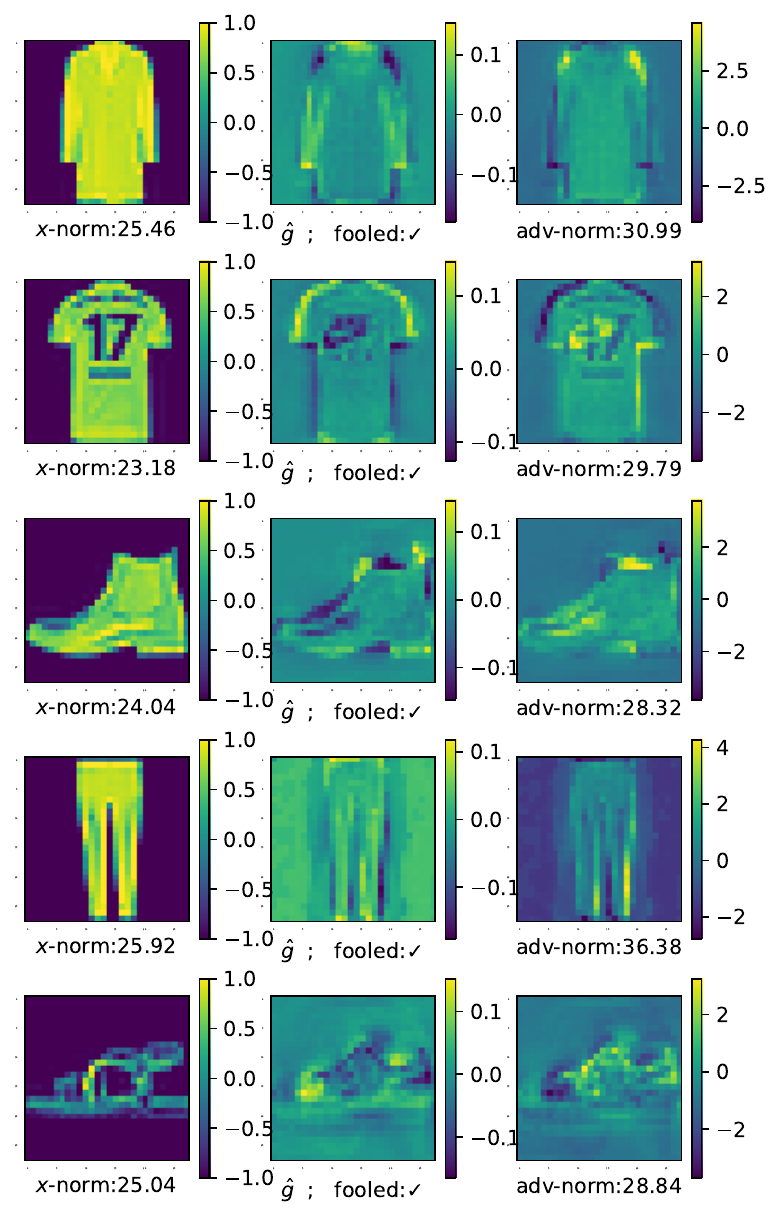}}
     \end{subfigure}
     \newline
        \caption{Adversarial Examples with FGM atack method and bound of [-10, 10] on model trained with center-leaning-rate scaled by 0.01 using data initialisation (Reference~\ref{fig:adv_rejection_init_compare}). The epsilons($\epsilon$) from left to right are 2.0, 15.12, 45.68. For low $\epsilon$ (2 here), not only adversarial examples but inputs($\vx$) are also rejected.}
        \label{fig:adv_examples_viz_data_init}
\end{figure*}

\begin{figure*}
     \begin{subfigure}{0.32\linewidth}
         \centering
         \fbox{\includegraphics[width=\textwidth]{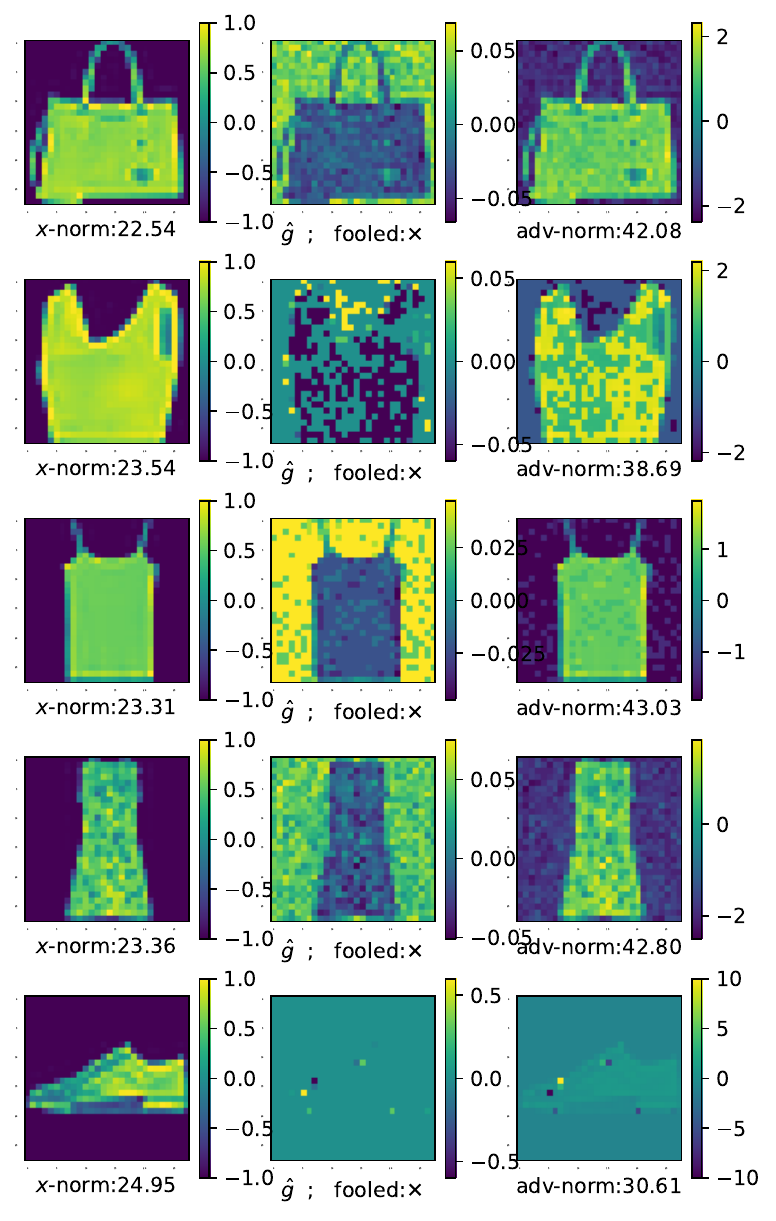}}
     \end{subfigure}\hfill
     \begin{subfigure}{0.32\textwidth}
         \centering
         \fbox{\includegraphics[width=\textwidth]{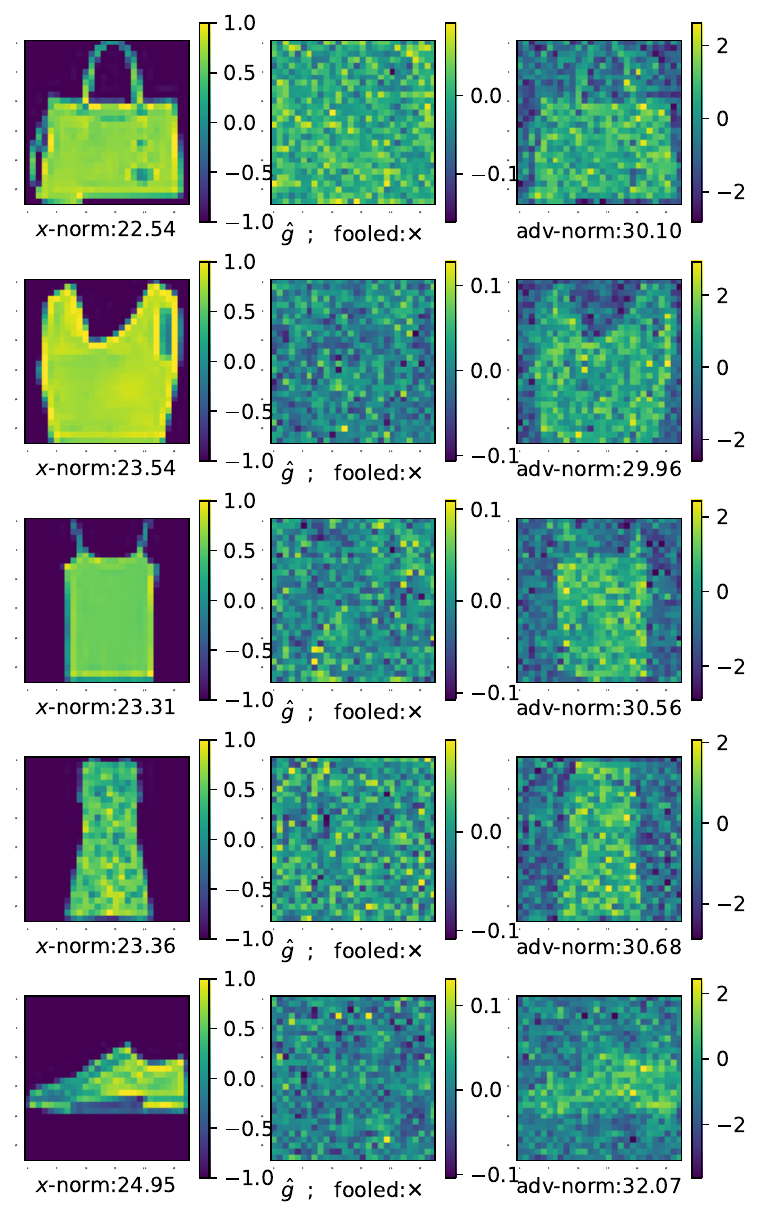}}
     \end{subfigure}\hfill
      \begin{subfigure}{0.32\linewidth}
         \centering
         \fbox{\includegraphics[width=\textwidth]{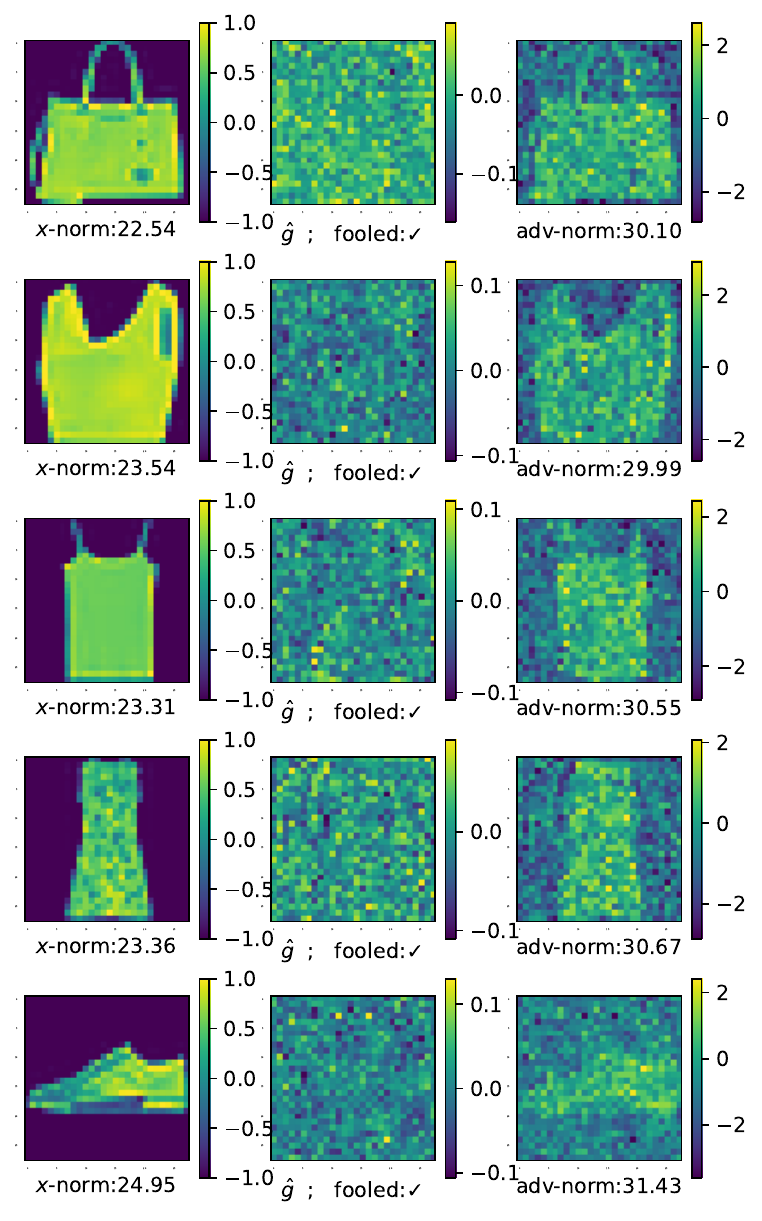}}
     \end{subfigure}
     \newline
        \caption{Adversarial Examples with FGM atack method and bound of [-10, 10] on model trained with center-leaning-rate scaled by 0.01 using random initialisation (Reference~\ref{fig:adv_rejection_init_compare}). The epsilons from left to right are 9.0, 25.31, 50.11. }
        \label{fig:adv_examples_viz_center_init}
\end{figure*}

When we take a look at adversarial attack with low epsilon, we find approximately negative of sample as gradient. This might need more evaluation to understand. Some select examples are shown in Figure~\ref{fig:adv_examples_viz_data_init} and \ref{fig:adv_examples_viz_center_init}.

\textbf{For Classification Layer}\\
We can simply reduce dictionary MLP to classifier along with uncertainty neuron in classification layer. For reduction, we need to have $I$dentity mapping as $V$alues of MLP. In such situation, number of neurons equals number of classes.


\subsubsection{$\epsilon$-Embedding for UMAP}

Similar to adding a epsilon value itself as a neuron, we add a point in embedding space to represent the epsilon neuron. We concatenate epsilon value to distance to calculate nearest-neighbours in the UMAP algorithm (We implemented UMAP, and is not identical with the library).

\begin{figure*}
 \includegraphics[width=0.245\textwidth]{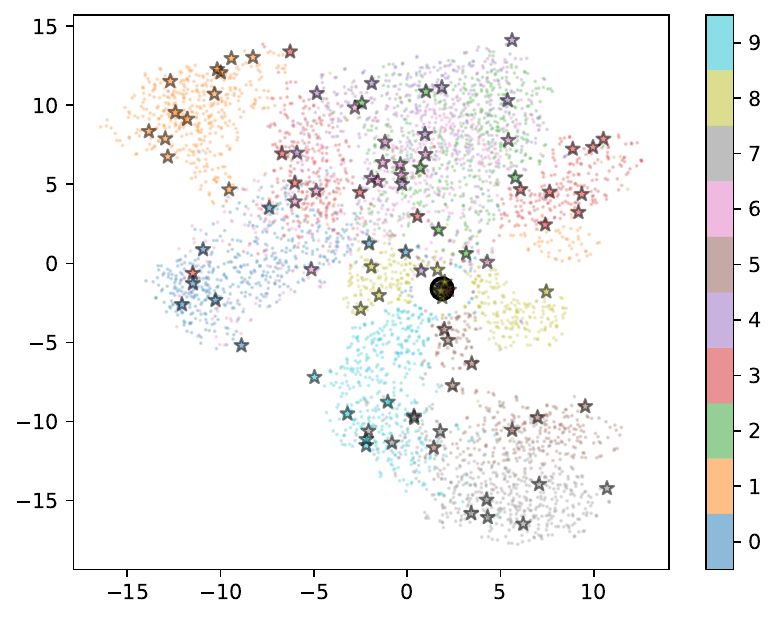}
 \includegraphics[width=0.245\textwidth]{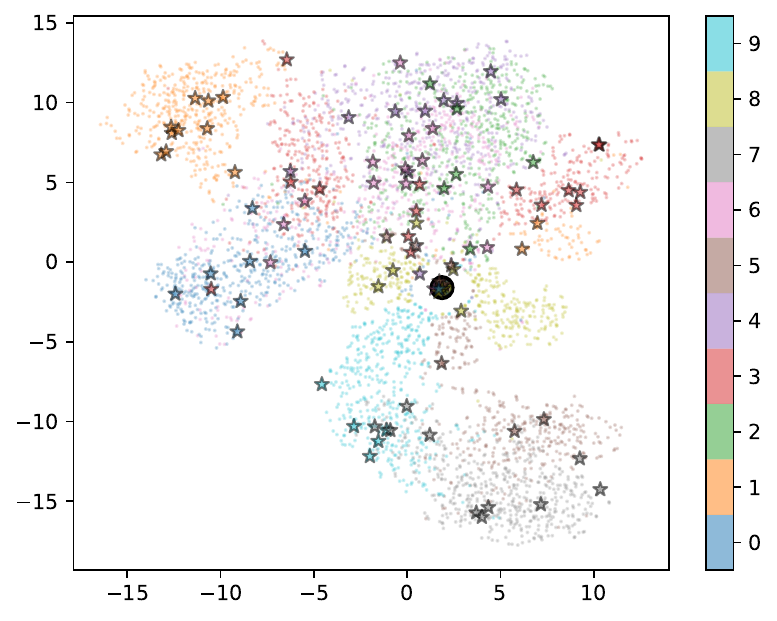}
 \includegraphics[width=0.245\textwidth]{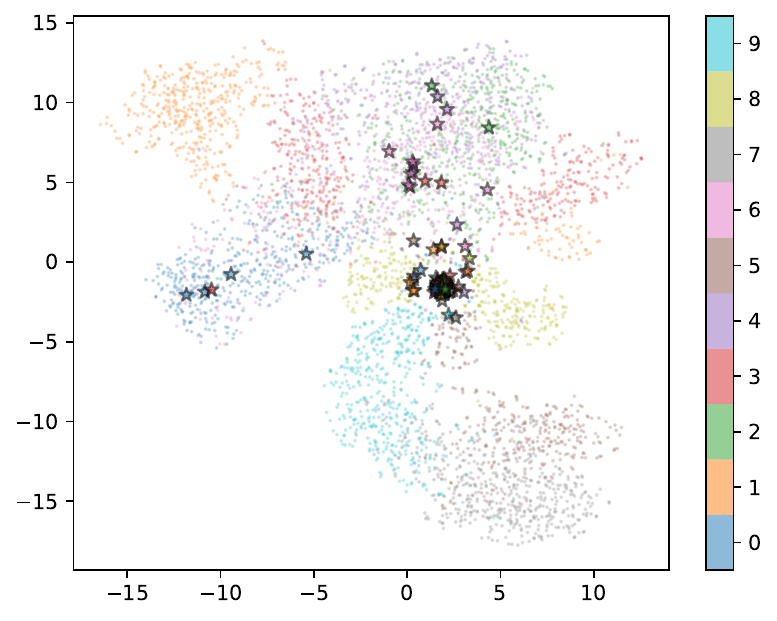}
 \includegraphics[width=0.245\textwidth]{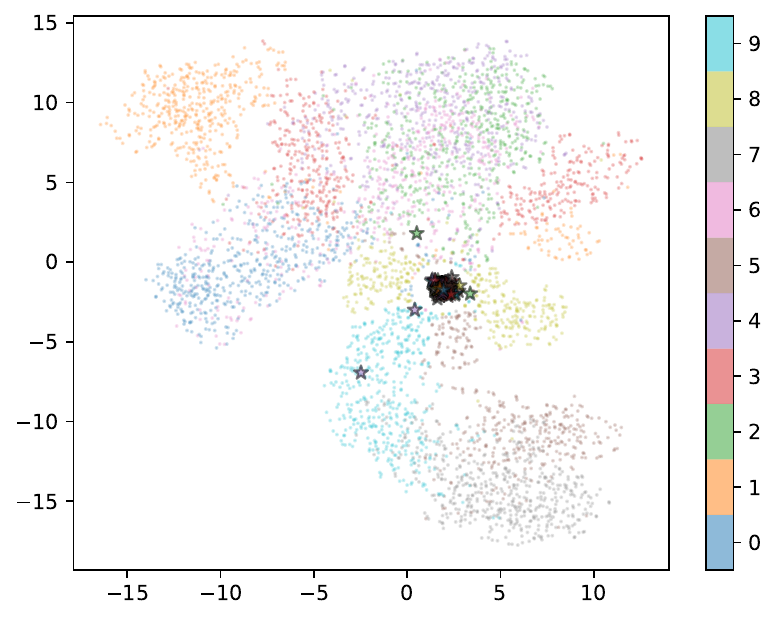}
    \caption{UMAP embedding with epsilon neuron. The stars are interpolated samples. The interpolation towards random - from left to right is [0.0, 0.3, 0.4, 0.5]. The circle in the middle is the $\epsilon$-embedding. We use $\epsilon=6$ with our own implementation of UMAP on PyTorch.}
    \label{fig:epsilon_on_umap}
\end{figure*}

In the experiment, we analyze the interpolation($\alpha$) between image($\vi$) and random uniform noise($\vr$) by plotting its embedding of $\vx = (1-\alpha)\vi + \alpha\vr$. As expected, we find random noise images move towards the embedding of $\epsilon$-distance. This means that value of $\epsilon$ is less than the distance with training samples in high dimension space. This is depicted by Figure~\ref{fig:epsilon_on_umap}

%% file: 14_noisy_center_search_and_data_init.tex
\subsection{Noisy Center Search}

Noisy search by adding, finetuening and pruning has been used in Architecture search~\cite{gordon2018morphnet, dai2020incremental, sapkota2022noisy}. In evolution based algorithms as well the solution is found by noisily mutating and selecting the population. In our case, we noisily search for best centers/target pair that reduces loss.

\begin{figure*}
     \begin{subfigure}{0.36\linewidth}
         \centering
         \includegraphics[width=\textwidth]{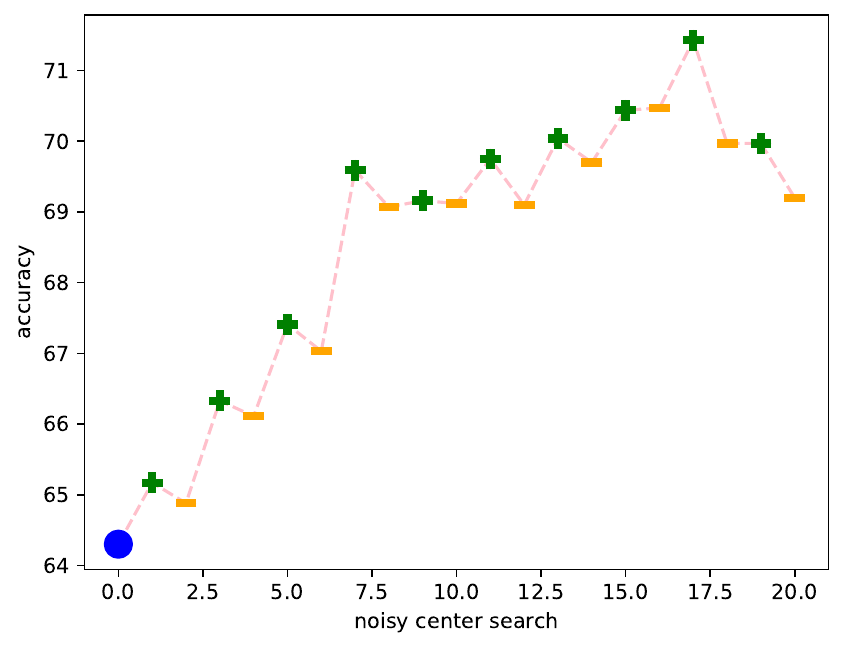}
     \end{subfigure}\hfill
     \begin{subfigure}{0.63\textwidth}
         \centering
         \includegraphics[width=\textwidth]{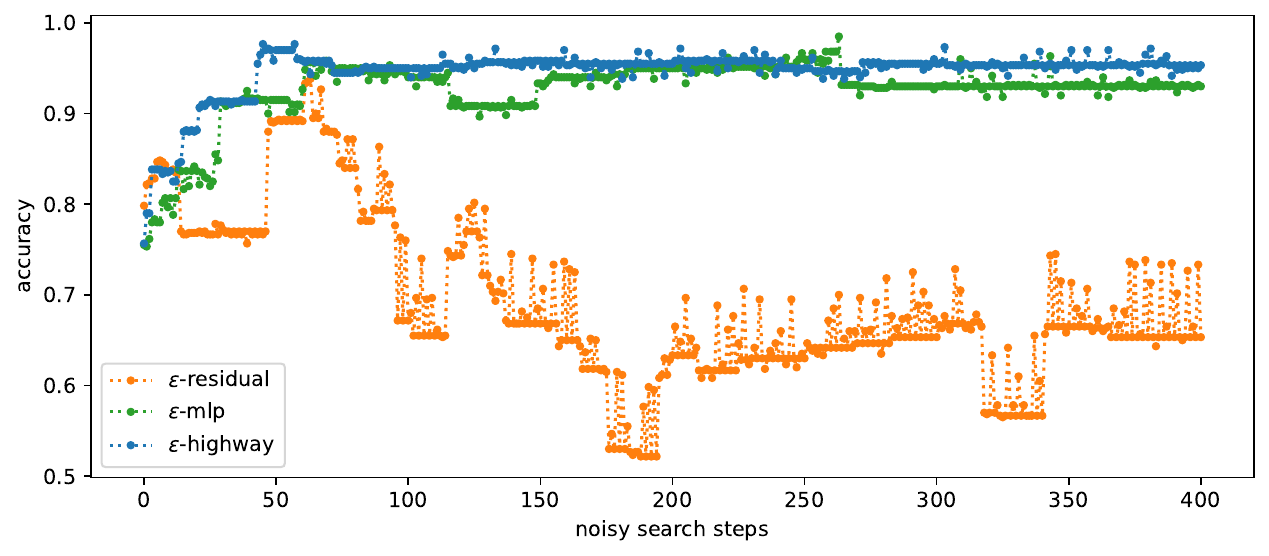}
     \end{subfigure}
     \newline
        \caption{(LEFT) Noisy search for FMNIST dataset. (RIGHT) Noisy search with $\epsilon$ based MLP, ResMLP, and HighwayMLP.}
        \label{fig:noisy_center_search}
\end{figure*}

In FMNIST experiment, we have constant hidden units of 100, and we have 30 search units. So, 30 neurons are added (optionally fine-tuned) and again 30 are removed. Our goal is to add random samples and remove worst performing local neurons recursively. Doing this, without finetuening improves the performance (accuracy) as shown in Figure~\ref{fig:noisy_center_search}. We were also successful in noisy search combined with finetuening.

This is important as centers can move away from exact points. If we want centers to be some of the data, we can optimize without backprop~\cite{}, by noisily searching for centers, and its target.

\paragraph{$\epsilon$-Highway MLP}
Furthermore, we tried noisy center search with Local-Residual but failed to get stable solution that was converging. This motivated us to explore and modify Highway-Nets~\cite{srivastava2015highway}, where gating is done by $\epsilon$-neuron. The $\epsilon$-Highway MLP is given as :
$$\va_{keys}, a_{\epsilon} = f_{\epsilon-softmax-sim} (\vx, \mathbf{K}, \tau)$$
$$\vy = a_{\epsilon}\mathbf{x} + \va_{keys}.\mathbf{V}$$

In 2D toy spiral classification dataset, we successfully converged with $\epsilon$-Highway MLP as shown in Figure~\ref{fig:noisy_center_search}. We use 20 neurons in hidden unit, and search using addition and removal of 1 extra neuron.\\

\textit{Difference in Credit-Assignment:}
The local residual shifts local region and keeps the un-represented regions intact. This mechanism passes the gradient to both residual stream, and to local neuron if the neuron fires. However, if we want to pass gradient to either residual stream or to a local neuron, then we want Highway Network. We can keep the input intact if no local neuron represents it, and output a value if a neuron represents.

%% file: 00_main.bbl
\begin{thebibliography}{40}
\providecommand{\natexlab}[1]{#1}
\providecommand{\url}[1]{\texttt{#1}}
\expandafter\ifx\csname urlstyle\endcsname\relax
  \providecommand{\doi}[1]{doi: #1}\else
  \providecommand{\doi}{doi: \begingroup \urlstyle{rm}\Url}\fi

\bibitem[Amos et~al.(2017)Amos, Xu, and Kolter]{amos2017input}
Brandon Amos, Lei Xu, and J~Zico Kolter.
\newblock Input convex neural networks.
\newblock In \emph{International Conference on Machine Learning}, pp.\  146--155. PMLR, 2017.

\bibitem[Ba et~al.(2016)Ba, Kiros, and Hinton]{ba2016layer}
Jimmy~Lei Ba, Jamie~Ryan Kiros, and Geoffrey~E Hinton.
\newblock Layer normalization.
\newblock \emph{arXiv preprint arXiv:1607.06450}, 2016.

\bibitem[Bancroft(1985)]{bancroft1985algebraic}
Stephen Bancroft.
\newblock An algebraic solution of the gps equations.
\newblock \emph{IEEE transactions on Aerospace and Electronic Systems}, \penalty0 (1):\penalty0 56--59, 1985.

\bibitem[Bishop \& Nasrabadi(2006)Bishop and Nasrabadi]{bishop2006pattern}
Christopher~M Bishop and Nasser~M Nasrabadi.
\newblock \emph{Pattern recognition and machine learning}, volume~4.
\newblock Springer, 2006.

\bibitem[Broomhead \& Lowe(1988)Broomhead and Lowe]{broomhead1988multivariable}
D~Broomhead and D~Lowe.
\newblock Multivariable functional interpolation and adaptive networks, complex systems, vol. 2, 1988.

\bibitem[Chen et~al.(2020)Chen, Wang, Xu, Shi, Xu, Tian, and Xu]{chen2020addernet}
Hanting Chen, Yunhe Wang, Chunjing Xu, Boxin Shi, Chao Xu, Qi~Tian, and Chang Xu.
\newblock Addernet: Do we really need multiplications in deep learning?
\newblock In \emph{Proceedings of the IEEE/CVF conference on computer vision and pattern recognition}, pp.\  1468--1477, 2020.

\bibitem[Clevert et~al.(2015)Clevert, Unterthiner, and Hochreiter]{clevert2015fast}
Djork-Arn{\'e} Clevert, Thomas Unterthiner, and Sepp Hochreiter.
\newblock Fast and accurate deep network learning by exponential linear units (elus).
\newblock \emph{arXiv preprint arXiv:1511.07289}, 2015.

\bibitem[Dai et~al.(2020)Dai, Yin, and Jha]{dai2020incremental}
Xiaoliang Dai, Hongxu Yin, and Niraj~K Jha.
\newblock Incremental learning using a grow-and-prune paradigm with efficient neural networks.
\newblock \emph{IEEE Transactions on Emerging Topics in Computing}, 10\penalty0 (2):\penalty0 752--762, 2020.

\bibitem[Elhage et~al.(2022)Elhage, Hume, Olsson, Nanda, Henighan, Johnston, ElShowk, Joseph, DasSarma, Mann, Hernandez, Askell, Ndousse, Jones, Drain, Chen, Bai, Ganguli, Lovitt, Hatfield-Dodds, Kernion, Conerly, Kravec, Fort, Kadavath, Jacobson, Tran-Johnson, Kaplan, Clark, Brown, McCandlish, Amodei, and Olah]{elhage2022solu}
Nelson Elhage, Tristan Hume, Catherine Olsson, Neel Nanda, Tom Henighan, Scott Johnston, Sheer ElShowk, Nicholas Joseph, Nova DasSarma, Ben Mann, Danny Hernandez, Amanda Askell, Kamal Ndousse, Andy Jones, Dawn Drain, Anna Chen, Yuntao Bai, Deep Ganguli, Liane Lovitt, Zac Hatfield-Dodds, Jackson Kernion, Tom Conerly, Shauna Kravec, Stanislav Fort, Saurav Kadavath, Josh Jacobson, Eli Tran-Johnson, Jared Kaplan, Jack Clark, Tom Brown, Sam McCandlish, Dario Amodei, and Christopher Olah.
\newblock Softmax linear units.
\newblock \emph{Transformer Circuits Thread}, 2022.
\newblock https://transformer-circuits.pub/2022/solu/index.html.

\bibitem[Fortune(2017)]{fortune2017voronoi}
Steven Fortune.
\newblock Voronoi diagrams and delaunay triangulations.
\newblock In \emph{Handbook of discrete and computational geometry}, pp.\  705--721. Chapman and Hall/CRC, 2017.

\bibitem[Goodfellow et~al.(2014)Goodfellow, Shlens, and Szegedy]{goodfellow2014explaining}
Ian~J Goodfellow, Jonathon Shlens, and Christian Szegedy.
\newblock Explaining and harnessing adversarial examples.
\newblock \emph{arXiv preprint arXiv:1412.6572}, 2014.

\bibitem[Gordon et~al.(2018)Gordon, Eban, Nachum, Chen, Wu, Yang, and Choi]{gordon2018morphnet}
Ariel Gordon, Elad Eban, Ofir Nachum, Bo~Chen, Hao Wu, Tien-Ju Yang, and Edward Choi.
\newblock Morphnet: Fast \& simple resource-constrained structure learning of deep networks.
\newblock In \emph{Proceedings of the IEEE conference on computer vision and pattern recognition}, pp.\  1586--1595, 2018.

\bibitem[He et~al.(2016)He, Zhang, Ren, and Sun]{he2016deep}
Kaiming He, Xiangyu Zhang, Shaoqing Ren, and Jian Sun.
\newblock Deep residual learning for image recognition.
\newblock In \emph{Proceedings of the IEEE conference on computer vision and pattern recognition}, pp.\  770--778, 2016.

\bibitem[Hooker(2021)]{hooker2021hardware}
Sara Hooker.
\newblock The hardware lottery.
\newblock \emph{Communications of the ACM}, 64\penalty0 (12):\penalty0 58--65, 2021.

\bibitem[Ioffe \& Szegedy(2015)Ioffe and Szegedy]{ioffe2015batch}
Sergey Ioffe and Christian Szegedy.
\newblock Batch normalization: Accelerating deep network training by reducing internal covariate shift.
\newblock In \emph{International conference on machine learning}, pp.\  448--456. pmlr, 2015.

\bibitem[Krizhevsky et~al.(2009)Krizhevsky, Hinton, et~al.]{krizhevsky2009learning}
Alex Krizhevsky, Geoffrey Hinton, et~al.
\newblock Learning multiple layers of features from tiny images.
\newblock 2009.

\bibitem[Krizhevsky et~al.(2012)Krizhevsky, Sutskever, and Hinton]{krizhevsky2012imagenet}
Alex Krizhevsky, Ilya Sutskever, and Geoffrey~E Hinton.
\newblock Imagenet classification with deep convolutional neural networks.
\newblock \emph{Advances in neural information processing systems}, 25:\penalty0 1097--1105, 2012.

\bibitem[Li et~al.(2022)Li, Parazeres, Oberman, Ghaffari, Asgharian, and Nia]{li2022euclidnets}
Xinlin Li, Mariana Parazeres, Adam Oberman, Alireza Ghaffari, Masoud Asgharian, and Vahid~Partovi Nia.
\newblock Euclidnets: An alternative operation for efficient inference of deep learning models.
\newblock \emph{arXiv preprint arXiv:2212.11803}, 2022.

\bibitem[Ma(2000)]{ma2000bisectors}
Lihong Ma.
\newblock \emph{Bisectors and Voronoi diagrams for convex distance functions}.
\newblock Citeseer, 2000.

\bibitem[McInnes et~al.(2018)McInnes, Healy, and Melville]{mcinnes2018umap}
Leland McInnes, John Healy, and James Melville.
\newblock Umap: Uniform manifold approximation and projection for dimension reduction.
\newblock \emph{arXiv preprint arXiv:1802.03426}, 2018.

\bibitem[Miller()]{0softmax}
Evan Miller.
\newblock Attention is off by one.
\newblock https://www.evanmiller.org/attention-is-off-by-one.html.

\bibitem[Nesterov et~al.(2022)Nesterov, Torres, Nagy-Huber, Samarin, and Roth]{nesterov2022learning}
Vitali Nesterov, Fabricio~Arend Torres, Monika Nagy-Huber, Maxim Samarin, and Volker Roth.
\newblock Learning invariances with generalised input-convex neural networks.
\newblock \emph{arXiv preprint arXiv:2204.07009}, 2022.

\bibitem[Norrdine(2012)]{norrdine2012algebraic}
Abdelmoumen Norrdine.
\newblock An algebraic solution to the multilateration problem.
\newblock In \emph{Proceedings of the 15th international conference on indoor positioning and indoor navigation, Sydney, Australia}, volume 1315, 2012.

\bibitem[Olah et~al.(2017)Olah, Mordvintsev, and Schubert]{olah2017feature}
Chris Olah, Alexander Mordvintsev, and Ludwig Schubert.
\newblock Feature visualization.
\newblock \emph{Distill}, 2017.
\newblock \doi{10.23915/distill.00007}.
\newblock https://distill.pub/2017/feature-visualization.

\bibitem[Olah(2014)]{olah2014neural}
Christopher Olah.
\newblock Neural networks, manifolds, and topology.
\newblock \emph{Blog post}, 2014.

\bibitem[Rauber et~al.(2020)Rauber, Zimmermann, Bethge, and Brendel]{rauber2017foolboxnative}
Jonas Rauber, Roland Zimmermann, Matthias Bethge, and Wieland Brendel.
\newblock Foolbox native: Fast adversarial attacks to benchmark the robustness of machine learning models in pytorch, tensorflow, and jax.
\newblock \emph{Journal of Open Source Software}, 5\penalty0 (53):\penalty0 2607, 2020.
\newblock \doi{10.21105/joss.02607}.
\newblock URL \url{https://doi.org/10.21105/joss.02607}.

\bibitem[Rumelhart et~al.(1986)Rumelhart, Hinton, and Williams]{rumelhart1986learning}
David~E Rumelhart, Geoffrey~E Hinton, and Ronald~J Williams.
\newblock Learning representations by back-propagating errors.
\newblock \emph{nature}, 323\penalty0 (6088):\penalty0 533--536, 1986.

\bibitem[Sapkota \& Bhattarai(2021)Sapkota and Bhattarai]{sapkota2021input}
Suman Sapkota and Binod Bhattarai.
\newblock Input invex neural network.
\newblock \emph{arXiv preprint arXiv:2106.08748}, 2021.

\bibitem[Sapkota \& Bhattarai(2022)Sapkota and Bhattarai]{sapkota2022noisy}
Suman Sapkota and Binod Bhattarai.
\newblock Noisy heuristics nas: A network morphism based neural architecture search using heuristics.
\newblock \emph{arXiv preprint arXiv:2207.04467}, 2022.

\bibitem[Srivastava et~al.(2015)Srivastava, Greff, and Schmidhuber]{srivastava2015highway}
Rupesh~Kumar Srivastava, Klaus Greff, and J{\"u}rgen Schmidhuber.
\newblock Highway networks.
\newblock \emph{arXiv preprint arXiv:1505.00387}, 2015.

\bibitem[Sukhbaatar et~al.(2015)Sukhbaatar, Weston, Fergus, et~al.]{sukhbaatar2015end}
Sainbayar Sukhbaatar, Jason Weston, Rob Fergus, et~al.
\newblock End-to-end memory networks.
\newblock \emph{Advances in neural information processing systems}, 28, 2015.

\bibitem[Szegedy et~al.(2013)Szegedy, Zaremba, Sutskever, Bruna, Erhan, Goodfellow, and Fergus]{szegedy2013intriguing}
Christian Szegedy, Wojciech Zaremba, Ilya Sutskever, Joan Bruna, Dumitru Erhan, Ian Goodfellow, and Rob Fergus.
\newblock Intriguing properties of neural networks.
\newblock \emph{arXiv preprint arXiv:1312.6199}, 2013.

\bibitem[Tolstikhin et~al.(2021)Tolstikhin, Houlsby, Kolesnikov, Beyer, Zhai, Unterthiner, Yung, Steiner, Keysers, Uszkoreit, et~al.]{tolstikhin2021mlp}
Ilya~O Tolstikhin, Neil Houlsby, Alexander Kolesnikov, Lucas Beyer, Xiaohua Zhai, Thomas Unterthiner, Jessica Yung, Andreas Steiner, Daniel Keysers, Jakob Uszkoreit, et~al.
\newblock Mlp-mixer: An all-mlp architecture for vision.
\newblock \emph{Advances in Neural Information Processing Systems}, 34:\penalty0 24261--24272, 2021.

\bibitem[Vaswani et~al.(2017)Vaswani, Shazeer, Parmar, Uszkoreit, Jones, Gomez, Kaiser, and Polosukhin]{vaswani2017attention}
Ashish Vaswani, Noam Shazeer, Niki Parmar, Jakob Uszkoreit, Llion Jones, Aidan~N Gomez, {\L}ukasz Kaiser, and Illia Polosukhin.
\newblock Attention is all you need.
\newblock \emph{Advances in neural information processing systems}, 30, 2017.

\bibitem[Voronoi(1908)]{voronoi1908nouvelles}
Georges Voronoi.
\newblock Nouvelles applications des param{\`e}tres continus {\`a} la th{\'e}orie des formes quadratiques. deuxi{\`e}me m{\'e}moire. recherches sur les parall{\'e}llo{\`e}dres primitifs.
\newblock \emph{Journal f{\"u}r die reine und angewandte Mathematik (Crelles Journal)}, 1908\penalty0 (134):\penalty0 198--287, 1908.

\bibitem[Wikipedia({\natexlab{a}})]{cosine-sim}
Wikipedia.
\newblock Cosine-similarity.
\newblock https://en.wikipedia.org/wiki/Cosine\_similarity, {\natexlab{a}}.

\bibitem[Wikipedia({\natexlab{b}})]{lp-space}
Wikipedia.
\newblock Lp-space.
\newblock https://en.wikipedia.org/wiki/Lp\_space, {\natexlab{b}}.

\bibitem[Wikipedia({\natexlab{c}})]{metric-space}
Wikipedia.
\newblock Metric-space.
\newblock https://en.wikipedia.org/wiki/Metric\_space, {\natexlab{c}}.

\bibitem[Wikipedia({\natexlab{d}})]{softmax}
Wikipedia.
\newblock Softmax function.
\newblock https://en.wikipedia.org/wiki/Softmax\_function, {\natexlab{d}}.

\bibitem[Wikipedia({\natexlab{e}})]{stereographic_projection}
Wikipedia.
\newblock Stereographic projection.
\newblock https://en.wikipedia.org/wiki/Stereographic\_projection, {\natexlab{e}}.

\end{thebibliography}
